\documentclass[acmsmall, manuscript, screen]{acmart}
\AtBeginDocument{%
  }

\setcopyright{acmlicensed}
\copyrightyear{2025}
\acmYear{2025}
\acmDOI{10.1145/3767735}
\acmJournal{TKDD}
\usepackage[T1]{fontenc}
\usepackage{babel}

\usepackage{graphicx}

\usepackage{amssymb}

\usepackage{type1ec}
\usepackage{algorithmic}
\usepackage{lscape} 
\usepackage{multirow}

\usepackage{blindtext}
\usepackage{multicol}

\usepackage{amsmath,amssymb,amsfonts}
\usepackage{textcomp}
\usepackage{xcolor}
\usepackage{color}
\usepackage{gensymb}
\usepackage{pstricks}
\usepackage{subcaption}
\usepackage{csquotes}
\usepackage{caption}
\usepackage[ruled,vlined]{algorithm2e}
\usepackage[dvipsnames]{xcolor}

\usepackage{mathtools}

\begin{document}

\title{"A 6 or a 9?": Ensemble Learning Through the Multiplicity of Performant Models and Explanations}


\author{Gianlucca Zuin}
\email{gianlucca@kunumi.com}
\orcid{0000-0002-0429-3280}
\affiliation{%
  \institution{Universidade Federal de Minas Gerais}
  \country{Brazil}
}\affiliation{%
\institution{Instituto Kunumi}
  \country{Brazil}
}

\author{Adriano Veloso}
\email{adriano@kunumi.com}
\orcid{0000-0002-9177-4954}
\affiliation{%
  \institution{Universidade Federal de Minas Gerais}
  \country{Brazil}
}\affiliation{%
\institution{Instituto Kunumi}
  \country{Brazil}
}

%\renewcommand{\shortauthors}{Double-blind review}

\begin{abstract}
  Creating models from past observations and ensuring their effectiveness on new data is the essence of machine learning. However, selecting models that generalize well remains a challenging task. Related to this topic, the Rashomon Effect refers to cases where multiple models perform similarly well for a given learning problem. This often occurs in real-world scenarios, like the manufacturing process or medical diagnosis, where diverse patterns in data lead to multiple high-performing solutions. We propose the Rashomon Ensemble, a method that strategically selects models from these diverse high-performing solutions to improve generalization. By grouping models based on both their performance and explanations, we construct ensembles that maximize diversity while maintaining predictive accuracy. This selection ensures that each model covers a distinct region of the solution space, making the ensemble more robust to distribution shifts and variations in unseen data. We validate our approach on both open and proprietary collaborative real-world datasets, demonstrating up to 0.20+ AUROC improvements in scenarios where the Rashomon ratio is large. Additionally, we demonstrate tangible benefits for businesses in various real-world applications, highlighting the robustness, practicality, and effectiveness of our approach.
\end{abstract}





\maketitle

\section{Introduction}

Model selection is crucial in industry and research, and the widely adopted approach is cross‐validation. Although cross‐validation generally provides robust risk estimation \cite{cvsurvey}, it can fail for specific problems depending on the goal of model selection, and empirical risk in a test set might not always correlate with real‐world performance \cite{damour2020underspecification}. Empirical risk can be significantly affected when different models perform equally well on the test set \cite{hinns2021initial}. This limitation of relying solely on empirical risk motivates us to explore alternative evaluation approaches that capture subtle differences in model behavior.

The Rashomon Effect, also known as the multiplicity of good models \cite{breiman2001statistical}, presents a phenomenon where many models perform equally well. However, these models may process data in substantially different ways, making it challenging to draw reliable conclusions or automate decisions based on a single model fit \cite{zuintese}. This inherent diversity in model behavior emphasizes why a single performance metric can be misleading. A significant challenge arises when a cross-validated model, carefully selected during training, encounters data drawn from a different distribution during production. In these cases, even small internal differences among models may lead to divergent outcomes. Cross-validation guarantees no longer apply to out-of-distribution data, resulting in unpredictable model performance and rendering held-out performance an unreliable risk estimate. To address this issue, we extend our analysis beyond empirical risk and explore additional axes to identify more robust models. In doing so, we not only measure performance but also examine how models adapt their behavior across varying data distributions.

Our main hypothesis posits that models exhibit similar behavior only when data are drawn from the same distribution as seen during training. If this holds, we can sample different models and 
verify their outputs during production (i.e., once deployed and used on new data). Disagreement among models would imply that the data are drawn from an unknown distribution, leading to untrustworthy predictions. Conversely, if the models agree, we gain extra confidence in the correctness of their predictions, as diverse models converge on the same conclusion. These models can be used to build an ensemble, and we may establish the Jensen-Shannon distance between the outputs of the ensemble constituents as a production risk metric.

However, we believe that diversity among individual models is still crucial for gaining an understanding of the data. The Rashomon Effect suggests that multiple explanations can exist for a given phenomenon, each consistent with the observed data. To systematically capture this diversity, we can group models based on the similarity of their explanations. Ideally, this leads to dense groups of models that share common factors, from which the most distinct representatives are selected. The traditional approaches attempting to induce a single model from all sub-populations may perform poorly in complex problems where data inherently contain several local structures and sub-populations~\cite{ballard2016dynamic, zuinportinari2020, liang2023accuracy, lorasdagi2025fitting}. Our strategy instead culminates in an ensemble that is both diverse and robust, with all its constituents being performant models focused on a subset of features, a local structure in the data.

Further, as each constituent offers a different explanation for the target phenomenon, the ensemble's output is directly linked to the trustworthiness of predictions. The consensus among the constituents indicates a match between the data distribution and the one seen during training, with all the cross-validation guarantees. Disagreement suggests that cross-validation properties may not be trusted. In our experiments, we consider a simple voting scheme ensemble and the returned voting ratio, that is, the probability predicted by our ensemble, as a measure of agreement. For binary classification problems, such as those evaluated in our experiments, ratios close to $50\%$ indicate the highest uncertainty. We coined this concept the Rashomon Ensemble. In summary, our approach involves the following steps:

\begin{enumerate}
\item {Sampling models from a pre-defined Rashomon subspace (i.e., a set of models with equivalent performance), achieved by drawing different random feature subsets.}
\item {Computing the explanation for each sampled model and quantifying the pairwise similarities among them.}
\item {Perturbing held-out test data through appropriate transformations to evaluate model stability.}
\item {Measuring the pairwise distances on the perturbed dataset to capture divergence in predictions.}
\item {Segmenting the Rashomon Set into subgroups based on the models' explanation vectors and distance metrics.}
\item {Selecting a set of models with contrasting explanations and divergent responses on the perturbed data.}
\item {Constructing an ensemble and evaluating the degree of agreement among models as a proxy for production risk.}
\end{enumerate}

We validate our approach on a set of public datasets for reproducibility and demonstrate its robustness in simulated scenarios. Our results show that Rashomon ensembles consistently outperform state-of-the-art ensemble learning approaches if the Rashomon Set is large enough. When exposed to data drift, our approach remained the performant one in most evaluated scenarios, providing further evidence of its reliability. Further, we also employ the Rashomon ensembles in four real-world applications partnered with various industries and institutions, studying the impact of our approach. We demonstrate how our approach leads to a tangible impact on business, with reported gains of over $R\$1.5$ million and a patent being filed. Our results also consistently show that Rashomon ensembles outperform state-of-the-art ensemble learning approaches when the Rashomon Set is sufficiently large.

The remainder of this work is structured as follows: Section 2 introduces background to support the understanding of the devised techniques, explaining concepts such as the Rashomon Effect, Rashomon Ratio, and our understanding of model explanation. Section 3 discusses related research, including data drift, domain adaptation, ensemble learning, and prior applications of the Rashomon Effect in machine learning. Section 4 formalizes our proposed Rashomon Ensemble, outlining its theoretical framework and implementation. Sections 5 and 6 present experimental evaluations. Section 5 addresses the analysis of our approach's robustness and limitations under publicly available datasets, while Section 6 focuses on multiple real-world case studies using unique collaborative data. Finally, Section 7 brings about the conclusion and proposes directions for future research.

\section{Background}

The idea of capturing model uncertainty by exploring the relationship between test points and the learned model is not new. Typical approaches include building an ensemble of models and measuring inter-model variance \cite{Madras2020Detecting}, or learning a scoring rule that captures ambiguity in targets \cite{simpledeepensembles, caruanaunknowns}. However, most recent research on this topic has mainly focused on Neural Networks and how they learn intermediate features. More specifically, state-of-the-art approaches to Out-of-Distribution (OoD) detection enrich the intermediate feature space beyond what would ordinarily be learned via supervised learning alone, such as encouraging a model to learn high-level task-agnostic semantic features \cite{winkens2020contrastive}, or employing an additionally labeled outlier dataset during training \cite{outlierexprosure}. When one cannot access the intermediate feature space, most of the mentioned approaches fail. As noted by \citet{oodanalyzier}, this type of approach has two drawbacks: first, models trained to identify OoD may fail to cover the entire data distribution; second, explaining the source of OoD may be non-trivial.

The key difference in our work lies in the analysis of additional unexplored axes, such as the decision-making process of a model via its explanatory factors~\cite{zuincartas2, shap}. A second key idea is to exploit the Rashomon Effect by looking for models with similar performance during training. Both of these propositions enable explanation of the risk metric by assigning importance to the factors leading to each model decision and comparing them. Further, our approach is algorithm-agnostic and reproducible with any model that handles tabular data. We therefore summarize three pivotal points underlying our approach: acknowledging that production data may fall outside the training distribution, recognizing the multiplicity of high-performing models, and analyzing the explanatory factors behind model decisions

\vspace{0.1in}
\noindent\textbf{Underspecification:} Underspecification in deep learning arises when models achieve similar in-sample performance but present divergent behaviors on out-of-sample data. This is problematic when some models perform significantly worse in production, creating challenges for proper model selection \cite{damour2020underspecification}. Although much of the underspecification literature focuses on deep neural networks, the phenomenon largely arises from the elevated number of optimized parameters \cite{bui2021underspecification, underantoposofic}. \citet{mei2019generalization} state that this issue is common to any machine learning pipeline. D’Amour et al. observed that repeating a training process can generate many models with identical test performance but significantly different behaviors, even when only minor perturbations are introduced, such as using a different random seed. This, in turn, differentiates each model by small arbitrary learning decisions. Although these differences are usually considered minor, the consequence is varying degrees of performance in the real world. As such, underspecification is closely tied to the Rashomon Effect.

\vspace{0.1in}
\noindent\textbf{Rashomon Effect:} The Rashomon Effect, as analyzed by \citet{rashomon_ml_concept}, refers to the set of models with accuracy close to that of the optimal model. From this set, they formally defined the concept of the \textit{Rashomon Set}, which represents the subspace of the universe of models summarizing the range of effective prediction strategies an optimal analyst might choose. The Rashomon Effect is further explored by \citet{rashomon_ml_subspaces}, who provide pertinent definitions concerning the generalization of the Rashomon Set, its shape, and its volume. In particular, they explore under which situations it is possible to obtain a sample of the model space such that the Rashomon properties of this subspace are similar to those of the full universe.

The Rashomon Set thus comprises a collection of close-to-optimal models that share similar explanations and performance due to the Rashomon Effect. For this, we need a comparison to some key reference model, denoted as $h_{ref}$. Fisher suggests that $h_{ref}$ can be derived from expert knowledge or from some quantitative decision rule implemented in practice. This prespecified reference model serves as a baseline for performance. Thus, if we establish $\epsilon$ as the maximum accepted error relative to $h_{ref}$ for considering a model part of the Rashomon Set, we can denote it as:

\begin{equation}
R(H, \epsilon):=\{h \in H : E[L(h, Z)] \leq E[L(h_{ref}, Z)] + \epsilon\}
\end{equation}

\noindent where $E$ denotes expectations concerning the population distribution, $L$ is some nonnegative loss function, $H$ is the hypothesis space, and $Z = [Y X]$ is the data. The $\epsilon$ metric takes into account models that might be arrived at due to differences in data measurement, processing, filtering, model parameterization, covariate selection, or other analysis choices.

Further, let $X_1 \in X$ be the set of all features that model $h_{ref}$ relies on to reach a prediction. This reliance metric has a direct relationship with model explanation. We can expect models that rely too heavily on $X_1$ to be prone to high variance, leading to poor performance. Likewise, models that rely too little on $X_1$ are prone to high bias, also leading to poor performance. Model reliance (MR) of variable $X_1$ can be computed as the increase in expected loss when the contribution of this variable is removed by random permutation. The range of all possible MR values within this class gives rise to the notion of Model Class Reliance (MCR), which helps define a minimum and maximum MR value to classify a model $f$ within the set defined by $h_{ref}, \epsilon, \text{ and } X_1$. These models comprise the set of high-performing models that also share a similar reliance on $X_1$ as the reference model $h_{ref}$, as proposed by Fisher et al.

\vspace{0.1in}
\noindent\textbf{Rashomon Ratio:}  
The concept of the Rashomon ratio, as introduced by Semenova and Rudin~\cite{rashomon_ml_subspaces}, quantifies the fraction of models within the hypothesis space that perform nearly as well as a reference model. Given a hypothesis space \(H\) and a subset \(R \subseteq H\) of good models (the Rashomon Set), the ratio is defined as:

\begin{equation}
R_{ratio} = \frac{|R|}{|H|}
\end{equation}

\noindent Here, the notation \( | \cdot | \) is used to denote the size of a set, and its interpretation depends on the hypothesis space shape. For discrete spaces, that is, when \(H\) is discrete and finite, \( |R| \) and \( |H| \) represent the cardinalities of the Rashomon Set and the full hypothesis space, respectively:
    \[
    |R| = \sum_{h \in H} \mathbf{1}\{h \in R\}, \quad |H| = \sum_{h \in H} 1
    \]
in which \(\mathbf{1}\{\cdot\}\) is the indicator function.

For continuous hypothesis spaces, that is, in scenarios where \(H\) is continuous or infinite, \( |R| \) and \( |H| \) can be interpreted as volumes under a chosen measure \(\mathcal{V}(\cdot)\), such that:
    \[
    |R| \equiv \mathcal{V}(R) \quad \text{and} \quad |H| \equiv \mathcal{V}(H)
    \]
    \noindent in which the ratio \( \frac{\mathcal{V}(R)}{\mathcal{V}(H)} \) is well-defined under the assumption of a uniform prior over \(H\).

However, even for discrete hypothesis spaces, the exact computation of \(R_{ratio}\) would involve evaluating every model \(f \in H\), which may be computationally infeasible. Thus, we approximate the Rashomon ratio by random sampling: models are drawn from \(H\) and the set, the fraction of sampled models, that lie in \(\hat{R}\) serve as an empirical estimate \(\hat{R}_{ratio}\) of this discrete hypothesis subspace. If the sample is large enough, this estimate holds guarantees of similarity to the true Rashomon Set, as stated by \citet{rashomon_ml_concept}.

\vspace{0.1in}
\noindent\textbf{Data drift:} Let $T$ be the train distribution of the source data, and $U$ be some unknown distribution from another dataset. \citet{candela2009dataset} defines data drift as a change in the joint distribution of features. That is:

\begin{equation}
P(x_t, y_t) \neq P(x_u, y_u)
\end{equation}

\noindent Probably approximately correct learning relies on the assumption that data are independently and identically distributed to estimate the empirical risk of a learning function. If we observe data drift, we cannot guarantee that the empirical risk is close to the true risk.

We can decompose $P(x,y) = P(x) \times P(y|x)$. Thus, if data drift occurs, it may stem from two sources: a change in $P(x)$ (covariate drift) or a change in $P(y|x)$ (concept drift). As stated by \citet{MORENOTORRES2012521}, covariate drift is tied to the distribution of the variables, while concept drift implies that the relationship between the target and predictors changes between datasets. Finally, both $P(x)$ and $P(y|x)$ may differ significantly from the original distributions, which is defined as dual drift. Overall, data drift can be stated as a phenomenon in which the statistical properties of a target domain change over time in an arbitrary way \cite{LU201411}.

In this work, we address data drift by building ensembles composed of models trained on independent feature subsets. We propose that if a small portion of features suffers from drift, accurate predictions can still be obtained from the unaffected features. Further, if the distributions $T$ and $U$ are completely different, we must be able to signal the low reliability of the prediction. Finally, to ensure diversity in constituent models, we rely on the notion of explainability.

\vspace{0.1in}
\noindent\textbf{Model Explanation:} Instead of the model reliance metric proposed by Fisher et al., another possible approach is presented by \citet{shap}. Shapley Additive Explanations (or simply SHAP) use Shapley values to interpret a prediction model. We represent how model $f'$ explains the data as a $d$-dimensional vector $S(f') = {s_1, s_2, \ldots , s_d }$, showing which features contribute most to the prediction. The Shapley value is a concept in cooperative game theory introduced by \citet{shapley1953value}. In each game, a unique distribution of the rewards generated by the cooperation of all players is provided. Many other feature attribution methods exist \cite{reason:BreFriOlsSto84a, saabas2014interpreting, lime2}, but as highlighted by \citet{hinns2021initial}, the sound mathematical foundation and ease of implementation make SHAP ideal for identifying underspecification. Further, SHAP is the only method with the three desirable properties:

\begin{itemize}
\item Local accuracy: the explanations truthfully explain the model.
\item Missingness: missing features have no attributed impact on the decisions.
\item Consistency: if a model changes so that some feature's contribution increases or stays the same regardless of the other features, that feature's attribution should not decrease.
\end{itemize}

In summary, the Rashomon Set reveals the existence of multiple valid model solutions with similar performance. Their explanations allow us to examine each model's decision rationale and understand the importance of different features, highlighting what makes each model in the Rashomon Set distinct. Further, it enables us to discern the unique aspects of each model that contribute to varied performance and responses under data drift. By combining these approaches, we gain a broader understanding of the problem and build ensembles of diverse models that provide complementary explanations for different facets of the data. This ensemble enhances the robustness of our solution, as each model's behavior under varying conditions is better understood and accounted for.

\section{Related Work}

Data drift is usually associated with the notion of online learning, in which a model is applied to production and is constantly updated as new instances arrive. Under online learning, a model must handle new concepts as they arrive, properly tuning itself to new data distributions. The main challenge consists of the fact that, as data drifts toward these new concepts, it negatively impacts the accuracy of the models that are learned based on past training instances \citep{GONCALVES20148144}. Therefore, early identification and adaptation to data drift are key aspects of such systems. \citet{learningconceptdrift2019} provides a basic framework underlying general drift detection:

\begin{itemize}
\item Stage 1 (Data Retrieval): retrieval of chunks from data streams to infer data distribution.
\item Stage 2 (Data Modeling): extraction of key features that present the most impact on the system in the presence of drift.
\item Stage 3 (Test Statistics Calculation): the measurement of a dissimilarity or distance metric.
\item Stage 4 (Hypothesis Test): evaluation of the statistical significance of the measured metric.
\end{itemize}

The main differences between methods lie in stages 3 and 4. Concerning stage 3, two of the main categories of drift identification are error-based and data-based algorithms. Most error-based drift detection employs a base classifier and tracks the change in the online error rate. The main hypothesis behind these methods relies on the fact that the base model will misclassify new instances when data drifts, thus increasing the error rate. This is the core idea behind the DDD of \citet{minku2011ddd}. There are many other error-based methods, but, as stated by \citet{learningconceptdrift2019}, DDD is perhaps the most referenced method. Under their framework, other methods can be summarized by changes to some stage of the drift detection, such as employing another hypothesis testing \citep{frias2014online} or changing some detail of the evaluated metric \citep{baena2006early}.

Data-based drift detection algorithms rely on directly quantifying the dissimilarity\linebreak between the distribution of historical and new data. The standard strategy is to define a fixed window for the past and a sliding window for new data during the online learning process \citep{kifer2004detecting}. If we ignore Stage 1 of the drift detection framework, the problem turns into a multivariate two-sample test evaluating if samples come from the same distribution. However, there remains a problem concerning actual and virtual drift.

The decomposition of Equation 3 presents the sources of data drift: covariate drift and concept drift. Covariate drift is often called virtual drift due to drift in $P(x)$ not affecting the decision boundary of models \citep{ramirez2017survey}. Retraining a model under covariate drift might not be necessary, as the learned conditional $P(y|x)$ remains unchanged. This is not the case for dual drift, however, when both $P(x)$ and $P(y|x)$ exhibit a shift under new data. It is important to highlight that the aforementioned approaches to drift detection are well-suited for online learning scenarios, which is not the case for our proposed problem. We can only compute error-based metrics if we know the correct label of new incoming instances. Sliding window data-based methods depend on the notion of temporal relationships. Further, knowledge of the labels of novel instances is necessary to differentiate between dual and virtual drift, which might not be possible in scenarios outside of online learning. We propose building an ensemble of models and using the intra-constituent agreement as a proxy for error rate, as described in Section \ref{sec:method}.

Existing research has explored methods to address the challenge of data distribution shifts. Domain adaptation, for instance, aims at mitigating performance degradation when a model trained on a source domain is applied to a different but related target domain. Farahani et al. categorize these approaches into shallow and deep methods, emphasizing strategies such as feature alignment, instance re-weighting, and adversarial training for settings where only source labels are available \cite{farahani2021brief}. These methods often focus on aligning feature distributions or adapting the model parameters to the target data \cite{pan2010domain, tzeng2014deep, long2015learning}.

Another related approach that tackles this problem is domain generalization, in which one seeks to train models that can generalize well to unseen target domains without access to this target data during training \cite{muandet2013domain}. The current approaches can be organized into categories such as data manipulation, representation learning, and learning strategies~\cite{wang2022generalizing}. For example, Mixup-based augmentation \cite{zhang2018mixup} enhances diversity by modifying training data through linear interpolation, while domain-adversarial training \cite{ganin2016domain} explicitly aligns distributions across domains by adversarial optimization. %In contrast to these methods, which focus on altering inputs or features to improve generalization, our approach instead detects distribution shifts by analyzing disagreement among diverse models in the Rashomon Set, bypassing the need for domain-specific adaptation or augmented training data.

A third approach consists of test-time adaptation methods. Unlike domain adaptation or generalization, which operate during training, this approach adapts pre-trained models directly to unlabeled test data in real time~\cite{liang2025comprehensive} and, as highlighted by Liang et al., they broadly fall into three categories: (i) test-time domain adaptation, which leverages pseudo-labeling or clustering to align entire target domains with source knowledge \cite{liang2020we, qu2022bmd}; (ii) test-time batch adaptation, which adjusts normalization statistics or fine-tunes parameters on small batches \cite{schneider2020improving, zhang2022memo}; and (iii) online test-time adaptation, which incrementally updates models on streaming data while attempting to mitigate catastrophic forgetting \cite{wang2022continual, niu2022eata}. Other approaches also include entropy minimization \cite{wang2021tent}, contrastive consistency \cite{chen2022contrastive}, or memory-augmented prototypes \cite{ding2023proxymix}. However, these methods assume that models must be adapted to the target distribution, requiring access to source model parameters and computational resources for updates. These may be prohibitively limiting in scenarios where model stability, interpretability, or deployment efficiency are crucial.

While these existing methods, be they domain or test-time, focus on adapting models to shifted data, our approach leverages the inherent diversity of high-performing models in the Rashomon Set. Instead of modifying parameters or normalizing statistics, we detect data drift through disagreement among ensemble constituents. Models in the Rashomon Set, though equally accurate on training data, rely on distinct features and hold different decision boundaries. Significant prediction divergence on new data hints at distribution shifts. This process requires no model updates, source data access, or computational overhead after deployment, providing a lightweight, proactive indicator for safety-critical or resource-limited settings, as is the case for many real-world applications.

One of the core motivations in this work arises from the insight that a dataset might be heterogeneous, thus inducing a large Rashomon Set. There might exist regions of the data that show complex correlations among a specific set of features and the target label, and the same correlations are not necessarily so strongly observed in other regions. If this is true, it would be more suitable if local behavior were represented by a local model, which can be incorporated into an ensemble~\citep{zuinaperam}. 
Sampling multiple local minima allows approximation of the global objective while expanding the representation space~\citep{dietterich2000ensemble}. This idea of employing local models aligns with the Rashomon Set concept, as it acknowledges the existence of multiple valid and diverse models that perform well in different regions of the data space. By exploring the Rashomon Set and considering models with contrasting explanations, we can identify subgroups of correlated features and build ensembles with diverse models that contribute unique explanations for different facets of the data~\cite{zuinnature}. Such an approach enhances robustness by leveraging the multiplicity of high-performing models with diverse decision-making processes.

\citet{dembczynski2008general} focuses on understanding how one can learn a performant rule-based ensemble via boosting. Starting from the standard initial rule, they iteratively add new rules to obtain an ensemble that can cover most of the data. To validate their approach, they also define the concept of coverage through a $\phi(x)$, this being an arbitrary axis-parallel region in the attribute space. The diversity of constituents is measured solely by the coverage $\phi$ of each rule. As noted by \citet{damour2020underspecification}, two rules may have the same coverage but exhibit divergent behavior in practice. Thus, using some other metric associated with the inner mechanism of the model and not simply the observed response may be relevant, such as a vector representation of the explainability of a model.

\citet{grosskreutz2008cascaded} propose splitting dataset rows into subgroups given a set of restrictions over its columns, and apply this approach to an unsupervised problem. If the groups are large enough, the associated restrictions express some significant pattern in the data. Grosskreutz focuses on tasks where there is no target variable. However, one can employ an equivalent technique regardless of this fact, similar to \citet{malik2008classification} and \citet{knobbe2009building}. All these works operate primarily within the data space, looking for relevant patterns, clusters, or subgroups that induce diverse models. Our approach, in contrast, operates within the model space, finding different groups of explanations. The Rashomon groups can be interpreted as a particular set of restrictions on the data, which in turn induce the subgroups presented. We improve upon previous work in the sense that the SHAP groupings aided by the Rashomon concept not only prune a large portion of the search space but also provide a direct measure of model behavior similarity while tackling the problem of data drift detection in domains outside of online learning.

Regarding the Rashomon Effect, many works have exploited its implications to gain insights about the solution space. \citet{predictivemultiplicity} explores the concept of predictive multiplicity, the ability of a prediction problem to admit competing models with conflicting predictions, which can be seen as a restriction on the Rashomon Set. \citet{kissel2021forward} searches for an entire collection of plausible models via a forward selection approach and resampling of the training dataset to account for uncertainty. \citet{dong2020exploring} introduces the notion of a variable importance cloud, mapping every variable to its importance for the Rashomon Set, and experimenting on criminal justice, marketing data, and image classification tasks. \cite{NING2022100452} performs a similar approach using Shapley values as a measure of importance. There is also relevant literature regarding Rashomon Sets and a specific learning algorithm of choice. For instance, \citet{ahanor2022diversitree} and \citet{danna2007generating} both look for the set of near-optimal solutions for integer linear programs, while \citet{xin2022exploring} restricts their analysis of the Rashomon Set to Decision Trees. To the best of the authors’ knowledge, building an ensemble from the Rashomon set is a novel idea.
\section{Method}
\label{sec:method}
We consider a supervised learning scenario and formulate a classification model as a function $h(X, Y; \theta)$ parameterized by $\theta$ that maps inputs $x_i \in X$ to labels $y_i \in Y$. During cross-validation, we train models on data $D_{train}$ coming from a distribution $T$. To estimate the predictive risk of each function, we employ additional data $D_{test}$ from the same distribution $T$ and evaluate $h_n \in H$ on this independent and identically distributed data. The standard model selection step involves selecting the function that minimizes the empirical predictive risk, providing performance guarantees when future data follows the same distribution $T$. However, these guarantees do not hold when dealing with data coming from other distributions, such as in the case of data drift.

Our main objective is to build a diverse ensemble comprising contrasting explanations for the same problem. Additionally, we aim to estimate the reliability of our predictions under uncertainty arising from an unknown data distribution $U$, which may contain drift compared to the training data distribution $T$. To achieve this, we explore how models behave when the differences between executions are only minor. We consider $\theta$ to encompass any choices made during training that lead to similar models exhibiting contrasting performances. We then introduce drift to the test data and evaluate its effects on each model.

Instead of simply mixing different structures into a single model and minimizing the objective function $h(x)$, we sample the model space by minimizing different functions $h(x')$, where $x' \subseteq x$ and $|x'| < |x|$, as in \citep{zuinportinari2020}. This sampling strategy approximates the Rashomon Set, acknowledging the existence of multiple valid and diverse models that perform well in different regions of the data space. By exploring the Rashomon Set and considering models with contrasting explanations, we can identify subgroups of correlated features and build ensembles with diverse models that contribute unique explanations for different facets of the data. This approach enhances the robustness of our solution by considering the multiplicity of well-performing models with varying decision-making processes.

We build our ensemble exploiting two concepts: diversity between individual models and stability between model explanation and empirical predictions~\citep{galit}. Diversity is crucial for gaining a general understanding of a phenomenon, assuming that problems are not tied to a single cause, which may vary in ways that are not directly intuitive. To promote diversity while finding patterns, we cluster the set of sampled models $H'$ based on the distance between their explanation vectors (i.e., SHAP values). In our experiments, we employ Euclidean distance and k-means clustering, though alternative metrics and clustering methods could be applied. Ideally, this creates groups of models that are internally dense and separated from other models in terms of their explanatory factors. Stability, on the other hand, refers to models within a cluster being associated with the same explanatory factors and performing similar predictions. 

To assess prediction-explanation stability, we cluster the model space based on the distance between the explanation vector associated with each model and project them into the prediction space. This allows us to locate different Rashomon subgroups inside the Rashomon Set and select models from each subspace. In practice, when we evaluate one constituent at a time, the remaining members of the ensemble act as reference models to verify consistency under new data distributions. If a candidate model’s predictions agree with the remainder of the ensemble, it is indicative of prediction stability. Further, we also require a stability check to ensure that, when searching for optimal constituents, adding or removing features does not significantly alter the model's explanation vector relative to its cluster centroid and neighborhood. Finally, to study the Rashomon Set for a given problem, we need to sample models from the complete model space. Algorithm~\ref{alg:mrce} describes the main steps of our ensemble learning approach. Here each model $h$ is represented by the unique feature set that it employs. As such, we overload the notation $h$ to denote both the model and its feature set, using them interchangeably in the algorithm description.

\begin{algorithm}[ht]
\caption{Rashomon Ensemble Algorithm}
\label{alg:mrce}
\KwIn{Feature set \(F\), evaluation dataset \(Z\), number of initial models \(n\), maximum model width \(m\), error margin \(\epsilon\)}
\KwOut{Ensemble of models \(M\)}
% [CHANGED] Initialization
Initialize pool \(P\) with \(n\) models, each built using a random subset of features from \(F\).\\
Choose a reference model \(h_{\text{ref}}\) (e.g., a baseline method).\\
Initialize an empty Rashomon Set \(R\).\\[1mm]

\For{each model \(h_i \in P\)}{
    Evaluate \(h_i\) on \(Z\).\\
    \If{\(E[L(h_i, Z)] \leq E[L(h_{\text{ref}}, Z)] + \epsilon\)}{
        Compute explanation vector \(\text{S}(h_i)\) (e.g., using SHAP).\\
        Add \(h_i\) along with \(\text{S}(h_i)\) to \(R\).\\[1mm]
        }
    }
Cluster the models in \(R\) based on the distance between their explanation vectors, forming clusters \(C\).\\
For each cluster \(c \in C\), select a representative model (e.g., the clusteroid) \(h_c\).\\[1mm]

Initialize the ensemble \(M\) with these representative models.\\
\For{each cluster \(c \in C\)}{
    Let \(h_c\) be the representative model for cluster \(c\).\\
    \While{\(|h_c| < m\)}{
        Identify the feature \(f \in F \setminus h_c\) that minimizes 
        \[
        E\left[L\left((M \setminus \{h_c\}) \cup \{h_c \cup \{f\}\}, Z\right)\right]
        \]
        while ensuring that \(h_c \cup \{f\}\) remains consistent with the explanatory profile of cluster \(c\) (e.g., does not fall into another cluster).\\
        Update \(h_c \leftarrow h_c \cup \{f\}\).\\
    }
}
\KwRet{\(M\)}
\end{algorithm}

\vspace{0.1in}
\noindent\textbf{Deriving an Ensemble:} We assume a factorial combinatorial space encompassed by all feature combinations constrained to a single learning algorithm. To induce the Rashomon Set, we aim to find a set of relevant features $K$ (with size $|K|$) that characterize an evaluated subspace. These features show complex correlations with the target label, which may not appear as strongly in other regions of the data space, thus inducing a Rashomon subspace.

To mitigate the curse of dimensionality, we restrict the model space to subsets of size $s$ where $|K| \leq s \leq S_{\text{max}}$, with $S_{\text{max}} \ll |F|$. This avoids the computational intractability of enumerating all $2^{|F|}$ subsets. The number of models containing the subset $K$ is derived by fixing $|K|$ features and choosing the remaining $s - |K|$ from $|F| - |K|$:

\begin{equation}
\label{EQUATION 5}
\sum_{s=|K|}^{S_{\text{max}}} \binom{|F| - |K|}{s - |K|}
\end{equation}

\noindent The probability of sampling a model containing $K$ is:

\begin{equation}
\label{EQUATION 6}
P_K = \frac{\sum_{s=|K|}^{S_{\text{max}}} \binom{|F| - |K|}{s - |K|}}{\sum_{s=|K|}^{S_{\text{max}}} \binom{|F|}{s}}
\end{equation}

We limit our scope to problems where $|K| \ll |F|$, as otherwise, the Rashomon ratio diminishes due to the combinatorial scarcity of subspaces containing $K$. To ensure computational tractability, we restrict models to sizes $s$ where $|K| \leq s \leq S_{\text{max}}$, avoiding the curse of dimensionality inherent in high-dimensional feature spaces. If we sample an arbitrary model from this constrained space, the probability of it \textit{not} containing $K$ is $1 - P_K$. From Equation \ref{EQUATION 6}, to guarantee that the subspace $K$ is present in at least one model with probability $\alpha$, we need to sample at least $\eta$ models:

\begin{equation}
\label{eq:sample}
\eta = \frac{\ln(1-\alpha)}{\ln\left(1 - P_K\right)}
\end{equation}

\noindent For example, for $|F| = 100$, $|K| = 4$, $S_{\text{max}} = 10$, and $\alpha = 0.95$:
\[
\eta \approx \frac{\ln(1 - 0.95)}{\ln\left(1 - 0.00005\right)} \approx 60{,}000
\]

\vspace{0.1in}
\noindent\textbf{Time Complexity:} In the experiments, we use Decision Trees as base models for the ensemble constituents. The time complexity of training and explaining a Decision Tree is $O(\log(F)ID + D^2)$~\cite{shap}, where $I$ is the number of instances, $F$ is the number of features, and $D$ is the maximum tree depth. Since we sample $T$ trees, other steps present negligible complexity in comparison to the sampling stage, resulting in $O(TI)$ complexity. However, as we sample models, the Rashomon ensemble substantially reduces the number of models that need to be evaluated, making the approach feasible in practice. For instance, any model with a loss close to random guessing is unlikely to present itself as a useful constituent. Thus, we do not need to explain the entire model space, and need only concern ourselves with the Rashomon Set.

\vspace{0.1in}
\noindent\textbf{Splitting the Rashomon Set:} To split the Rashomon Set into clusters, we represent how a model $h'$ explains a phenomenon as a d-dimensional vector $S(h') = [e_1; e_2; ... ; e_d]$ showing which features $[x_1, x_2, ... x_d]$ drive the model's prediction. We use K-Means clustering with a suitable number of clusters, identified by maximizing the silhouette value. This splits the Rashomon Set into well-divided clusters based on their explanatory factors, leading to compact and well-separated clusters. As discussed previously, there usually exists a small subset of key features that are only present in models from one cluster and absent in the remaining ones. The presence of this subset leads to these models being close in the feature preference space since cohesion values are relatively high and lead to concise and well-divided clusters.

\vspace{0.1in}
\noindent\textbf{Prediction Distance:} We compare models within the Rashomon Set to estimate the risk under an unknown distribution $U$. We compute the Jensen-Shannon distance (JSD) \cite{jensenshannondist} as our metric of choice for a measure of risk, indicating how similar the predictions of the two models are. Let $P$ be the probability distributions returned from a model $h_p$, and we wish to compute a metric that estimates the risk of selecting it in production. Further, let $Q$ be the probability distribution from a model $f_q$ that ideally behaves similarly to $h_p$. As shown by \citet{mackay2003information}, we can compute the error of $P$ from $Q$ by the cross-entropy between $P$ and $Q$ as:

 \begin{equation} H(P, Q) = H(P) + D_{KL}(P || Q) \end{equation}

We could also evaluate the Kullback-Leibler (KL) divergence between these models under the unknown distribution $U$, and thus estimate risk. The main drawback of employing the Kullback-Leibler divergence is that it is non-symmetric. That is, $D_{KL}(P, Q)$ might be different from $D_{KL}(Q, P)$. To avoid confusion, we instead opt to employ the Jensen-Shannon distance as our metric of choice for a measure of risk. If $P$ and $Q$ agree (low JSD), we have a strong indicator that $U$ should be similar to $T$, and predictions can be trusted. Contrasting $P$ and $Q$ (high JSD) suggests that $U$ differs from $T$, and the returned predictions cannot be trusted.

\vspace{0.1in}
\noindent \textbf{Constituent search:} In summary, we verify that looking at the explanatory factors in isolation is not enough to observe meaningful patterns. In our preliminary experiments, we find instances of models with similar SHAP but contrasting predictions as well as contrasting SHAP but similar predictions. The choice of a $h_q$ model to estimate the risk of the target constituents becomes a challenging task. We propose performing a controlled transformation in $T$ to create a simulated production dataset. This should enable us to estimate model behavior in an out-of-distribution scenario. Namely, the transformation employed over data drawn from $T$ consists of adding Gaussian noise to the input features such that $y_i = h(x_i + \epsilon_i)$ and $\epsilon_i \sim N(0, \sigma^2)$. We can then select models that have contrasting explanations and predictions. Among possible transformations, we choose Gaussian noise for its simplicity and the ease of computing exact feature distortion. In many real-world scenarios, Gaussian noise might not be the closest representative of divergence. However, we verify that this simple transformation is enough to induce large changes in model behavior and enable our ensemble learning approach.

Further, not all variables are relevant for prediction, and some features may even be detrimental. To find a set of relevant features to induce the Rashomon Set, we represent the model space as a directed acyclic graph (DAG) in which each node represents a distinct feature subset, and vertex $A\rightarrow B$ is connected if $B$ can be reached by simple feature addition from $A$, thus representing a transitive reduction of the more complex combinatorial complete model space. This modeling approach has two desirable properties: (i) any vertex is reachable from the $[\emptyset]$ model, and (ii) a topological ordering exists such that for every edge, the start vertex occurs earlier in the sequence than the ending vertex of the edge for any feature set path. These properties imply a partial ordering of the graph starting from the root node, which allows us to search it in an orderly manner. It has been shown that this modeling approach is effective for the task at hand~\citep{zuinaperam, zuinnature}.

We can, for example, apply the A* algorithm ~\citep{astar}, employing as a heuristic the performance of the model represented by the feature set of a given vertex and the Jensen-Shannon distance to the predictions of the remaining Rashomon subgroup clusteroids. We hypothesize that there exists a set of optimal feature expansions that lead to the best-performing models for each specific base task. This allows us to search the $F!$ combinatorial space of feature subsets to select the best-performing specialized models and build the Rashomon ensemble.

\section{Open datasets}% from the Rashomon Set}
\label{sec:ensemblerashomon}

We present our experiments related to the Rashomon Set for a given problem and the process of obtaining ensemble constituents using the Rashomon Sets. The goal is to explore the usefulness of Rashomon Sets as a method for model space partitioning and to understand their effectiveness in addressing the problem akin to underspecification in ensembles. To study the Rashomon Set for a given problem, we sampled models from the complete model space. We considered the $N!$ combinatorial space, encompassing all feature combinations constrained to a single learning algorithm. We aimed to evaluate whether Rashomon Sets could serve as a valuable tool for partitioning the model space and generating diverse ensembles.

To achieve this, we proposed a process of Rashomon Set partitioning based on clustering models by their explainability vectors. The K-Means algorithm was used to induce clusters, and we determined the optimal number of clusters (K) using silhouette scores. By creating ensembles composed solely of models located close to the centers of each Rashomon subgroup, we aimed to generate diverse ensembles capable of covering a wider region of the solution space.

\subsection{Benchmark Suite and Datasets}

To verify the effectiveness of the Rashomon ensemble learning technique, we considered a benchmark suite including a series of open-source datasets from the UCI machine learning repository \citep{asuncion2007uci} and the OpenML database \citep{bischl2017openml}. The benchmark suite consists of the following datasets:

\vspace{0.1in} \noindent\textbf{APS Failure:} The dataset used for the 2016 IDA Industrial Challenge \citep{apsfailure}. It consists of data collected from heavy Scania trucks in everyday usage, and the problem is formulated as a binary classification task to predict component failures for a specific component of the APS system after a small amount of noise was introduced to the data.

\vspace{0.1in} \noindent\textbf{Diabetes Readmission:} This dataset was submitted on behalf of the Center for Clinical and Translational Research, Virginia Commonwealth University \citep{diabetes}. It represents 10 years of clinical care at 130 US hospitals and integrated delivery networks. The problem is a binary classification task to predict whether a given patient will be readmitted to a hospital.

\vspace{0.1in} \noindent\textbf{Heart Disease:} This dataset from the Cleveland database focuses on the diagnosis of coronary artery disease \citep{heart}. The goal is to predict the presence of heart disease in the patient, with a severity indicator valued from 0 (no presence) to 4. We have focused on the binary counterpart of this problem, in which we simply attempt to distinguish presence (value 1, 2, 3, 4) from absence (value 0).

\vspace{0.1in} \noindent\textbf{MADELON:} This artificial dataset contains data points grouped in 32 clusters placed on the vertices of a five-dimensional hypercube and randomly labeled +1 or -1, and it was one of five datasets used in the NIPS 2003 feature selection challenge \cite{mandelon}. The problem is a binary classification task to separate examples into two classes.

\vspace{0.1in} \noindent\textbf{MAGIC:} This dataset is composed of a series of Monte Carlo simulations regarding the registration of high-energy gamma particles in a ground-based atmospheric Cherenkov gamma telescope (Major Atmospheric Gamma Imaging Cherenkov Telescope project, MAGIC) \cite{magic}. The problem is a binary classification task to discriminate the patterns caused by primary gammas (signal) from the images of hadronic showers initiated by cosmic rays in the upper atmosphere (background).

\vspace{0.1in} \noindent\textbf{Nursery:} This dataset was derived from a hierarchical decision model originally developed to rank applications for nursery schools, thus constituting the Nursery Database \citep{nursery}. The goal is to predict the final decision, ranging from not recommended to priority. We have focused on the binary counterpart of this problem, in which an applicant was given either a priority recommendation or not.

\vspace{0.1in} \noindent\textbf{Speed Dating:} This dataset was gathered from participants in experimental speed dating events from 2002 to 2004 \citep{speeddating}. The problem was formulated as a binary classification task to predict whether both participants would like to date each other again, given each participant's questionnaire responses and characteristics.

\vspace{0.1in} \noindent\textbf{WDBC:} This dataset is composed of features computed from a digitized image of a fine needle aspirate (FNA) of a breast mass associated with breast cancer \citep{wdbc}. The problem was formulated as a binary classification task to predict the presence of malignant tumor cells.

\vspace{0.1in} \noindent\textbf{Wine Quality:} This dataset is composed of chemical analysis of wines grown in the same region in Italy but derived from three different cultivars \citep{wine}. The analysis determined the quantities of 13 constituents found in each of the three types of wines, and the end goal is to predict the wine quality score, ranging from 0.0 to 8.0. We have focused on the binary counterpart of this problem, in which we wish to predict whether a given wine is of high quality (>5) or not.

\subsection{Rashomon Ensemble Learning}

Table \ref{tab:benchmark} summarizes our comparison between our approach and classic and state-of-the-art algorithms. Specifically, we employ AdaBoost~\cite{adaboost}, Random Forests~\cite{randomforest}, XGBoost~\cite{chen2016xgboost}, LightGBM~\cite{lightgbm}, Catboost~\cite{catboost2018}, and GRANDE~\cite{grande} as baseline algorithms. We also consider an evolutionary approach using the NSGA-II algorithm~\cite{nsgaii, gupta2022nsga} and evaluating the same number of models as our Rashomon Ensemble. The proposed evolutionary framework optimizes both the performance and feature diversity of the population and induces an ensemble from the best individuals of the last generation. For a fair comparison to the ensemble and boosting methods, we only employed decision trees as base constituents for our Rashomon ensembles. In our experiments, we sampled $100,000$ decision trees to guarantee a minimum subset diversity and trained a meta-model to combine constituent outputs in a stacking ensemble. No hyper-parameter tuning was employed, either on the baselines or the Rashomon ensembles, to ensure a fair comparison. The number of trees in all ensemble algorithms was limited to 50.

\begin{table}
\centering
\small
\setlength\tabcolsep{1.5pt}
\caption{Mean AUROC results on binary classification tasks after 10 repetitions. Datasets with no pre-defined test set were subject to an 80-20 cross-validation split.}
\label{tab:benchmark}
\small
\begin{tabular}{lcccccccccccc}
\multicolumn{3}{c|}{Benchmark} & \multicolumn{8}{c|}{Baseline Algorithm} & \multicolumn{2}{c}{Rashomon} \\ %\hline \\
%& & & & & & & & & & \\
Dataset & Rows & \multicolumn{1}{l|}{Cols} & DT & Ada & RF & XGB & LGBM & CatBoost & {GRANDE} & \multicolumn{1}{c|}{NSGA-II} & Ensemble & Ratio \\
\hline \\
APS & 76000 & \multicolumn{1}{l|}{172} & .866 & .824 & .869 & .835 & .853 & .888 & .855 & \multicolumn{1}{c|}{.859} & \textbf{.911} & $12.4\%$ \\
Diabetes & 101766 & \multicolumn{1}{l|}{1691} & .544 & .614 & .599 & .615 & .616 & \textbf{.619} & - & \multicolumn{1}{c|}{\textbf{.619}} & \textbf{.618} & $17.4\%$ \\
Heart & 303 & \multicolumn{1}{l|}{171} & .748 & .787 & .826 & .796 & .830 & .834 & .825 & \multicolumn{1}{c|}{.806} & \textbf{.839} & $50.3\%$ \\
Nursery & 12630 & \multicolumn{1}{l|}{784} & \textbf{.999} & \textbf{.999} & \textbf{.999} & .991 & \textbf{.999}  & \textbf{.999} & \textbf{.999} & \multicolumn{1}{c|}{\textbf{.999}} & \textbf{.999} & $83.2\%$ \\
WDBC & 569 & \multicolumn{1}{l|}{903} & .949 & \textbf{.973} & .967 & .963 & .967  & \textbf{.974} & \textbf{.975} & \multicolumn{1}{c|}{\textbf{.973}} & \textbf{.974} & $21.5\%$ \\
Wine & 4898 & \multicolumn{1}{l|}{13} & .762 & .722 & .802 & .755 & .764 & .782 & .802 & \multicolumn{1}{c|}{\textbf{.804}} & \textbf{.805} & $8.9\%$ \\ 
MAGIC & 19020 & \multicolumn{1}{l|}{102} & .808 & .830 & .857 & .837 & .850 & .850 & \textbf{.897} & \multicolumn{1}{c|}{.809} & .848 & $19.4\%$ \\ \hline \\
MADELON & 2000 & \multicolumn{1}{l|}{502} & .764 & .598 & .694 & .828 & .832 & \textbf{.852} & .594 & \multicolumn{1}{c|}{.674} & \emph{.746} & \emph{$<0.5\%$} \\
Speeddating & 8378 & \multicolumn{1}{l|}{123} & .650 & .673 & .630 & .639 & .642 & .668 & \textbf{.801} & \multicolumn{1}{c|}{.751} & \emph{.632} & \emph{$<0.5\%$} \\ \hline
\end{tabular}
\end{table}

\begin{figure}[!t]
\centering
\begin{subfigure}{0.48\textwidth}
\centering
\includegraphics[width=\linewidth]{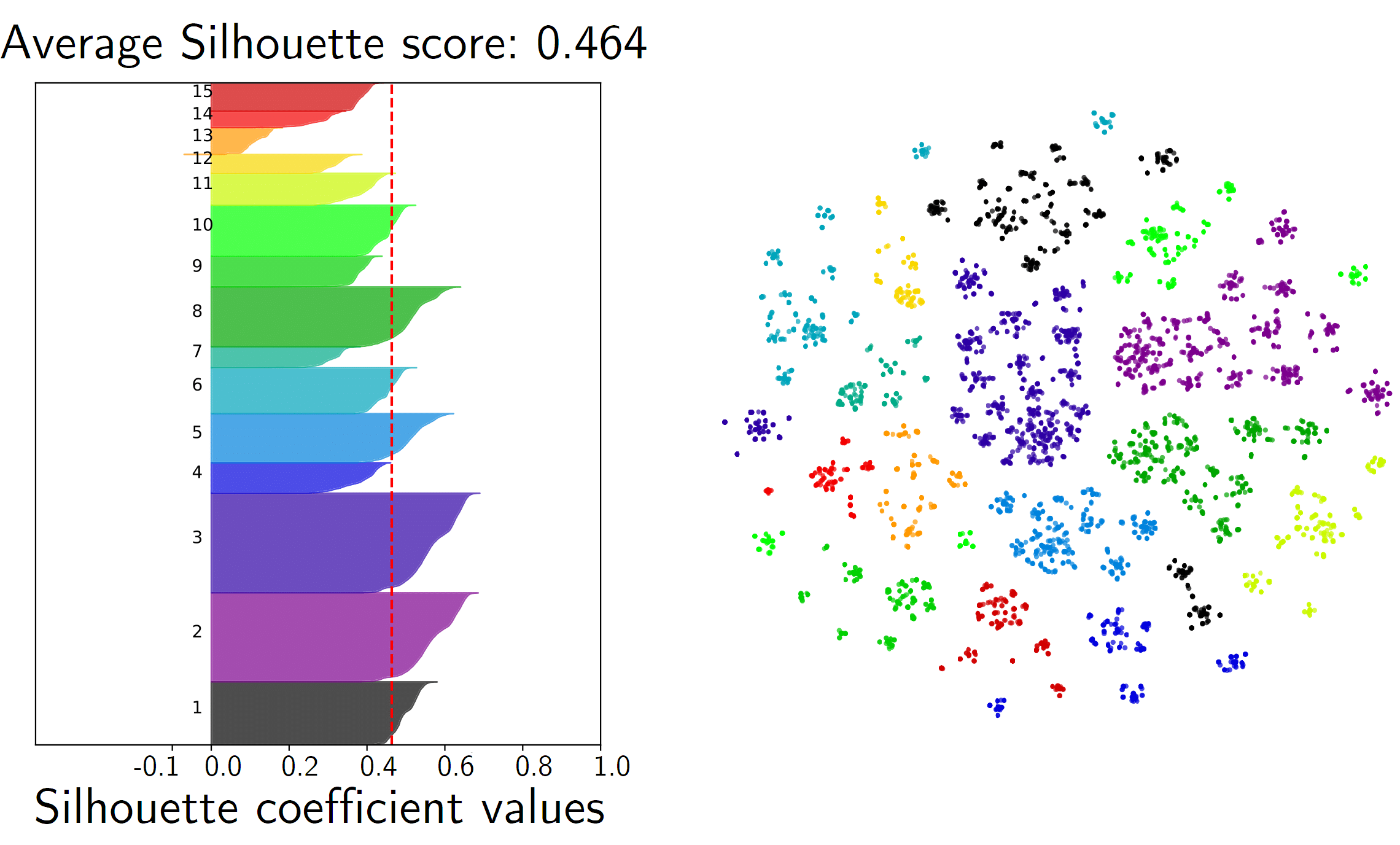}
\caption{Optimal k (15) for MAGIC models.}
\label{fig:kmeans}
\end{subfigure}
\begin{subfigure}{0.48\textwidth}
\centering
\includegraphics[width=\linewidth]{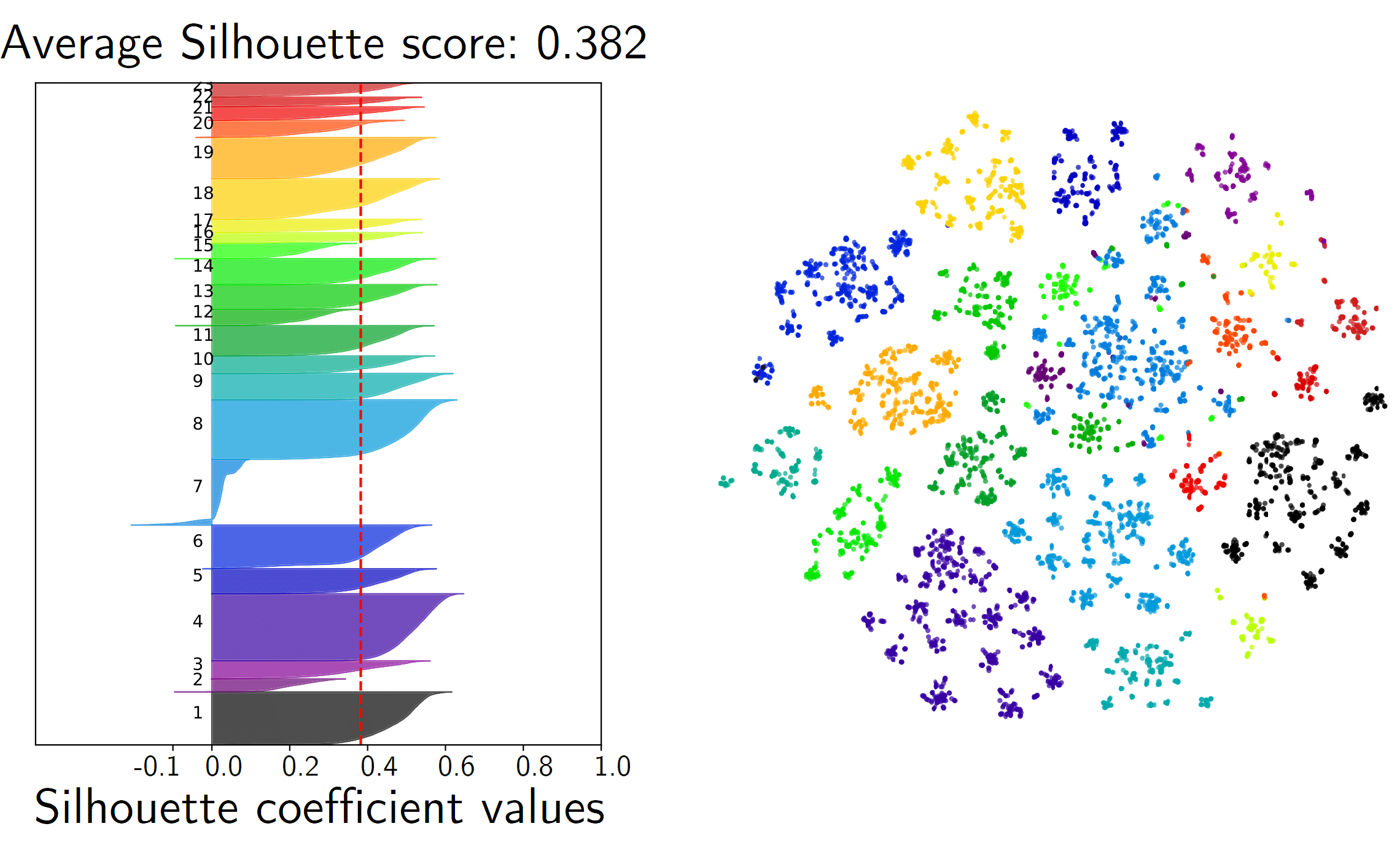}
\caption{Optimal k (23) for APS Failure models.}
\label{fig:kmeanswdbc}
\end{subfigure}
\caption{TSNE reduction of the Rashomon space and optimal silhouette for each subgroup.}
\end{figure}

In the Nursery dataset, we verify that nearly all models lie inside the Rashomon space. This implies that the problem is relatively easy, and nearly any model is performant. In this scenario, the choice of using Rashomon ensembles or any other learning algorithm becomes less meaningful, and we observe that all baselines can achieve an AUROC of 0.99. In other scenarios, where the Rashomon ratio is large but not excessive (between 8\% and 50\%), we observe statistically significant gains when using our approach. The poor performance on the Speed Dating and MADELON datasets can be explained by the scarcity of contrasting explanations, represented by the small size of the Rashomon Set. The same cannot be said of the MAGIC dataset. One hypothesis for this behavior is related to MAGIC being a purely synthetic dataset. There might be some underlying pattern guiding the feature creation that is not present in the remainder datasets, which were crafted from different real-world problems. Figures \ref{fig:kmeans} and \ref{fig:kmeanswdbc} illustrate some Rashomon subspaces and their respective Rashomon partitions after sampling. The figures show a TSNE reduction of the Rashomon space and the optimal silhouette scores for each subgroup.

\subsection{Evaluating Ensemble Composition and Robustness}

\begin{figure}[htb]
\centering
\begin{subfigure}{0.48\textwidth}
\centering
\includegraphics[width=\textwidth]{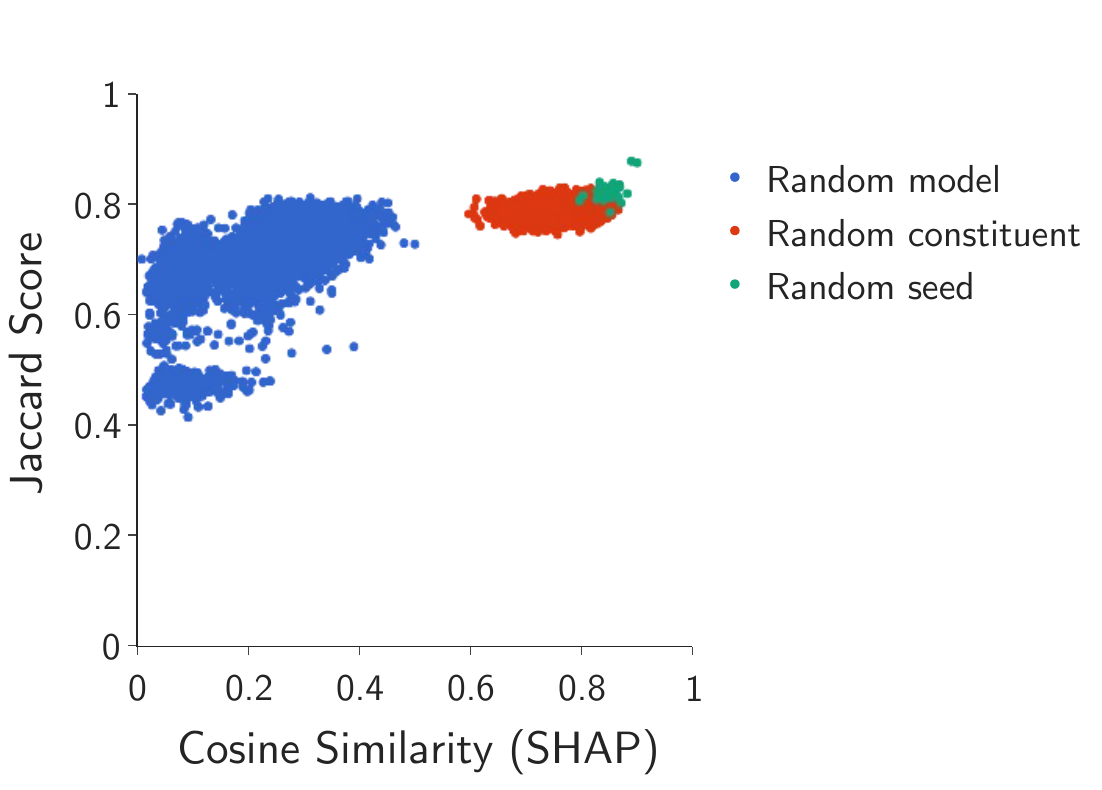}
\caption{Rashomon ensembles' sensibility to random constituent or seed selection.}
\label{MAGIC}
\end{subfigure}
\hspace*{0.02\linewidth}
\begin{subfigure}{0.48\textwidth}
\centering
\includegraphics[width=\textwidth]{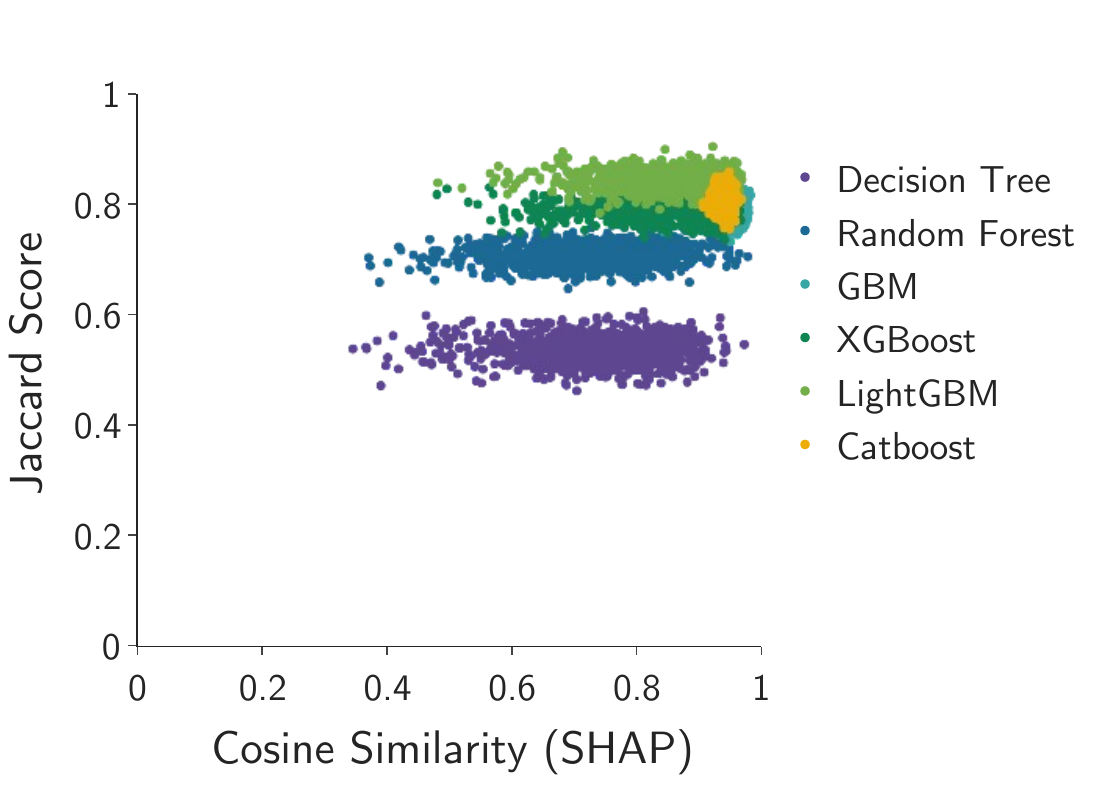}
\caption{Remaining baseline methods' sensibility to random seed selection.}
\label{MAGICOTHERS}
\end{subfigure}
\caption{Similarity to a reference model found from running the Rashomon and baseline pipelines, filtering models with statistically worse performance than the proposed threshold. MAGIC dataset.}
\label{fig:underexp}
\end{figure}

To assess the ensemble composition and robustness, we considered three scenarios to isolate the impact of key components in our Rashomon pipeline (Figure \ref{fig:underexp}). Scenario I (green points) evaluates the stability of the full proposed method, in which we ran the complete pipeline 30 times with different random seeds to test sensitivity to stochasticity. Scenario II (red points) serves as a cluster-aware ablation, in which we retained the Rashomon subgroups but selected constituents randomly from each, over a total of 10,000 runs. This scenario represents omitting the intra-cluster optimization step. Finally, in Scenario III (blue points), we conduct a subgroup-free ablation. We ignored clustering entirely and selected models randomly from the entire Rashomon Set over 10,000 runs.

In Figure \ref{MAGIC}, Scenario I (green) demonstrates stable similarity to the reference model across seeds, while Scenarios II (red) and III (blue) illustrate how omitting optimization or clustering increases variability. Figure \ref{MAGICOTHERS} presents a comparison against the remaining baseline methods employed in our previous experiments.

To compare these different scenarios, we used a visualization scheme that jointly considered the Jaccard Index and SHAP values' similarity between the models found in each scenario and a reference model (performance in Table \ref{tab:benchmark}). The Jaccard Index provided insights into the degree of agreement between two sets of predictions, accounting for differences in prediction patterns that might be overlooked by standard performance metrics. On the other hand, the SHAP values similarity helped gauge the robustness of the explanations and whether the explanatory factors remained consistent despite the stochastic nature of the algorithms.

Upon analyzing the results presented in Figure \ref{fig:underexp}, we observed that both Scenario I and Scenario II demonstrated models with high Jaccard Index and SHAP values similar to the reference model, indicating better ensemble compositions and increased robustness in terms of explanations. In contrast, Scenario III, which directly selected models from the whole Rashomon Set without considering subgroups, yielded models with statistically inferior performance compared to the reference model.

The outcomes of Scenarios I and II provide evidence that the Rashomon pipeline offers a viable solution to enhance ensemble robustness and credibility. By maintaining Rashomon subgroups and selecting ensemble constituents from each cluster, the Rashomon ensemble generation process appears to mitigate challenges related to ensemble underspecification, as discussed by \citet{damour2020underspecification}, which mentions that only observing the performance of models poses an ineffective way to judge underspecification and thus, the potential multiplicity and divergence of seemly equal models under production settings. We also observe that performing the intra-cluster optimization step, represented by the green cloud of points on Figure \ref{MAGIC}, severely reduces the variability on SHAP. That is, different runs of the algorithm seem to result in the same set of variables and key constituents being chosen. This same behavior cannot be verified for most of the remaining baseline methods in Figure \ref{MAGICOTHERS}, except for Catboost.

\subsection{Robustness to Distribution Drift}

To evaluate the robustness of Rashomon ensembles to distribution drift, we conducted experiments to address concerns about out-of-distribution data. We considered two scenarios: the addition of Gaussian noise with increasing values of ${\sigma}^2$ to simulate data drift and shuffling feature values to evaluate the reliance on core key features.

In the first scenario, we added Gaussian noise with increasing ${\sigma}^2$ values to the datasets, mimicking shifts in the data distribution. We then evaluated the performance of Rashomon ensembles and other models under these perturbations. The results of this data drift scenario are summarized in Table \ref{tab:underfing}, where each approach's performance is represented as a ring plot ordered by performance. The mean AUROC (Area Under the Receiver Operating Characteristic curve) after 30 repetitions is provided as a measure of performance.

In the second scenario, we shuffled the feature values within the datasets, disrupting the relationship between features and the target variable. This scenario aimed to evaluate whether models could extrapolate from global information rather than relying on specific local patterns. The results of this data shuffle scenario are also presented in Table \ref{tab:underfing}.

\definecolor{rashcolor}{rgb}{0.18, 0.08, 0.22}
\definecolor{catcolor}{rgb}{0.62, 0.23, 0.31}
\definecolor{lgbmcolor}{rgb}{0.82, 0.70, 0.62}
\definecolor{rfcolor}{rgb}{0.65, 0.75, 0.80}
\definecolor{gluoncolor}{rgb}{0.62, 0.44, 0.23}
\definecolor{grandecolor}{rgb}{0.28, 0.39, 0.38}

\begin{table}[b]
\centering
\setlength\tabcolsep{6pt}
\caption{Performance loss comparison between {\rule{7pt}{7pt}}, Random Forest \textcolor{rfcolor}{\rule{7pt}{7pt}}, LightGBM \textcolor{lgbmcolor}{\rule{7pt}{7pt}}, CatBoost \textcolor{catcolor}{\rule{7pt}{7pt}} and Rashomon ensembles \textcolor{rashcolor}{\rule{7pt}{7pt}}. Mean AUROC after 30 repetitions.}
\small{
\begin{tabular}{lcccccccccc}
%\hline
& \multicolumn{5}{c|}{Data Drift (${\sigma}^2$) } & \multicolumn{5}{c}{Data Shuffle (n)} \\
& $0.4$ & $0.8$ & $1.2$ & $1.6$ & \multicolumn{1}{l|}{$2.0$} & $10\%$ & $30\%$ & $50\%$ & $70\%$ & $90\%$ \\ \hline

& & & & & & & & & & \\
APS & \includegraphics[width=0.04\linewidth]{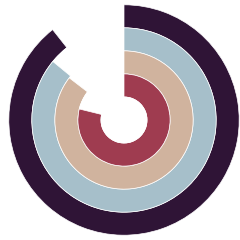} & \includegraphics[width=0.04\linewidth]{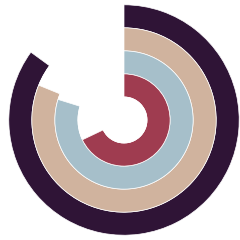} & \includegraphics[width=0.04\linewidth]{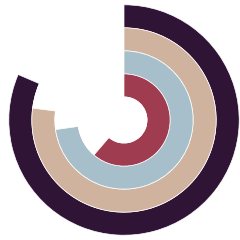} & \includegraphics[width=0.04\linewidth]{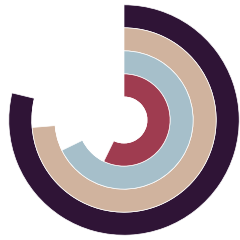}& \multicolumn{1}{c|}{\includegraphics[width=0.04\linewidth]{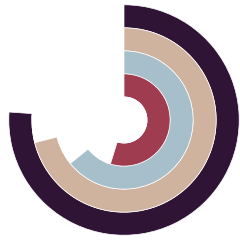}} & \includegraphics[width=0.04\linewidth]{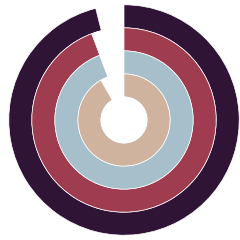} & \includegraphics[width=0.04\linewidth]{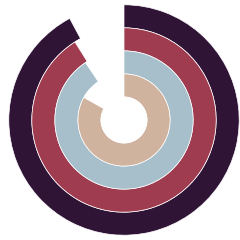} & \includegraphics[width=0.04\linewidth]{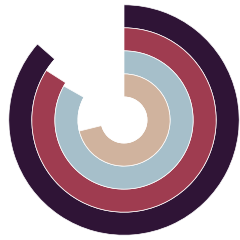} & \includegraphics[width=0.04\linewidth]{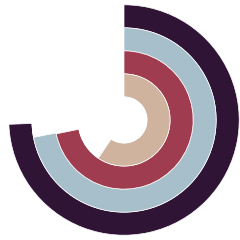} & \includegraphics[width=0.04\linewidth]{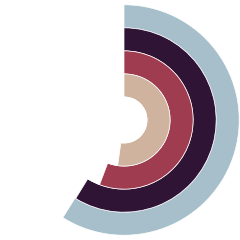}\\
Heart & \includegraphics[width=0.04\linewidth]{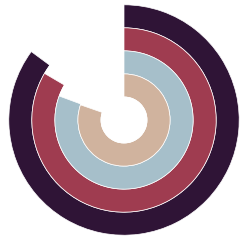} & \includegraphics[width=0.04\linewidth]{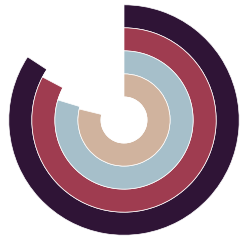} & \includegraphics[width=0.04\linewidth]{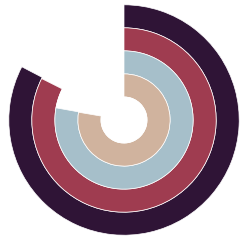} & \includegraphics[width=0.04\linewidth]{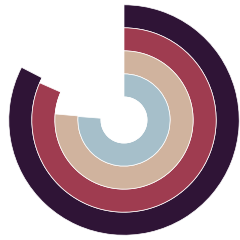}& \multicolumn{1}{c|}{\includegraphics[width=0.04\linewidth]{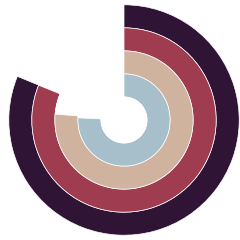}} & \includegraphics[width=0.04\linewidth]{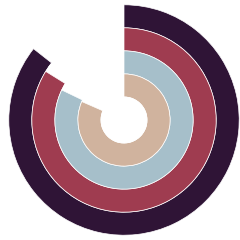} & \includegraphics[width=0.04\linewidth]{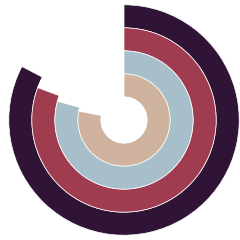} & \includegraphics[width=0.04\linewidth]{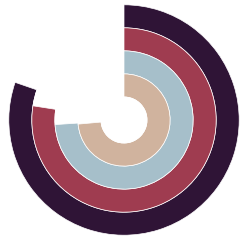} & \includegraphics[width=0.04\linewidth]{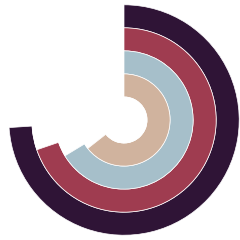} & \includegraphics[width=0.04\linewidth]{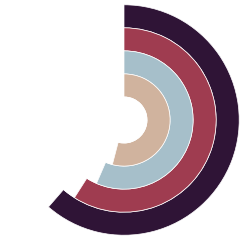}\\
MAGIC & \includegraphics[width=0.04\linewidth]{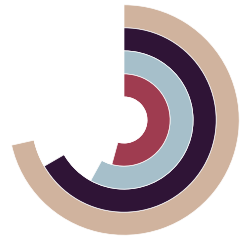} & \includegraphics[width=0.04\linewidth]{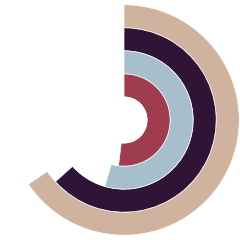} & \includegraphics[width=0.04\linewidth]{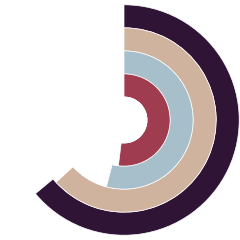} & \includegraphics[width=0.04\linewidth]{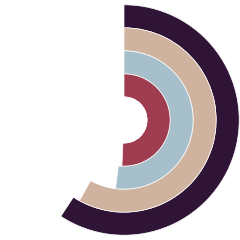}& \multicolumn{1}{c|}{\includegraphics[width=0.04\linewidth]{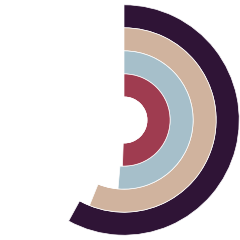}} & \includegraphics[width=0.04\linewidth]{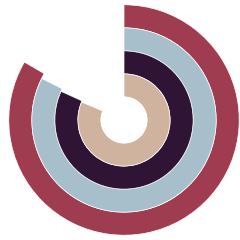} & \includegraphics[width=0.04\linewidth]{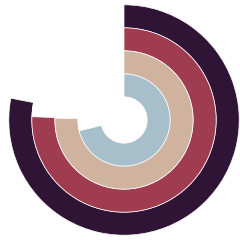} & \includegraphics[width=0.04\linewidth]{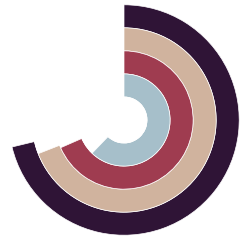} & \includegraphics[width=0.04\linewidth]{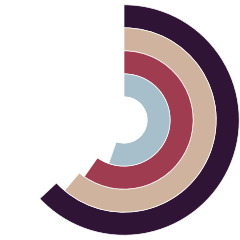} & \includegraphics[width=0.04\linewidth]{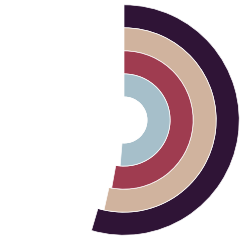}\\
Nursery & \includegraphics[width=0.04\linewidth]{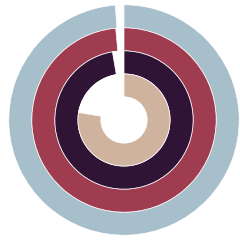} & \includegraphics[width=0.04\linewidth]{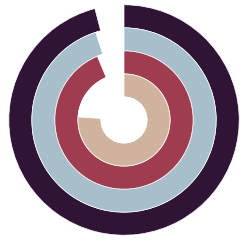} & \includegraphics[width=0.04\linewidth]{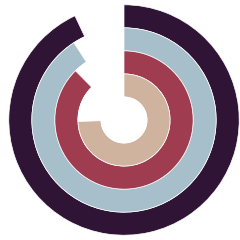} & \includegraphics[width=0.04\linewidth]{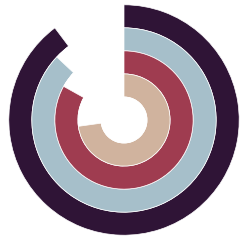}& \multicolumn{1}{c|}{\includegraphics[width=0.04\linewidth]{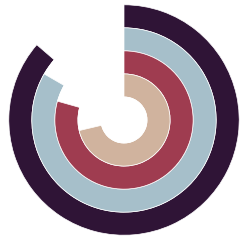}} & \includegraphics[width=0.04\linewidth]{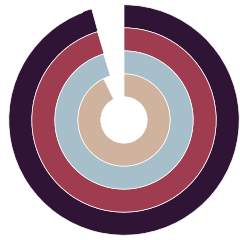} & \includegraphics[width=0.04\linewidth]{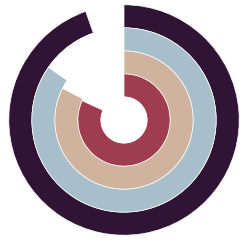} & \includegraphics[width=0.04\linewidth]{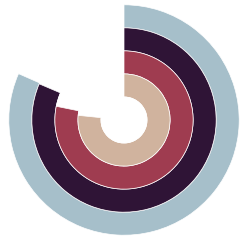} & \includegraphics[width=0.04\linewidth]{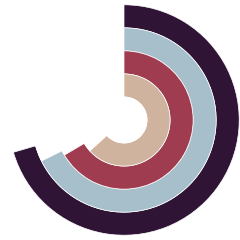} & \includegraphics[width=0.04\linewidth]{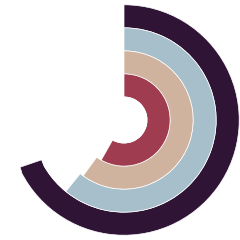}\\
WDBC & \includegraphics[width=0.04\linewidth]{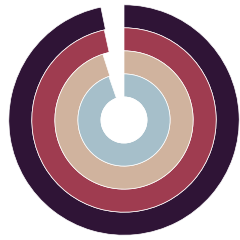} & \includegraphics[width=0.04\linewidth]{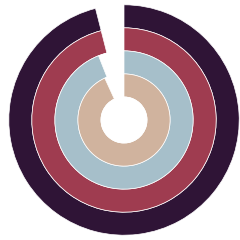} & \includegraphics[width=0.04\linewidth]{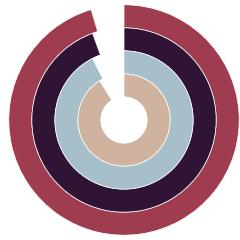} & \includegraphics[width=0.04\linewidth]{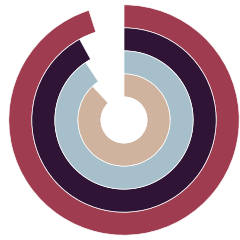}& \multicolumn{1}{c|}{\includegraphics[width=0.04\linewidth]{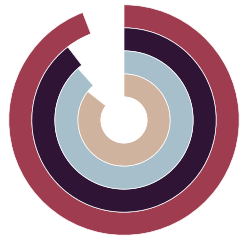}} & \includegraphics[width=0.04\linewidth]{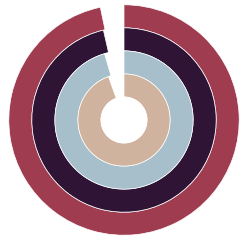} & \includegraphics[width=0.04\linewidth]{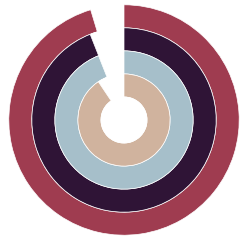} & \includegraphics[width=0.04\linewidth]{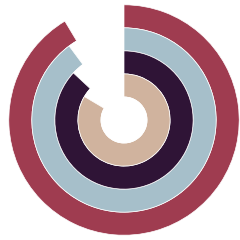} & \includegraphics[width=0.04\linewidth]{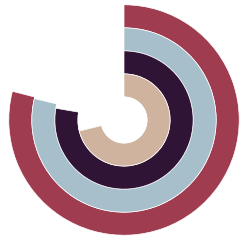} & \includegraphics[width=0.04\linewidth]{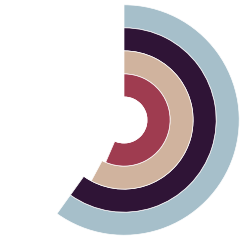}\\ \hline
\end{tabular}
\label{tab:underfing}
}
\end{table}

Upon analyzing the results, we observed that Rashomon ensembles consistently outperformed other models in both data drift and data shuffle scenarios, demonstrating their robustness to distribution changes. The Rashomon ensemble's ability to maintain superior performance under these perturbations showcases its capacity to adapt and generalize well to variations in data distributions.

The comparative analysis provided in Table \ref{tab:underfing} highlights the strengths of Rashomon ensembles in handling distribution drift. The findings suggest that the ensemble's ability to leverage diverse subgroups of models contributes to its robustness and adaptability, making it a promising approach for real-world applications where data distributions may evolve over time.

\subsection{Intra-Model Associations}

Out of the three datasets in which our approach was not able to beat the state-of-the-art, two can be explained by the Rashomon ratio. In most scenarios, a fair share of the sampled models presented performance statistically close to or superior to the all-in-one model, resulting in explanation diversity. In the cases of the Speed Dating and MADELON problems, less than $0.5\%$ of the sampled models (less than 500) were comparable to the all-in-one model. This entailed scarcity in possible different solution paths and the possibility that nearly all the present features provided a complementary view of the solution. This harmed our sub-space division, leading to non-representative groups and poor predictive power.

When exploring ensemble composition and robustness, we verified that once we filtered the underperforming models of III, this group population represented a sample of several ensembles that could be reached by direct optimizations over their constituents. On the opposite extreme, the group I represented the ensembles found following our proposed pipeline. It was important to remark that our approach depended on sampling the extremely complex model space, making it highly unlikely that the clusteroids found in each repetition were the same. However, the high Jaccard coefficient associated with the high cosine similarity between the SHAP vectors provided strong evidence that the centroids found in each repetition were contiguous, resulting in similar clusteroids that led to similar ensembles. Finally, group II represented a sample of possible optimization paths within the respective Rashomon Sets.

In all experiments, group III not only presented the lowest values of Jaccard and SHAP similarity but also consisted of the sparsest point cloud. Groups I and II were more cohesive and concentrated over high values of similarity with the reference model. When considering that all models had a statistically equal or higher performance than the reference model, it was reasonable to conclude that the pipeline involving Rashomon Sets reduced the impact of underspecification while retaining concise predictions. When further exploring drift by introducing Gaussian noise and shuffling feature values, the robustness of Rashomon ensembles became evident. In most explored scenarios, our approach remained the best-performing model, even when considering the MAGIC dataset, in which Rashomon ensembles had slightly worse performance than other models.

\begin{figure}
\centering
\begin{subfigure}[b]{0.485\textwidth}
\centering
\includegraphics[width=\textwidth]{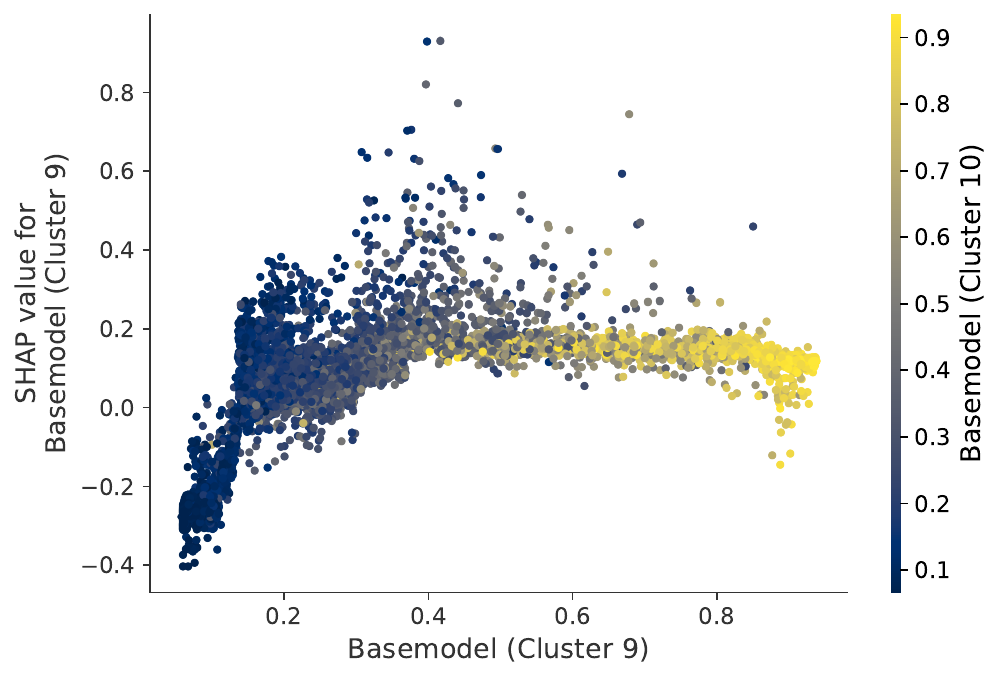}
\caption{Dependence between models 9 and 10.}
\label{dep8-10}
\end{subfigure}
\begin{subfigure}[b]{0.485\textwidth}
\centering
\includegraphics[width=\textwidth]{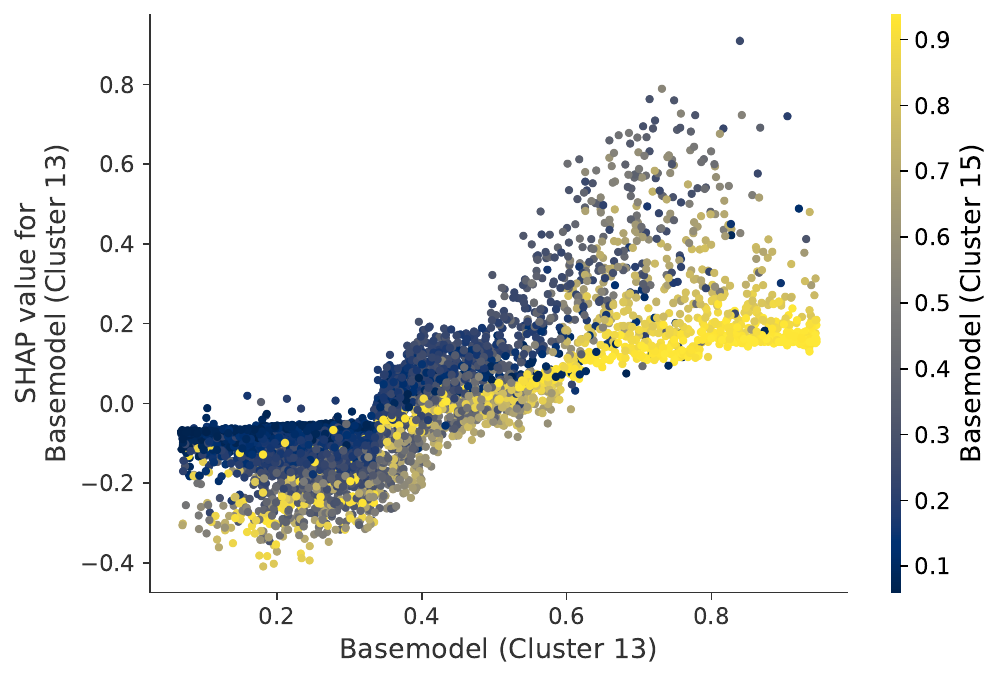}
\caption{Dependence between models 13 and 15.}
\label{dep13-15}
\end{subfigure}
\hspace{0.01\textwidth}
\caption{Dependence between relevant base models in the MAGIC Rashomon ensemble.}
\label{fig:dependence}
\end{figure}

Further investigation of the relationships learned in the ensemble revealed a variety of interesting patterns, as illustrated by Figure \ref{fig:dependence} with partial dependence plots derived from the constituents' Shapley values. For instance, in Figure \ref{dep8-10}, we observed that the ensemble learned to rely on the output of the 9th base model to give mostly positive predictions, with nearly all points above the 0.2 probability threshold presenting a positive SHAP value. We also verified that models 9 and 10 provided a complementary view of the problem, as higher prediction values of 9 were likewise associated with high prediction values of 10. Figure \ref{dep13-15} showed that models 13 and 15 mostly contrast. Whenever there was disagreement, the relative importance of Model 13 increased. Similarly, model 13 presented Shapley values close to zero when both models agreed, resulting in a concentration of yellow points on the lower side of the distribution. 

\section{Unique collaborative datasets}

To validate the effectiveness of our approach in real-world scenarios, we present the results of our evaluation across distinct applications conducted in collaboration with various companies and institutions: stainless steel surface defect detection, COVID-19 hemogram detection from blood counts, energy consumption forecasting, and medical bills auditing. In all cases, new unique handcrafted datasets were created to explore each of the mentioned problems. Although these problems may seem vastly different, they share a common characteristic: the absence of a clear consensus among specialists on the best solution. Instead, they appear to exhibit multiple possible and effective solutions without a definitive optimal model or explanatory factors. This implies the presence of a large Rashomon Set fit for the application of our approach.

\subsection{Surface Defects in Stainless Steel Manufacturing}
\label{chap:aperam}

The quality of duplex stainless steel can be compromised by surface defects, such as slivers, which increase production costs due to their detection occurring only during the final inspection stage. In collaboration with \textit{APERAM South America}, we analyzed the chemical composition and hot rolling process variables of duplex stainless steel plates. Chemical compositions were measured using spectrometers, capturing the relative abundances of 20 elements, while 1,160 temporal hot-rolling variables were collected. We extended the feature space by considering elemental ratios, increasing the total to 220 chemical attributes. For the hot rolling data, we discretized the temporal series into 30-second intervals and calculated statistical moments, resulting in 11,488 hot rolling features after filtering non-actionable variables. This data was used to predict the likelihood of sliver formation as a binary classification problem.

As described by \citet{barbosa2007valuation}, identifying factors contributing to sliver formation is challenging, as these defects can arise from a combination of process variables or chemical compositions at various steelmaking stages. We hypothesize that different data structures might correspond to different models and that models with similar feature importance distributions likely reflect similar mechanisms. Our analysis identified correlations between feature sets and predictions, indicating that some features are more sensitive to specific defect formation mechanisms~\citep{zuinaperam}.

\begin{figure}[b]
	\begin{subfigure}[b]{0.17\linewidth}
	\includegraphics[height=4cm]{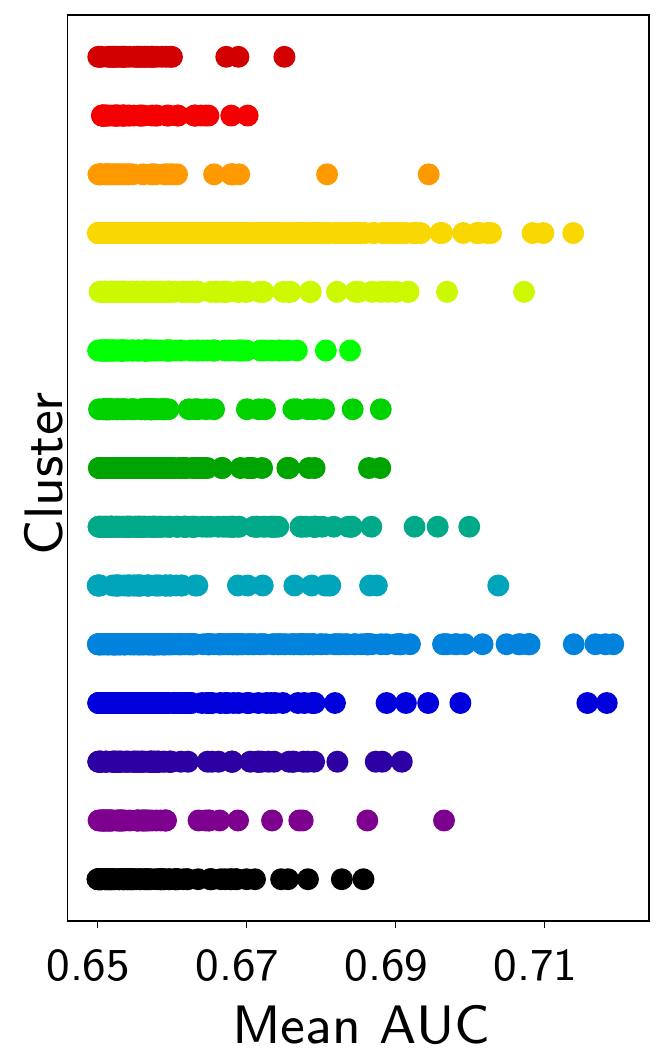}
	\caption{AUCs.}
	\label{fig:corrmean}
	\end{subfigure}
	\centering
	\begin{subfigure}[b]{0.39\linewidth}
	\includegraphics[height=4cm]{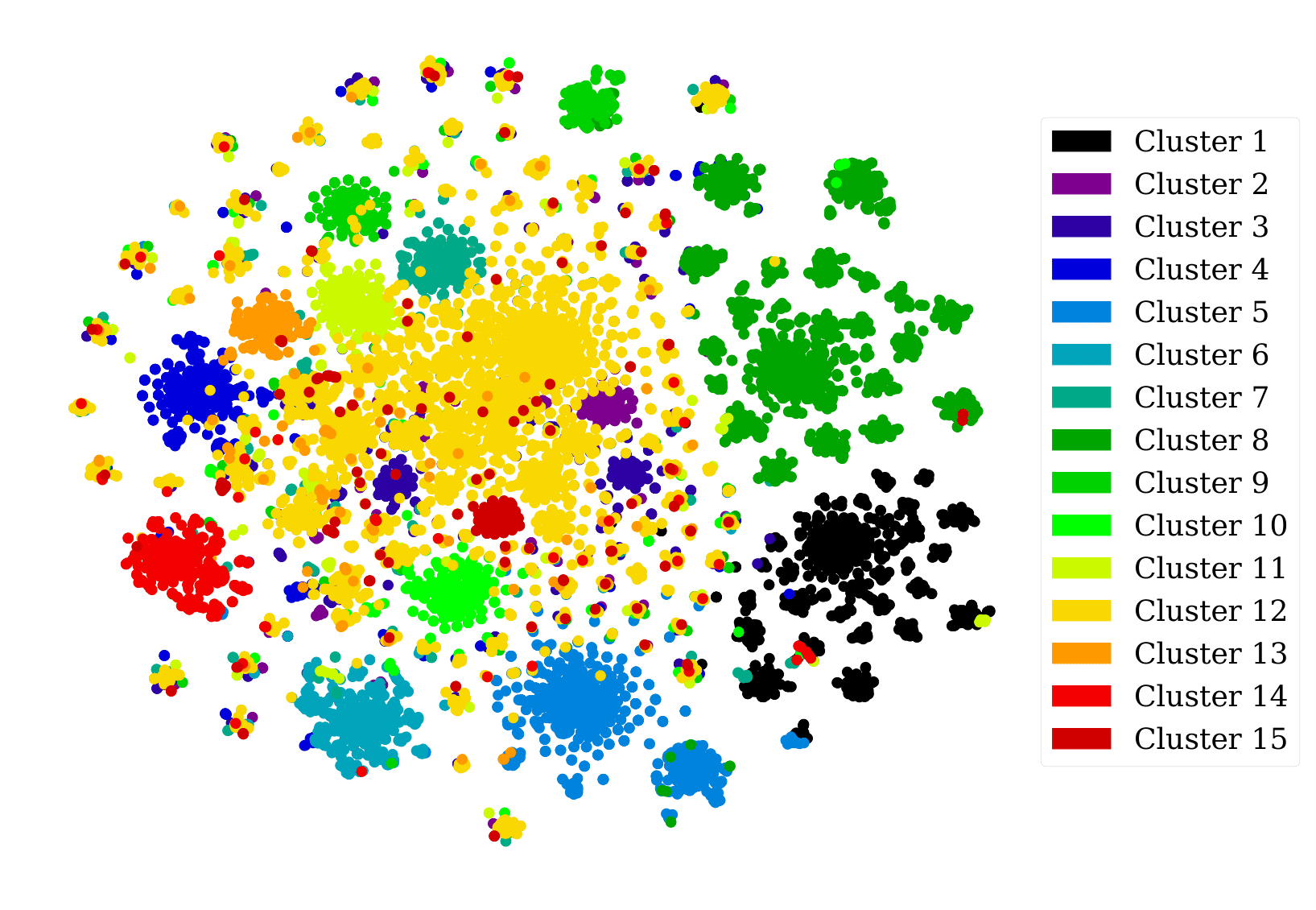}
	\caption{All found clusters.}
	\label{fig:map1a}
	\end{subfigure}
	\hspace*{0.02\linewidth}
	\begin{subfigure}[b]{0.39\linewidth}
	\includegraphics[height=4cm]{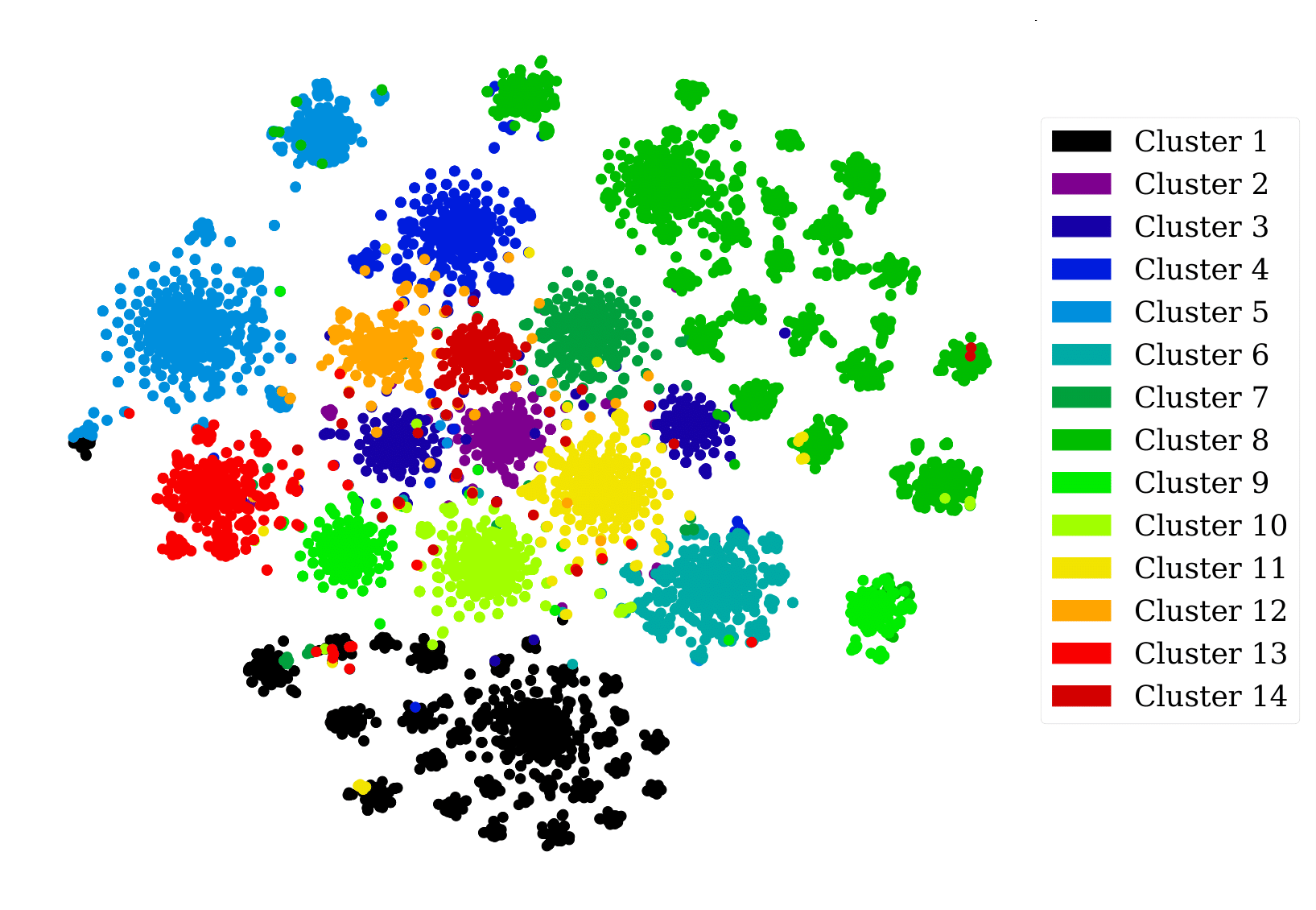}
	\caption{Filtering the sparse cluster.}
	\label{fig:map1b}
	\end{subfigure}
\caption{T-SNE visualization of the sampled Rashomon space for models trained on the steel plate defects problem. Each point represents a model. Models are placed according to the defect explanations assigned to each steel plate so that models that possess similar SHAP values are placed next to each other in space. The color indicates the cluster for which the model was assigned.}
\label{fig:map1}
\end{figure}

To explore the model space, we constructed a Rashomon Set by sampling 75,000 models for each feature set size until no significant performance improvement was observed, resulting in 1,049,999 models. Since no strong baseline exists in the literature, we compared it against an all-in-one model, which included all features and achieved an AUC of 0.62. Models surpassing this threshold formed the Rashomon Set, comprising 63,374 models (6.04\%). For visualization purposes, Figure \ref{fig:map1} presents a t-SNE projection of 2,500 models from the Rashomon Set, clustered by their explanations.

\begin{figure}
\centering
\includegraphics[width=0.6\linewidth]{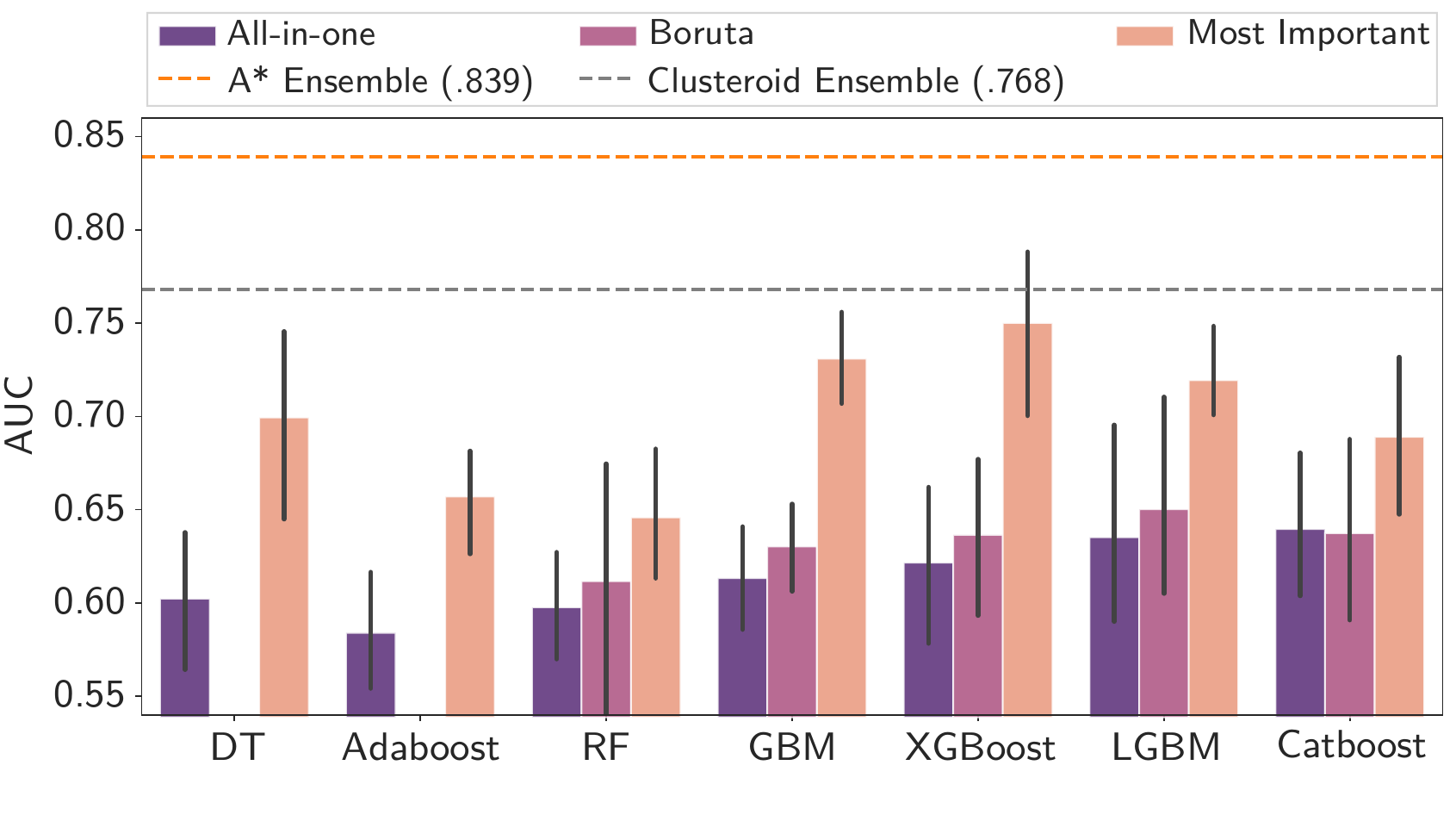}
\caption{Comparison of different algorithms to our approach in the steel manufacturing defects problem. Even when employing the clusteroid ensemble, in which most constituents are underperforming, our approach exceeds other state-of-the-art results.}
\label{fig:ensembleaperam}
\end{figure}

We selected the optimal model for each cluster using two approaches. In the first one, we simply elect the clusteroids as the ensemble constituents. In the second approach, we reframe the problem of model selection as a graph search problem. We consider each possible model as a node in a graph, and two models are connected if they can be reached from simple feature addition. For each cluster, we employ the A* search algorithm in this graph using as a heuristic the AUROC of each model and penalizing paths that lead to models that lie outside the cluster boundaries.  Figure \ref{fig:ensembleaperam} compares the performance of our ensemble (both A* and clusteroid models) against state-of-the-art tree-based ensemble techniques and other classical algorithms with and without feature selection.

Once representative models were found, we asked for insights from metallurgical experts. The main lesson was that there were cases where some conclusions did not fit with realistic scenarios. For example, some models hinted towards increasing carbon concentration to such high levels that the steel plates could not be classified as Duplex anymore. These inconsistencies highlight the advantage of our approach, which allows domain experts to discard unrealistic models at production time without involving data scientists and the need to retrain and evaluate new models, since, during training time, we sample from the complete model. We verified that this sort of approach drastically increases the power of domain experts and helps build trust in the models, as they feel in control of these domains and business-specific decisions concerning model development. This insight led to a patent further explained in Section \ref{sec:patent}. After filtering unrealistic patterns, the most relevant ones were turned into production rules and employed in the 2019 and 2020 steelmaking processes. A reduction from $49\%$ to $3\%$ in the occurrence of heating slivers was reported, showing the potential of this strategy in real-world problems and validating the proposed framework.

%%%%%%%%%%%%%%%%%%%%%%%%%

\subsection{Diagnosing COVID-19 from Complete Blood Counts}

In late 2019, Severe Acute Respiratory Syndrome Coronavirus 2 (SARS-CoV-2) emerged in Wuhan, China, leading to a global outbreak of Coronavirus Disease 2019 (COVID-19) within weeks~\citep{wu2020nowcasting,hui2020continuing}. By the time of writing, over 630 million COVID-19 cases and 6.67 million deaths have been reported worldwide. Diagnosing COVID-19 is complicated by initial symptoms such as fever, dry cough, and tiredness, which overlap with many respiratory diseases~\citep{dias2020orientaccoes}.
Complete blood counts (CBC) are commonly used for diagnostic purposes~\citep{walters1996interpretation}. As a low-cost test, CBC measures various analytes and can provide insights into potential diseases, including infectious ones. However, correlating specific CBC results with particular diagnoses can be challenging, as similar changes may occur across different diseases.

In analyzing complete blood counts of individuals with COVID-19 infection in isolation, we find some changes to be quite characteristic of the disease~\citep{formica2020complete, foldes2020plasmacytoid, hu2020characteristics}, hinting at the potential for automated detection and screening of the disease using machine learning. However, many possible analyte combinations might lead to the same conclusion regarding a target disease, thus elucidating the Rashomon Effect and posing a suitable problem to deploy our ensembling approach. Numerous models have been proposed for automated COVID-19 diagnosis using CBC and omics data. We argue that the detection performance of these models may be biased due to non-unique patterns not exclusive to SARS-CoV-2. Our study utilizes a dataset from 2016 to 2021, in collaboration with \textit{Grupo Fleury}, encompassing blood tests and RT-PCR results across Brazil for COVID-19 and other pathologies, including Influenza-A and H1N1~\citep{zuinnature}.

Data collection for 2020 and 2021 includes $900,220$ unique individuals, $809,254$ CBCs, and $1,088,385$ RT-PCR tests, with $21\%$ ($234,466$) positive results and less than $0.2\%$ ($1,679$) inconclusive results. This work does not consider demographic, prognostic, or clinical data, such as ethnicity or hospitalization. We frame the task as a binary classification problem and analyze two timeframes: the early pandemic stage (the first wave of COVID-19 in Brazil) and a second stage post-November 2020, coinciding with the emergence of the \textit{P1} variant that led to a health crisis in Amazonas~\citep{he2021new}.

Our algorithm's first step involved sampling $100,000$ models from the complete model space, examining both raw analytes and analyte ratios as features. Previous studies have explored the use of CBC for COVID-19 detection through various machine-learning methods and, as such, provide reference points for our Rashomon space induction. Notable examples include a naive Bayes classifier with an AUROC of 0.84~\citep{avila2020hemogram}, a gradient boosting machine achieving 0.81~\citep{silveira2020prediction}, and a neural network and random forest model reaching an AUROC of 0.94~\citep{banerjee2020use}. Other studies report AUROC values ranging from 0.88 to 0.94, with one analysis involving $114,957$ individuals in a COVID-negative cohort~\citep{soltan2020rapid}.

\begin{figure}[h!]
\centering
\includegraphics[width=0.425\linewidth]{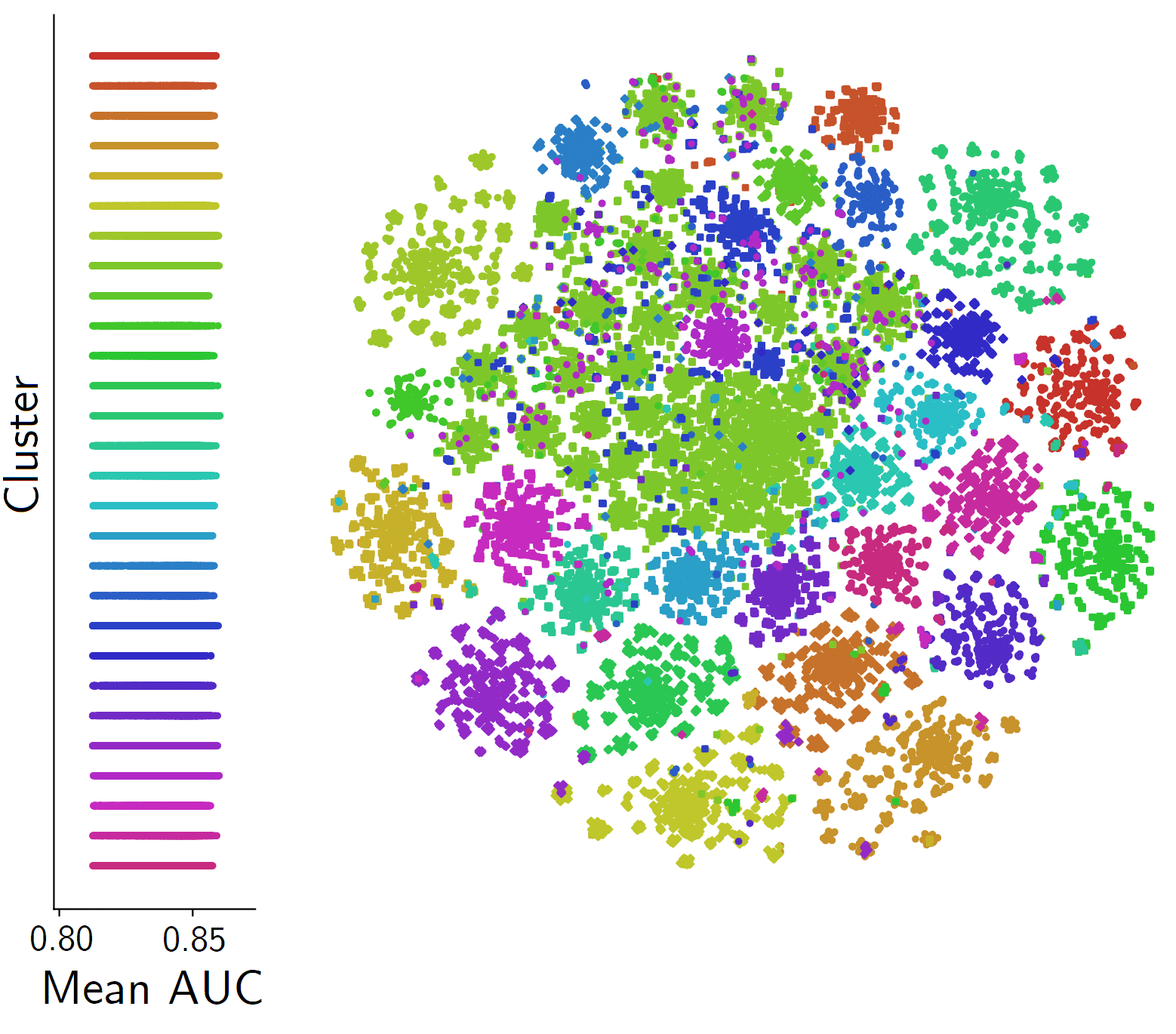}
\begin{center}
\end{center}
\caption{TSNE visualization of the COVID-19 Rashomon space of models trained to predict COVID-19 diagnosis. No clear relationship exists between cluster assignment and predictive power. Cluster 11 appears to be more spread over the space, overlapping with other clusters, while the remaining ones are mostly concise, reminiscent of the steel-plate defects of Rashomon space.}

\label{fig:fleuryrashomon}
\end{figure}

Based on this literature, we established a performance threshold of AUROC 0.81 to define a minimally performant model for our Rashomon Set, resulting in a sampled model space $\mathcal{H}'$ containing 47,708 models (47.71\% performing better than the literature threshold). This substantial Rashomon Set indicates that blood-related features are significant for preliminary disease diagnosis. Figure \ref{fig:fleuryrashomon} illustrates the induced Rashomon space, highlighting divisions after clustering models based on their explanatory vectors.

\begin{figure}
\includegraphics[width=\linewidth]{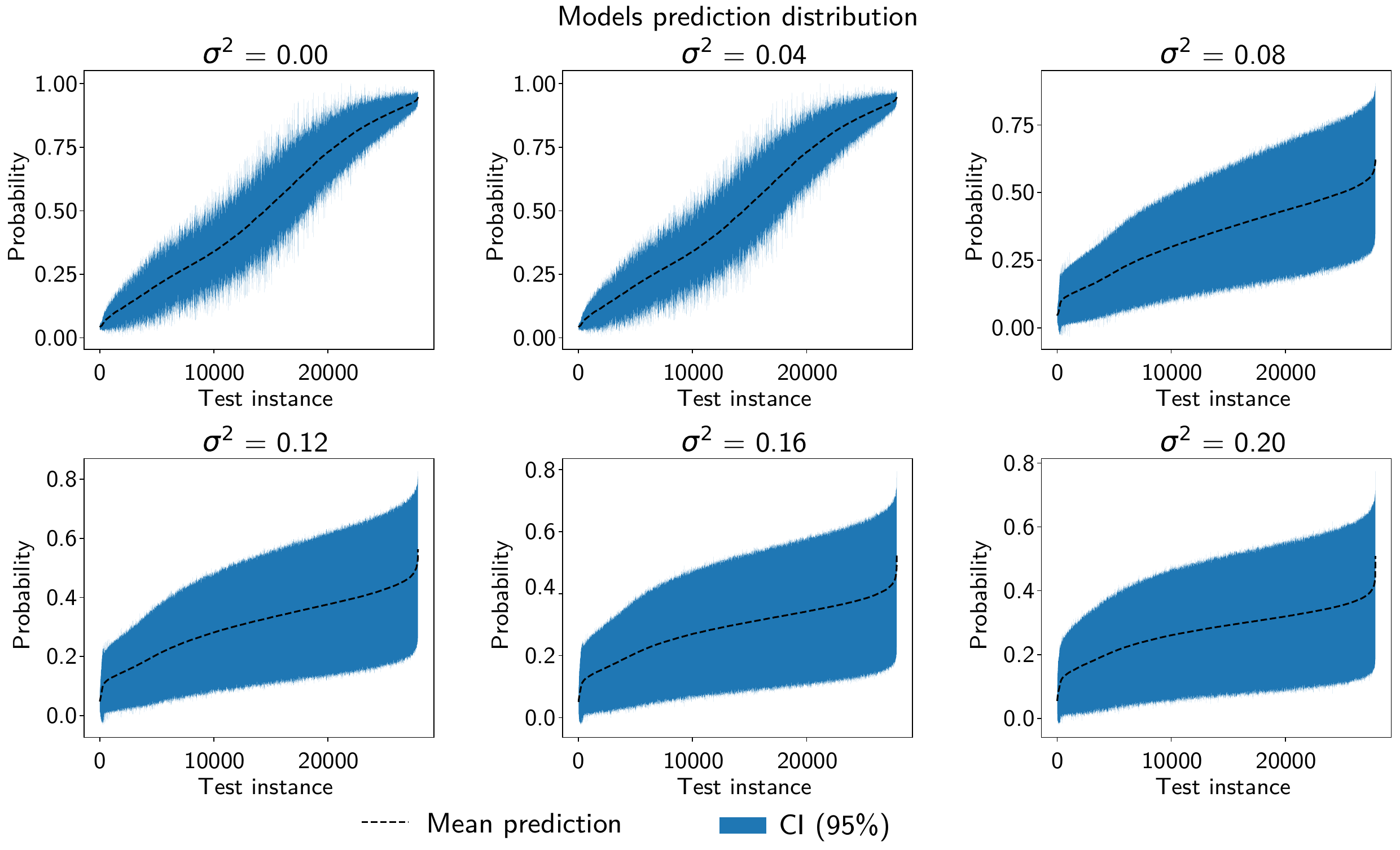}
\caption{Effect of introducing noise to input features of models trained for COVID-19 diagnosis from blood analytes.}
\label{fig:sigmafleury}
\end{figure}

In line with previous findings in the literature, not all CBC analytes are relevant for differentiating the target diseases, and some may detract from model performance. To refine feature selection for each representative Rashomon model, we represented the model space as a directed acyclic graph (DAG), where each node corresponds to a feature subset. The A* algorithm~\citep{astar} was applied, utilizing the AUROC of models represented by each vertex as a heuristic. Once models were selected, their suitability as Rashomon constituents was assessed. Our hypothesis posits that models exhibiting disagreement under data drift and with diverse explanations can form a more robust ensemble. In Figure \ref{fig:sigmafleury}, we introduced Gaussian noise to normalized features and examined the probability distributions returned by each constituent. We observed a direct correlation between noise and confidence intervals, indicating increased divergence among models under drift and thus supporting the appropriateness of our constituent selection.

\begin{figure*}
\centering
\includegraphics[width=0.8\textwidth]{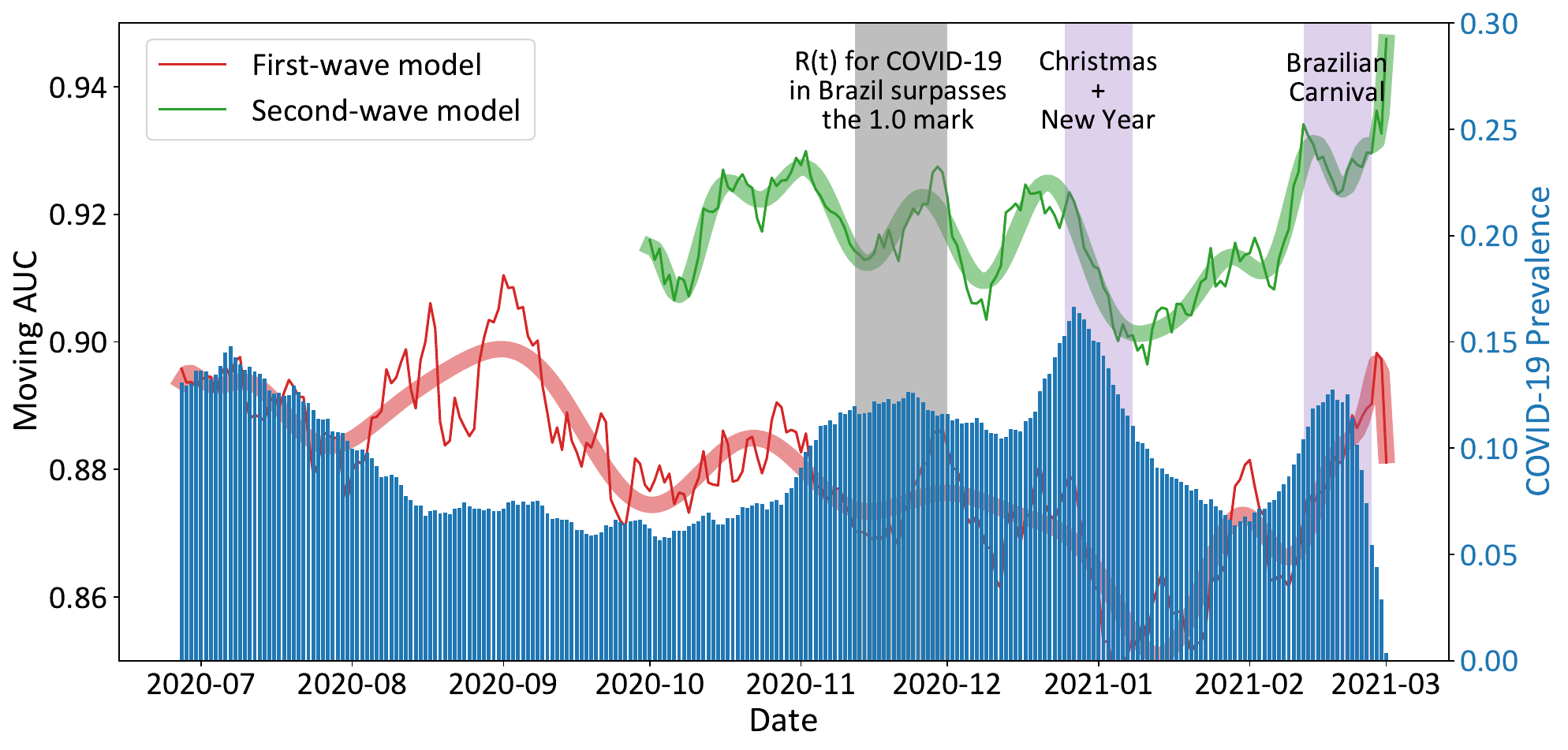}
\caption{AUROC fluctuation over time considering a 7-day sliding window on $357\,956$ CBCs. The red line represents the model trained only on the first wave of COVID-19 in Brazil data (up to 2020-06), while the green line represents a model trained with data immediately before the start of the second wave of COVID-19 in Brazil (up to 2020-10). Thinner lines depict the measured AUROC values, while thicker lines illustrate their respective trends. The second-wave model can retain performance during the second wave while the performance of the first-wave model deteriorates. Key events are marked in gray and purple.}
\label{fig:movingauc}
\end{figure*}

By mid-November 2020, Brazil entered the second wave of COVID-19, which eventually led to the collapse of the health system in Manaus, the capital of  Amazonas, a state in Brazil \citep{emmerich2021comparisons}. One of the explanations raised by the local government was the emergence of a new  COVID-19 variant, known as $20J/501Y.V3$ - or simply P.1 \citep{he2021new}. To evaluate the performance of our COVID-19 model as the SARS-CoV-2 virus mutates, we trained it on two distinct points in time. The first one, which we will refer to as the `First-wave model', was trained using the training set associated with the first wave. The second, which we will refer to as the `Second-wave model', was trained using the training set associated with the second wave in Brazil. Figure \ref{fig:movingauc} presents AUROC values obtained from these models during the pandemic until March 2021, utilizing a 7-day sliding window and depicting the prevalence of COVID-19 cases over time. We focus on three periods of interest: when the reproduction number exceeded 1.00, during the holiday season, and during Carnival, which included gatherings despite event cancellations.

\begin{figure}[h!]
\centering
\includegraphics[width=0.6\linewidth]{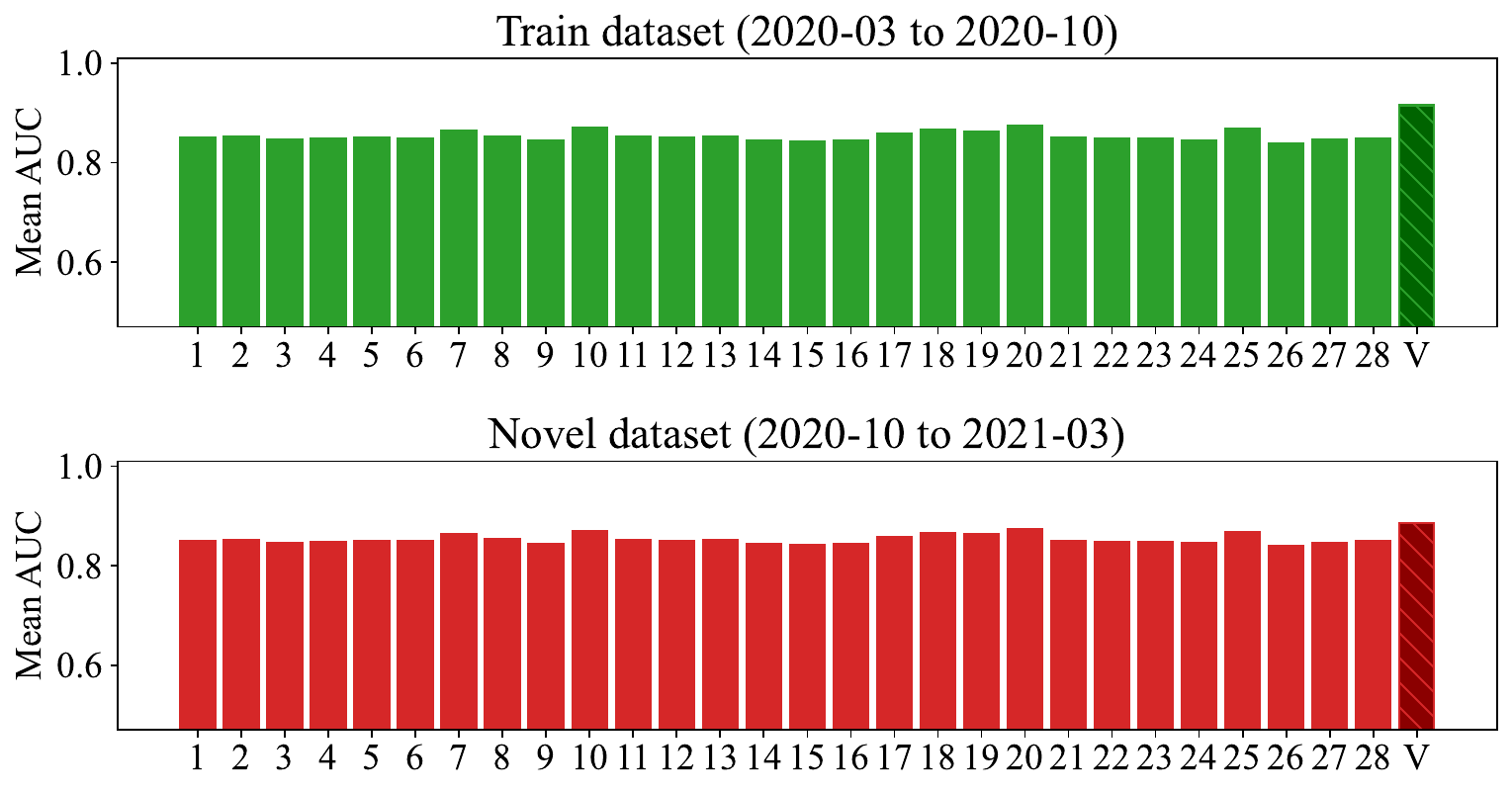}
\begin{center}
\end{center}
\caption{Comparison of model performances across periods. Each constituent model is represented by the Cluster from which it hailed. Both in the train and novel datasets, we can observe that all constituent models behave similarly. We should expect data distributions from late 2020 and early 2021 to be properly represented in data before October 2020.}
\label{fig:rashensnovel}
\end{figure}

We evaluate the COVID-19 Rashomon ensemble on both periods using the model trained with data up to October, illustrated in Figure~\ref{fig:rashensnovel}. Performance on both periods appears to be comparable, thus implying that the constituents were able to properly generalize to the second wave. We also verify that the Rashomon ensemble remains a suitable approach, outperforming all constituents in either scenario. Further, the empirical risk found during training can be used to estimate the empirical risk on production, as no significant divergences were observed. Overall, leveraging our Rashomon ensemble technique, we predicted COVID-19 RT-PCR outcomes using CBC data, achieving an AUROC of 0.917.

%%%%%%%%%%%%%%%%%%%%%%%%%%%%%%%%%%

\subsection{Including specialists in the model creation process}
\label{sec:patent}

A patent was filed focusing on incorporating experts and decision-makers into the AI model development pipeline~\cite{liapatente}.
The core of the patent is to foster a sense of shared responsibility and co-creation, where researchers make decisions regarding the technical aspects while domain experts provide guidance on domain-specific topics such as feature selection, exclusion, and the interpretation of discovered patterns. This balance ensures that the model development process integrates both technical rigor and domain knowledge.

Among the methods included in the patent is the usage of the developed Rashomon ensembles framework, which identifies multiple contrasting patterns in the data as per the Rashomon Effect. Frequently, some of these patterns are closely aligned with experts' experience and existing domain literature, enhancing the specialists' trust in the model. This alignment thus leads to more constructive discussions, especially regarding patterns that deviate from established knowledge. Since experts find some patterns consistent with their mental models, we verified that they are more likely to engage with the results and question potential gaps in the literature rather than dismissing them outright.

\begin{figure*}[h!]
\centering
\begin{subfigure}{0.43\linewidth}
\includegraphics[height=5.8cm]{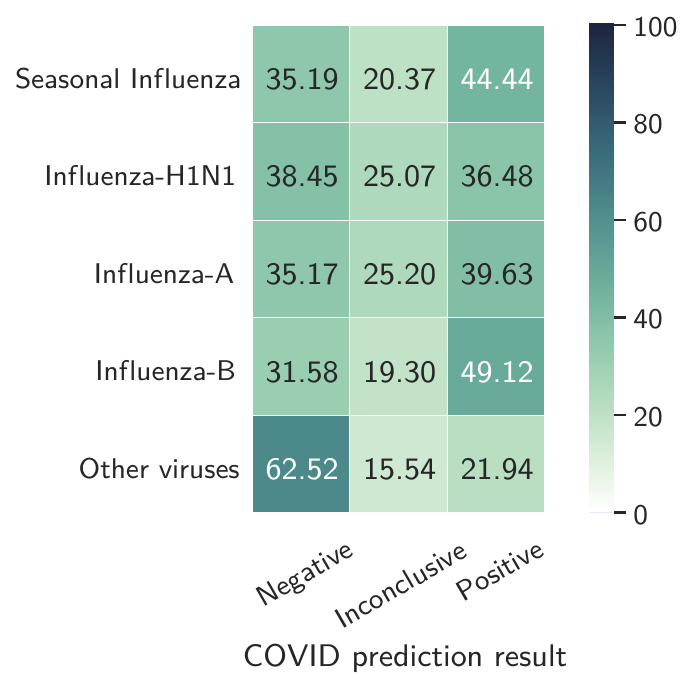}
\caption{Before expert input.}
\end{subfigure}
\begin{subfigure}{0.265\linewidth}
\includegraphics[height=5.8cm]{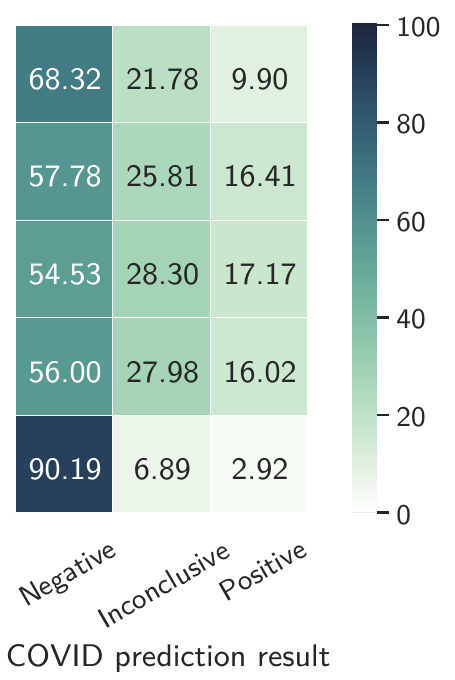}
\caption{After expert input.}
\end{subfigure}
\caption{Results of different models evaluated on $11\,116$ CBCs from 2019 individuals with confirmed RT-PCR results for diverse viruses, including Influenza-A, Influenza-B, Influenza-H1N1, and Seasonal Influenza. Left is a model trained only on SARS-CoV-2 data. CBC ($\mathbf{-}$) includes only COVID-19 ($\mathbf{-}$). The model to the right was trained using data of diverse viruses, including SARS-CoV-2. CBC ($\mathbf{-}$) also includes viruses other than SARS-CoV-2.}
\label{fig:res1}
\end{figure*}

In the specific case involving COVID-19 diagnosis from blood-count data, this methodology prompted experts to question whether a model trained solely on COVID-19 cases could distinguish it from other respiratory diseases. This line of inquiry proved crucial in improving the model's final performance, demonstrating the value of expert involvement in refining and validating the patterns uncovered by the AI models. Figure~\ref{fig:res1} shows how different models perform specifically on individuals who were infected by some viruses in 2019. The ideal result would be all predictions being negative for COVID-19. However, models trained solely on COVID-19 data failed to do so (Figure~\ref{fig:res1}a). Including viruses other than SARS-CoV-2 during training increases the performance of 2019 data (Figure~\ref{fig:res1}b).

%%%%%%%%%%%%%%%%%%%%%%%%%%%%%%%%%%

\subsection{Data Divergence in Hospital Settings}

We investigated the behavior of Rashomon ensembles under train and production divergence using two datasets: one related to COVID-19 and another to Alzheimer's disease. Both datasets contain information from different contexts, indicative of shifts in data distribution. Due to the lack of a strong baseline in the literature, we adopted the "all-in-one" approach to establish a reference model ($f_{ref}$) and quantify uncertainty ($\epsilon$) for the ensemble.

\vspace{0.1in} \noindent\textbf{COVID-19:} This dataset is an initiative of the S\~ao Paulo Research Foundation (FAPESP) and includes pseudonymized data from two Brazilian hospitals: \textit{Benefic\^encia-Portuguesa Hospital} (HBP) with $91,648$ exams and \textit{S\'irio-Liban\^es Hospital} (HSL) with $37,643$ exams. The data encompasses clinical and laboratory exams as well as hospitalization information. The binary classification task aims to predict the death prognosis of COVID-19 patients 20 days prior. The training dataset comprises exams from $453$ individuals hospitalized at HBP, while our considered production dataset consists of exams from $4,018$ individuals hospitalized at HSL, highlighting potential distribution drift.

\vspace{0.1in} \noindent\textbf{Alzheimer’s Disease:} This dataset includes patients with suspected Alzheimer's Disease symptoms, featuring attributes such as gender, age, education level, and lab results. The binary classification task predicts whether a patient is diagnosed with Alzheimer's. Data is sourced from the Geriatrics and Neurology departments, each presenting unique socio-economic characteristics. The training dataset consists of $154$ exams from the geriatrics department, and our considered production dataset consists of $166$ exams from the neurology department, ensuring divergence due to the inclusion of non-geriatric patients.

\begin{figure*}
\centering
\begin{subfigure}{0.48\linewidth}
\centering
\includegraphics[height=4.25cm]{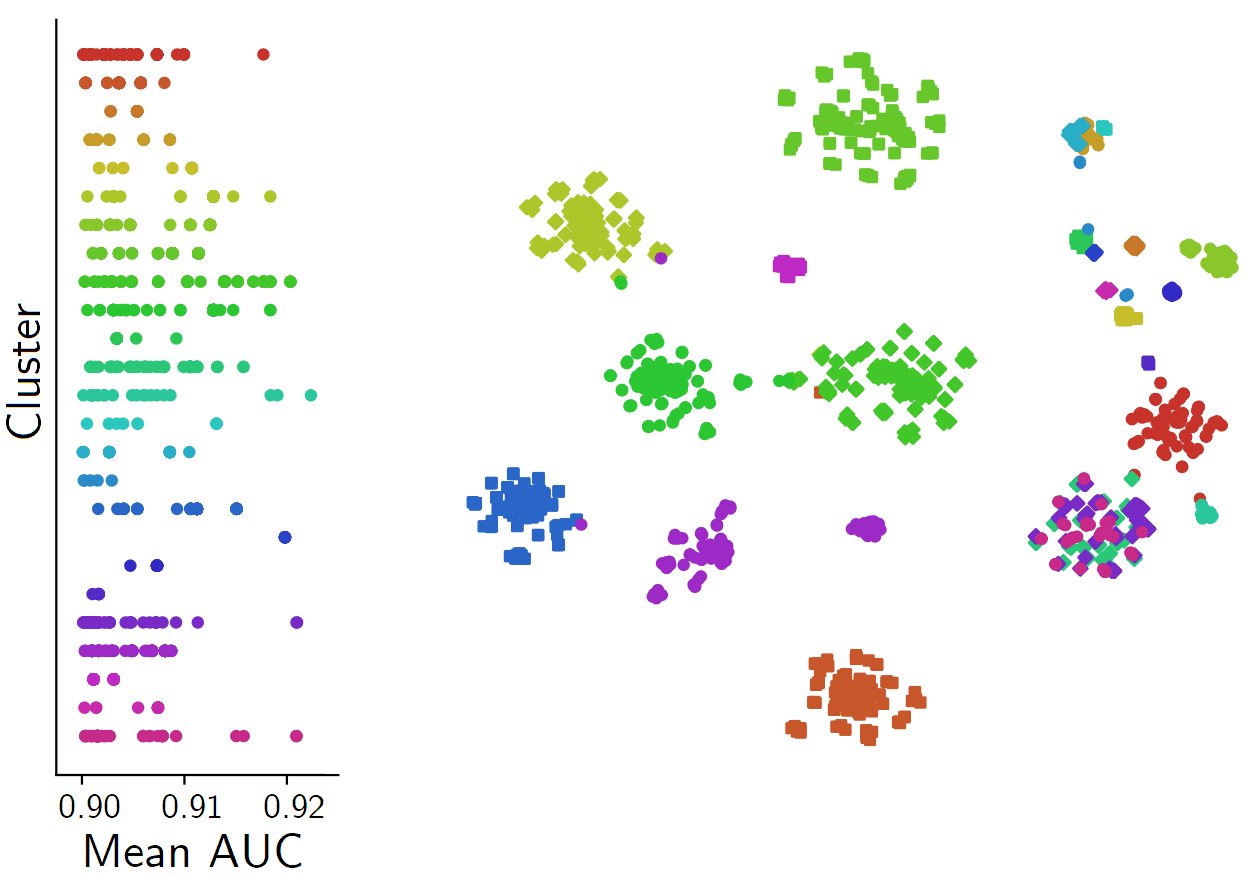}
\caption{COVID-19}
\label{fig:tsnecovid}
\end{subfigure}
\begin{subfigure}{0.48\linewidth}
\centering
\includegraphics[height=4.25cm]{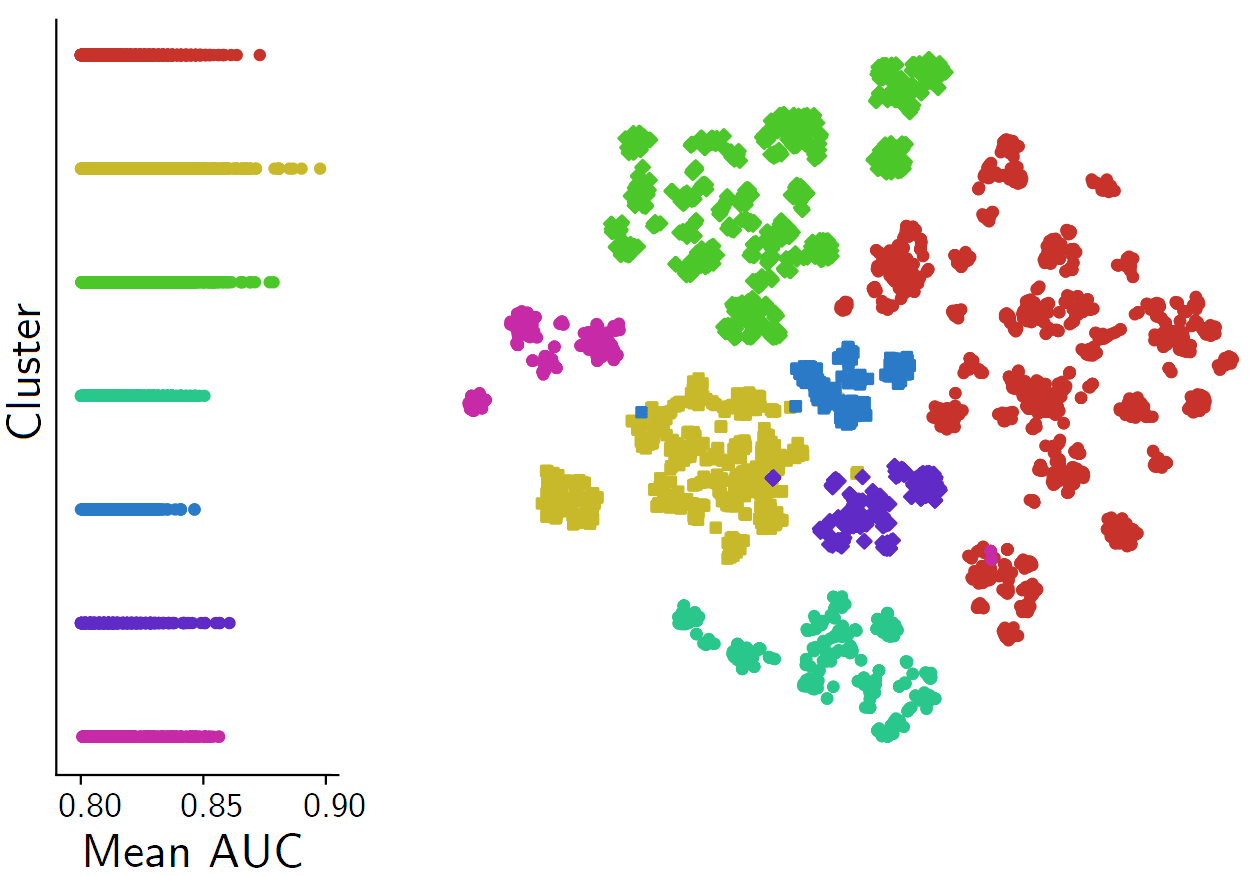}
\caption{Alzheimers'}
\label{fig:tsnealz}
\end{subfigure}
\caption{TSNE visualization of the Rashomon space of each problem.}
\label{fig:tsnerashos}
\end{figure*}

The all-in-one approach involves creating a comprehensive model that utilizes the entire dataset for predictions. We trained a decision tree model with all available data and features, referred to as the "all-in-one" model, which captured overall patterns. The model achieved average AUROC values of $0.90$ and $0.81$ for the COVID-19 and Alzheimer’s datasets, respectively, establishing benchmarks for minimal performance in the Rashomon Set. After sampling 100,000 models for each dataset, this resulted in subspaces containing $2,554$ and $6,251$ models, leading to Rashomon ratios of $2.5\%$ and $6.2\%$. Figure \ref{fig:tsnerashos} illustrates the Rashomon subspaces identified through clustering. We employed t-distributed Stochastic Neighbor Embedding (t-SNE) for visualization, where each point represents a model, colored by cluster assignment. K-means clustering was utilized to define subspaces, with the number of clusters determined by maximizing the silhouette value.

To evaluate model performance under train and production divergence, we combined models in a voting scheme. Since the task is a binary classification, the probability returned for the ensemble constitutes the absolute agreement rate of the constituents (e.g., for an ensemble of 10 constituents, a probability of 80\% means that 8 out of the 10 constituents agreed on predicting the positive class). Thus, probabilities near $0\%$ or $100\%$ indicate strong agreement, while values close to $50\%$ suggest uncertainty. Voting provides an interpretable measure of prediction reliability, beneficial in cases of production divergence and unknown data distributions. However, we also considered a stacking scheme where a meta-model learns to optimally combine the constituent outputs given the patterns present in the training data.

\begin{figure}
\centering
\begin{subfigure}{0.65\linewidth}
\includegraphics[width=0.95\linewidth]{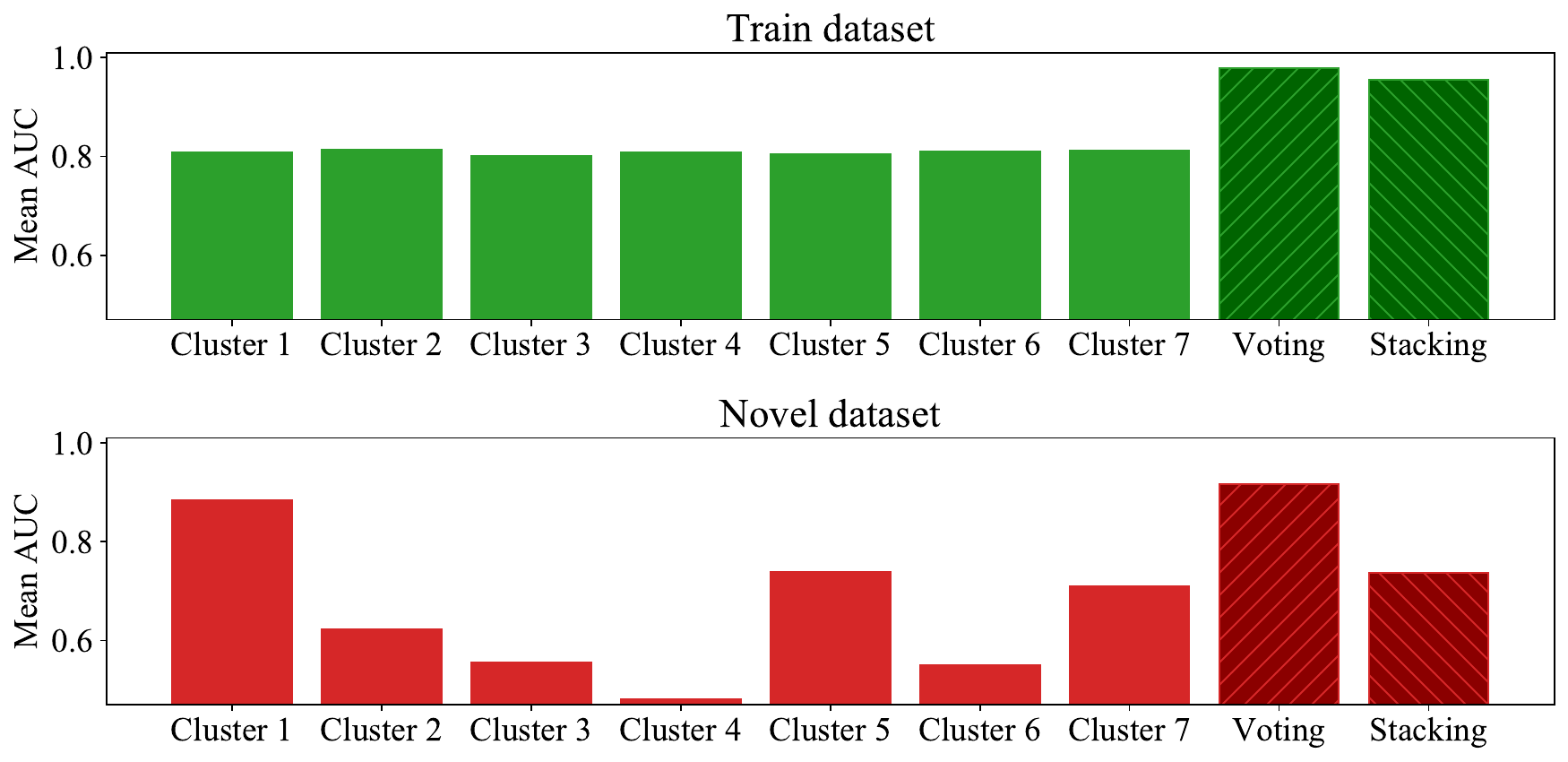}
\caption{COVID-19 dataset pairs.}
\label{fig:covidrashomonens}
\end{subfigure}
\begin{subfigure}{0.65\linewidth}
\includegraphics[width=0.95\linewidth]{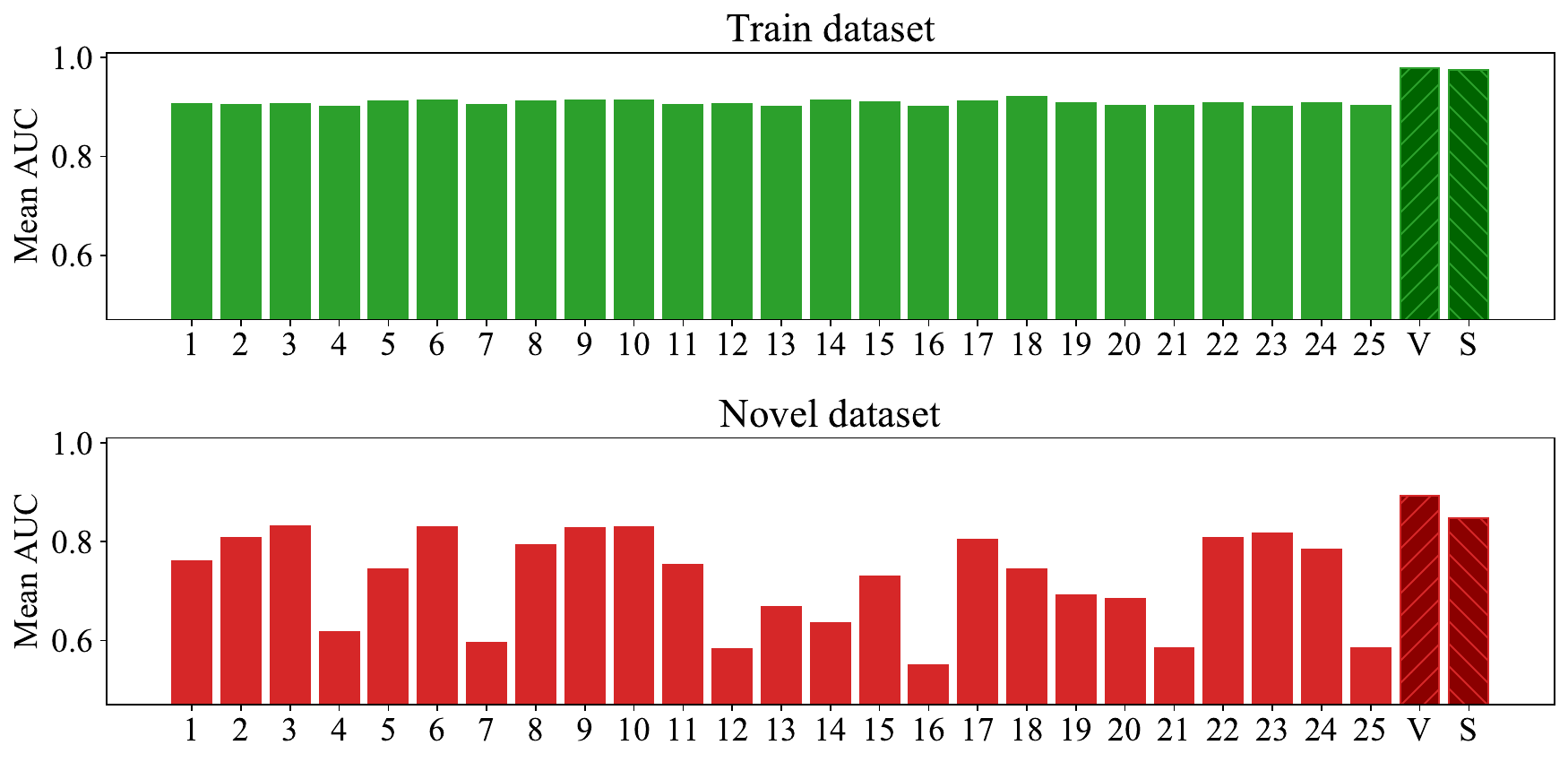}
\caption{Alzheimer's dataset pairs.}
\label{fig:alzrashomonens}
\end{subfigure}
\caption{Comparison of model performances across datasets. Each constituent model is represented by the Cluster from which it hailed. In the train datasets, we can observe that all constituent models behave similarly. On the novel datasets, under the unknown $U$ distribution, performance becomes unpredictable. However, we can verify that the voting approach always outperforms the best constituent model, thus presenting itself as a suitable technique to mitigate this behavior.}
\label{fig:rashoens}
\end{figure}

Figure \ref{fig:rashoens} displays performance comparisons for each base model and ensemble on the COVID-19 and Alzheimer's datasets. While all constituent models performed similarly on training data, voting consistently outperformed stacking on new, unseen data, confirming our hypothesis regarding erratic behavior across models. Both ensemble techniques exceeded the performance of individual models and state-of-the-art methods.

\begin{figure*}
\centering
\begin{subfigure}{0.45\linewidth}
\includegraphics[width=0.9\linewidth]{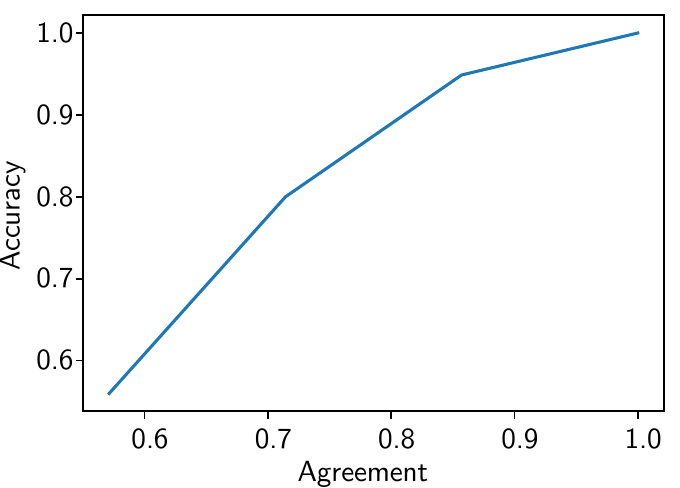}
\caption{COVID-19 production dataset.}
\label{fig:covidrashomonens}
\end{subfigure}
\begin{subfigure}{0.45\linewidth}
\includegraphics[width=0.9\linewidth]{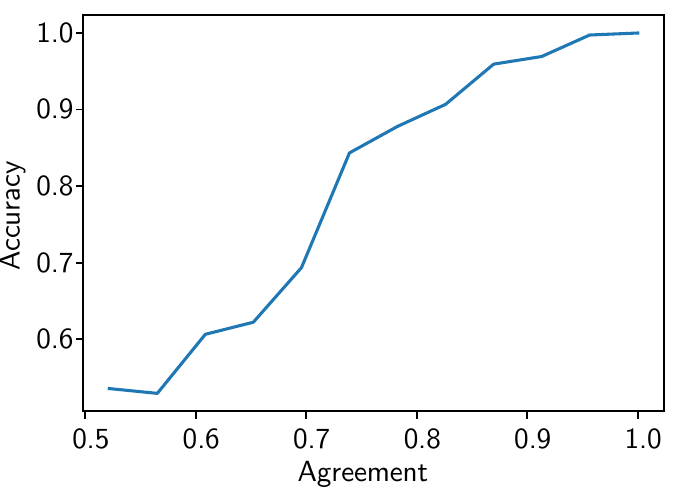}
\caption{Alzheimer's production dataset.}
\label{fig:alzrashomonens}
\end{subfigure}
\caption{Relationship between Rashomon Ensemble accuracy and intra-constituent agreement. As we hypothesized, there exists a direct relationship between ensemble performance and agreement. When constituents agree, accuracy is close to 1, implying that the observed instance is similar to what was seen in training, and we can trust the prediction with high confidence.}
\label{fig:agreementras}
\end{figure*}

To understand the relationship between model agreement and prediction confidence, we stratified test data points based on ensemble agreement and evaluated performance as shown in Figure \ref{fig:agreementras}. A direct correlation was observed between ensemble performance and intra-constituent agreement. When the models agreed, ensemble accuracy approached 1, indicating high confidence. Conversely, as agreement neared $50\%$, ensemble accuracy resembled random guessing, supporting our hypothesis that model agreement is critical for prediction reliability.
This relationship has significant implications for deploying Rashomon ensembles in dynamic data environments. Consensus among constituent models suggests instances similar to those encountered during training, enhancing trust in predictions. Conversely, divergence indicates that observations may fall outside training data, leading to unreliable predictions.

\subsection{Determinants of energy consumption}

Energy systems are increasingly linked with economic, social, and climate factors. Understanding these interconnections is crucial for electricity planning, particularly regarding how they impact electricity consumption and supply. Climate and weather significantly influence energy demand, with temperature being a key variable \citep{davies1959relationship, hor2005analyzing}. Various temperature-related metrics effectively approximate energy consumption \citep{huang1986climatic}. While degree days are commonly used for load forecasting, recent studies indicate that other weather variables, such as humidity, also affect electricity demand, particularly during hot days \citep{maia2020critical, WOODS2022726}. This suggests a range of potential weather predictors, indicating the Rashomon Effect and aligning with our analytical approach. Given the strong connection between weather and Brazil's electrical system, we focus on weather determinants of consumption, as presented in our previous work~\citep{zuinenergystanford, zuinenergystanford2}.

We utilized two primary datasets in our study: the Brazilian National Energy System Operator (ONS) historical reports \citep{ons2018operador} and the ERA5 reanalysis \citep{hersbach2020era5}. Energy data from the ONS website spans from 1999 to the present, including daily measures of load, maximum consumption, mean and total daily megawatts (MWd), and hourly megawatts (MWh). The ERA5 dataset offers global hourly estimates from 1950-2021 for atmospheric variables at a spatial resolution of 0.25 degrees (approximately a 30 x 30 km grid). Our final weather feature subset includes daily temperature minimums, means, and maximums, humidity, wind speed, precipitation, heating degree days (HDD), cooling degree days (CDD), heat index, wind chill index, apparent temperature, and the derived HDD and CDD from respective indices.

The primary objective is to predict consumption in the absence of abnormal events, enabling a direct comparison between predicted and actual consumption. We formulate this as a regression problem. Given a set $w \in W$ of weather descriptors and a set $t \in T$ of time descriptors, we apply a function $f(w; t; \sigma)$ parameterized by $\sigma$ to map a period to consumption. To exclude disruptive factors, we identify optimal subsets $W' \subset W$ and $T' \subset T$. 

We propose three main groups of factors influencing electricity consumption based on existing literature: load growth, historical events, and weather~\citep{giannakopoulos2006trends}. We observed a logistic growth trend in yearly energy consumption. Normalizing daily consumption by the load growth function, derived from yearly load interpolation while filtering atypical events, enables the construction of a counterfactual model focused on weather and temporal factors. This allows for the development of various models $f'(w; t; \sigma')$ with different feature sets, forming an ensemble that captures potential explanation biases in line with the Rashomon Effect.

\begin{figure}
\centering
\begin{subfigure}{0.48\linewidth}
\centering
\includegraphics[width=0.85\linewidth]{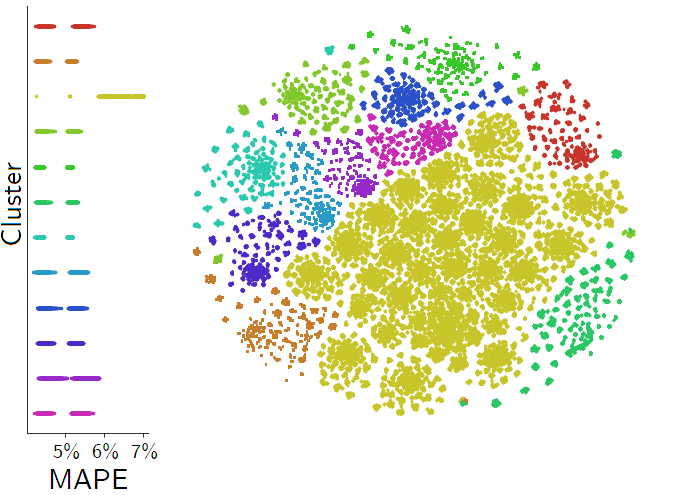}
\caption{$\epsilon=7\%$.}
\label{fig:stanfordrashomonA}
\end{subfigure}
\begin{subfigure}{0.48\linewidth}
\centering
\includegraphics[width=0.85\linewidth]{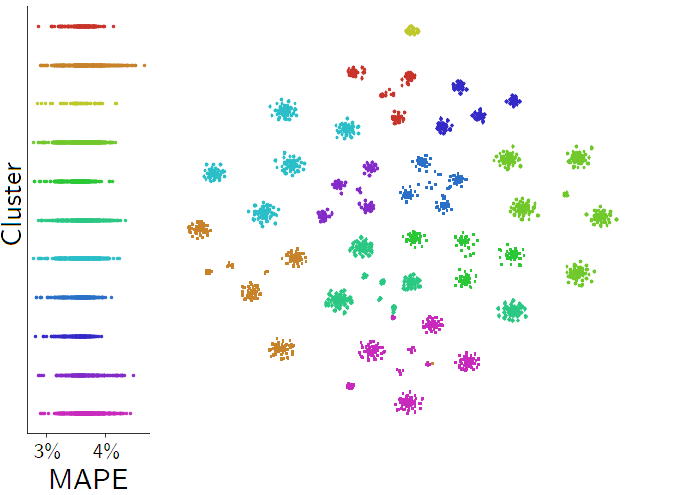}
\caption{$\epsilon=5\%$.}
\label{fig:stanfordrashomonB}
\end{subfigure}
\caption{Induced Rashomon spaces with $\epsilon$ thresholds of 7\% and 5\% MAPE. Overestimating $\epsilon$ results in a larger Rashomon space and undesirable correlations between cluster assignments, explanatory factors, and performance.}
\label{fig:stanfordrashomon}
\end{figure}

The first step in building our ensemble involves sampling models to estimate the Rashomon space. We chose to use the Mean Average Percentile Error (MAPE) for inducing Rashomon Sets. Figure \ref{fig:stanfordrashomon} illustrates the found Rashomon space after sampling 100\,000 models and also depicts the impact of different choices for the MAPE $\epsilon$ threshold. For MAPE of $7\%$, we found a Rashomon ratio of $.82$ (82\,114 models presented $MAPE\leq0.07$) while decreasing this threshold to $5\%$ reduced the ratio to $.04$ (4\,064 models presented $MAPE\leq0.05$). In Figure \ref{fig:stanfordrashomonA}, we noted that underperforming models clustered together, hinting that including these models in the ensemble might not be productive. By properly tuning the $\epsilon$ value, we reduced the Rashomon space and extracted more meaningful clusters, as depicted in Figure \ref{fig:stanfordrashomonB}.

\begin{figure}
\includegraphics[width=\linewidth]{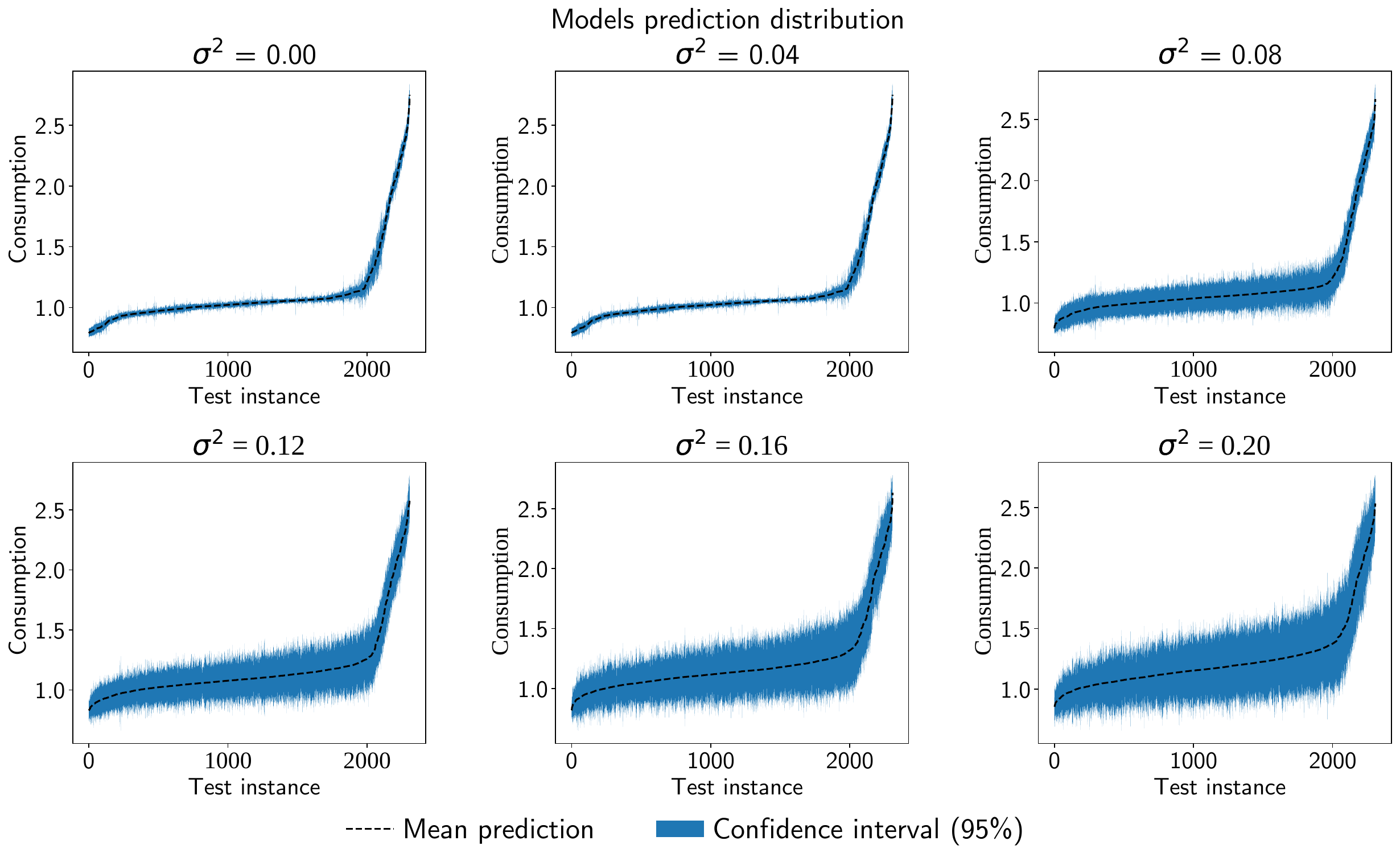}
\caption{Effect of noise on ensemble constituents' input features and predictions for models trained to predict Brazilian energetic consumption.}
\label{fig:sigmastanford}
\end{figure}

Following our algorithm, we searched for optimal representatives within each explanation cluster. Once again, we represented the model space as a directed acyclic graph (DAG) and searched for optimal constituents. In Figure \ref{fig:sigmastanford}, we added Gaussian noise to the normalized features and observed the normalized consumption estimates from each model. In the absence of noise, all models behaved similarly within a narrow confidence interval. However, introducing noise led to increased divergence among models, confirming the direct relationship between noise levels and the width of the confidence interval, indicating reduced ensemble reliability. Although the mean ensemble prediction remained stable, minor noise introduced significant prediction variability.

As shown in all the other experiments, agreement is a key metric to measure the reliability of the Rashomon ensembles. However, defining agreement in regression contexts is challenging since visual inspections of confidence intervals are impractical. While in classification, agreement occurs when two models predict the same class; in regression, we consider predictions 'similar' if they are within a context-dependent range. We selected the coefficient of variance ($C_V$) between regressors as our agreement measure, defined as:

\begin{equation}
C_V = \frac{\sigma}{\mu}
\end{equation}

Here, $\sigma$ represents standard deviation, while dividing by $\mu$ provides a dimensionless metric indicating the variability extent relative to population means. A higher $C_V$ reflects greater dispersion and disagreement between constituents. In a voting scheme, the ensemble prediction is $\mu$, while $C_V$ indicates the degree of divergence among individual constituents.

\begin{figure}
\centering
\includegraphics[width=0.75\linewidth]{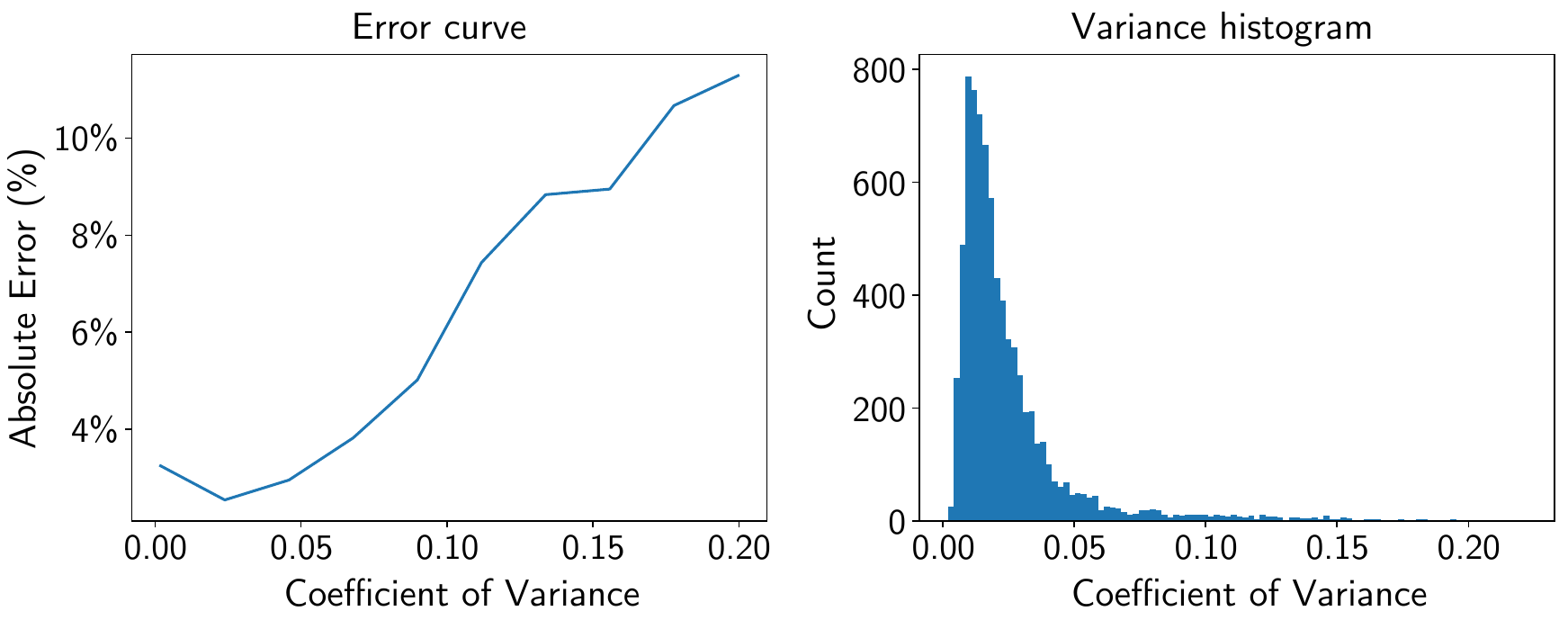}
\caption{Relationship between constituent agreement and ensemble performance when predicting energetic consumption. The coefficient of variance between constituent predictions was used as an agreement metric.}
\label{fig:stanfordacccurve}
\end{figure}

We validated our ensemble by comparing performance with constituent prediction dispersion $C_V$, as shown in Figure \ref{fig:stanfordacccurve}. A direct relationship between these metrics was observed, with most instances below $C_V=0.05$, indicating less than 5\% dispersion among ensemble constituents. In such scenarios, we anticipate an MAPE below 4\%, which is favorable. As dispersion increased, error escalated, confirming that disagreement among constituents compromises prediction reliability.

\begin{figure}
\centering
\begin{subfigure}{0.48\linewidth}
\centering
\includegraphics[width=\linewidth]{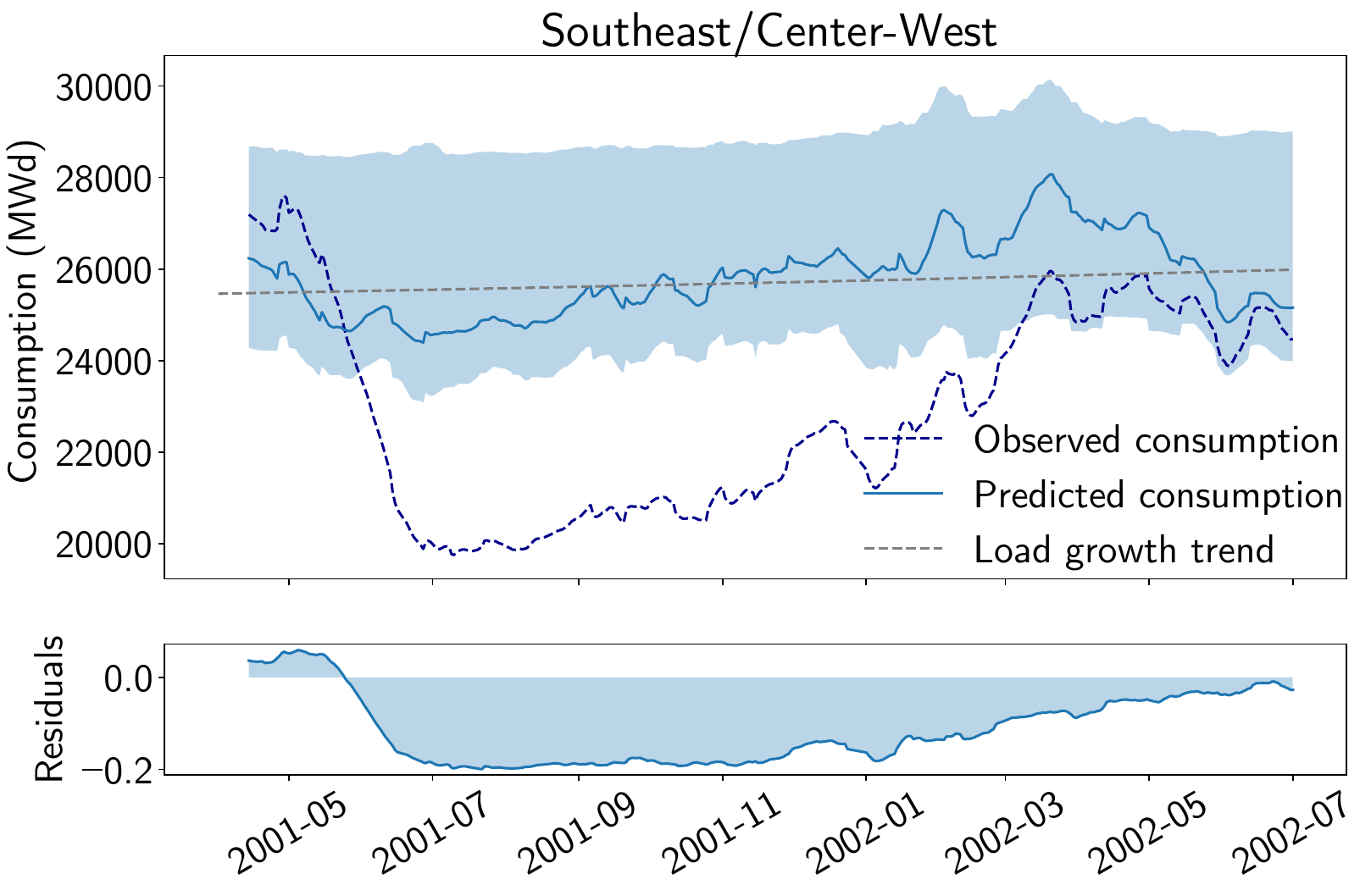}
\label{fig:sud2001}
\vspace*{-0.285cm}
\end{subfigure}
\begin{subfigure}{0.48\linewidth}
\centering
\includegraphics[width=\linewidth]{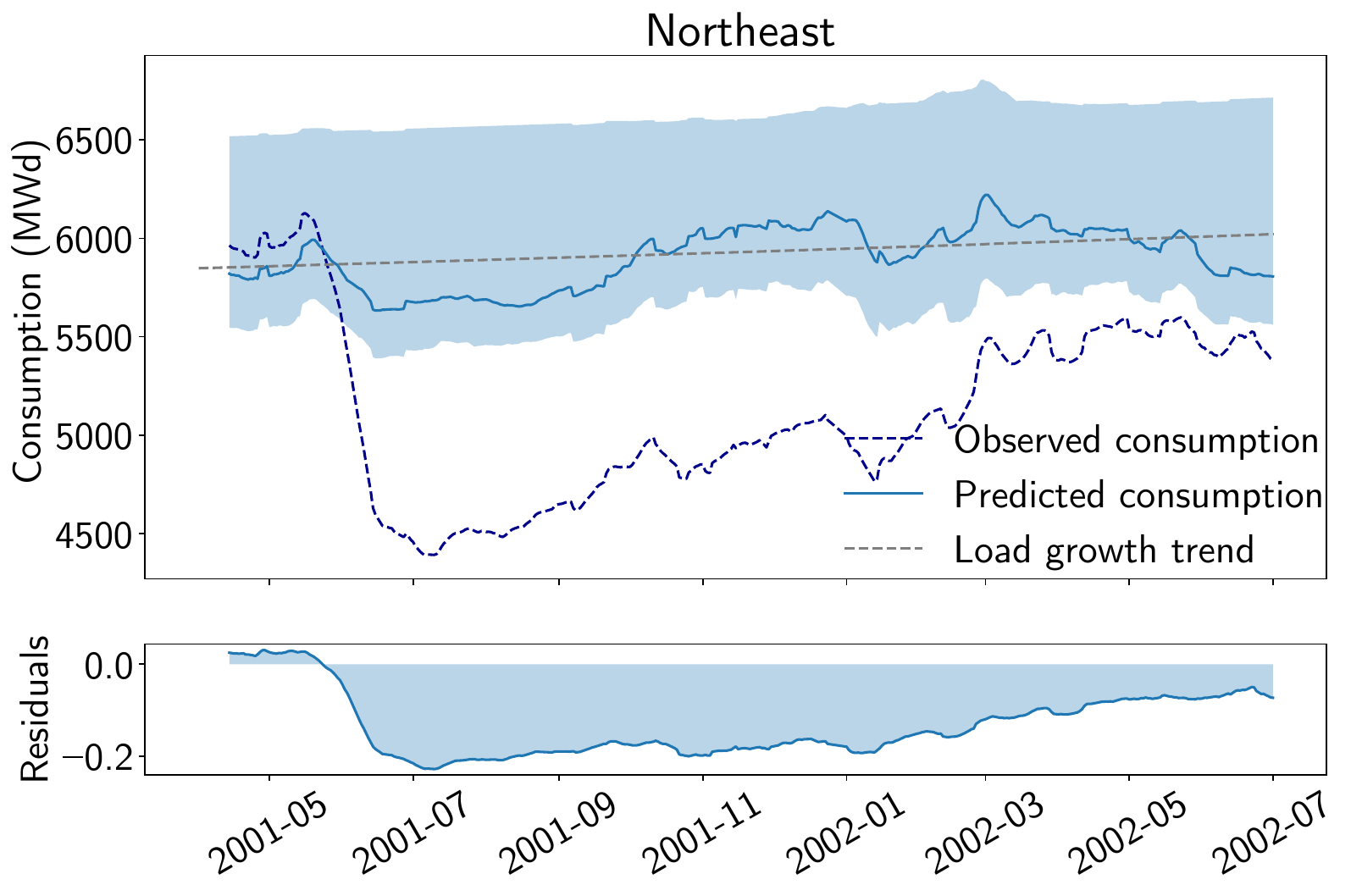}
\label{fig:sul2001}
\vspace*{-0.3cm}
\end{subfigure}
\begin{subfigure}{0.48\linewidth}
\centering
\includegraphics[width=\linewidth]{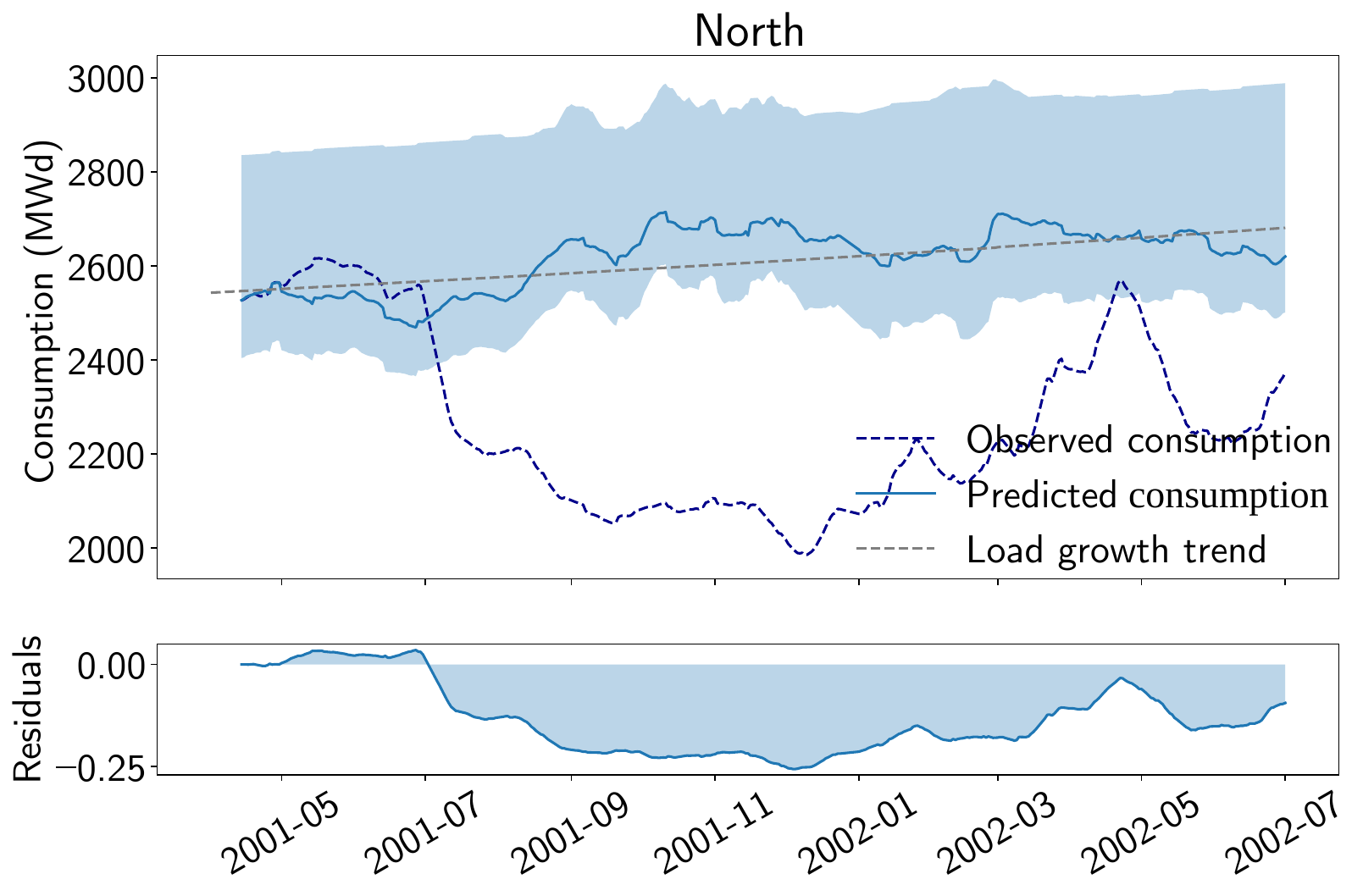}
\label{fig:sul2001}
\vspace*{-0.3cm}
\end{subfigure}
\begin{subfigure}{0.48\linewidth}
\centering
\includegraphics[width=\linewidth]{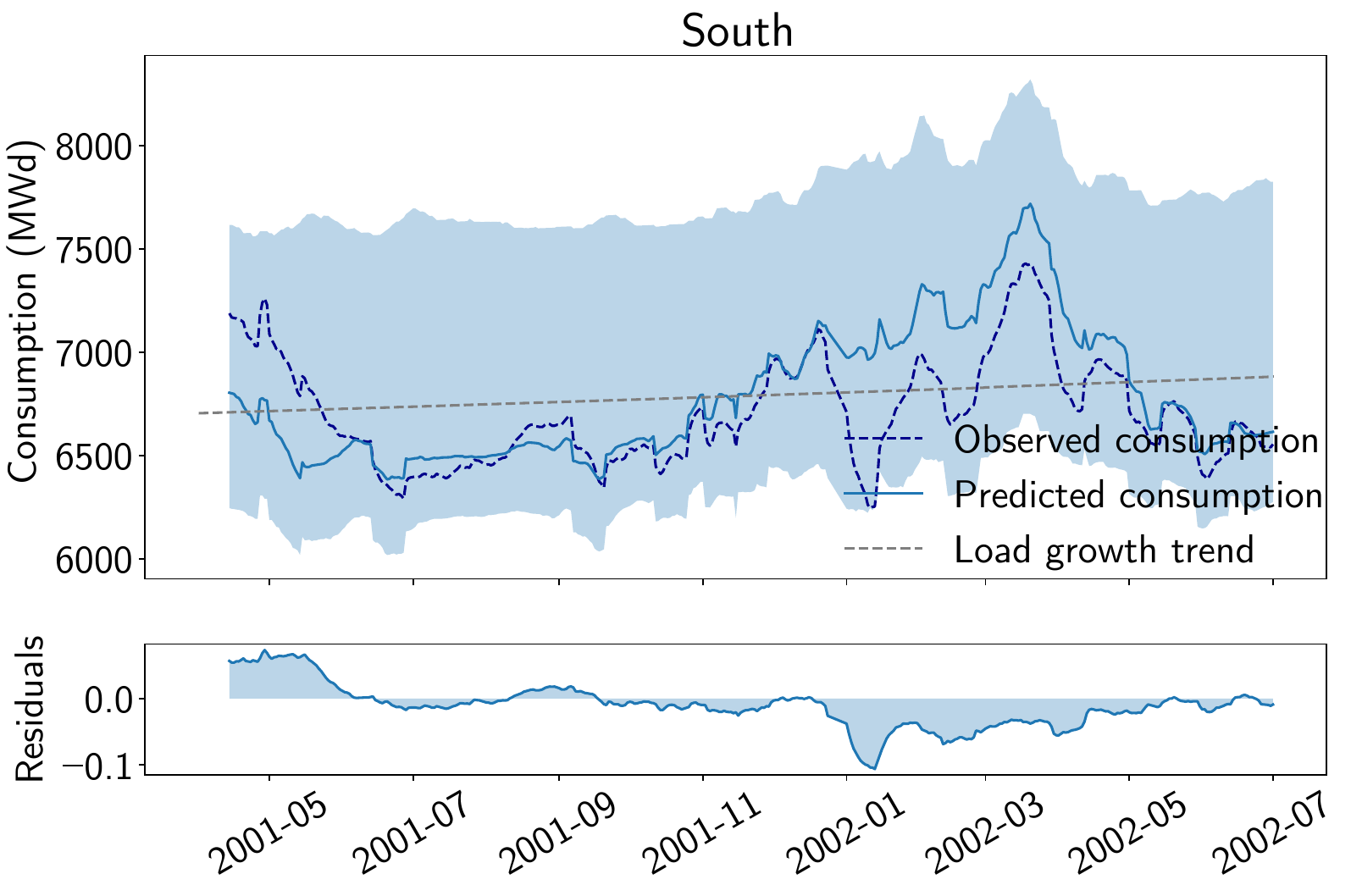}
\label{fig:sul2001}
\vspace*{-0.3cm}
\end{subfigure}
\caption{Consumption during Brazil's 2001 \textit{Apag\~ao} (Blackout). The South region did not adhere as closely to the restrictions as the other regions.}
\label{fig:2001}
\end{figure}

We evaluate the robustness of our approach through a case study in Brazil from 2001 to 2002. Severe drought and low reservoir levels in early 2001 raised concerns about a potential grid collapse, prompting the Federal Government to implement policies aimed at reducing energy consumption by 20\% \citep{bardelin2004efeitos}. These measures included awareness campaigns, peak-hour price increases, and incentives to limit non-essential energy use. This period provides a clear opportunity to measure expected impacts, allowing for a direct evaluation of our method. Figure \ref{fig:2001} illustrates the effects of these policies across regions. Our counterfactual model revealed mean and median relative residuals for the Center-West region from June 2001 to January 2002 of $-18.1\% (\pm1.6\%)$ and $-18.6\%$, closely aligning with the anticipated $-20\%$ reduction during the restriction period from July 1, 2001, to February 19, 2002. Similar trends were observed in the North and Northeast regions, with relative residuals of $-19.1\%$ and $-18.9\%$, respectively.

\begin{figure}
\centering
\includegraphics[width=0.5\linewidth]{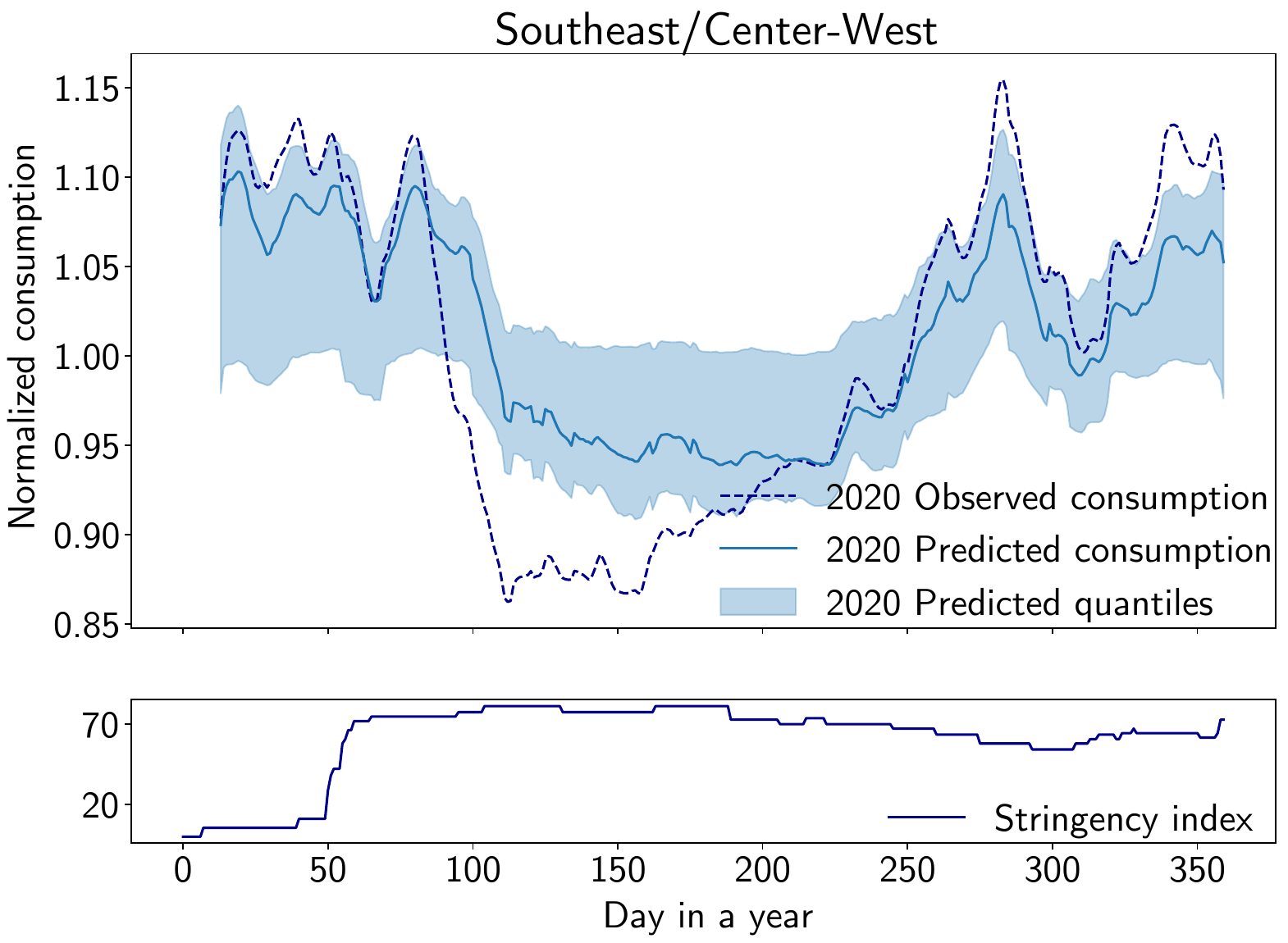}
\caption{Impact of COVID-19 on Brazil in 2020. Stringency data from \citet{owidcoronavirus}.}
\label{fig:sud2020}
\vspace*{-0.2cm}
\end{figure}

We also examine the year 2020, particularly during the early COVID-19 pandemic, which imposed significant mobility restrictions and reduced global GDP. Figure \ref{fig:sud2020} compares electricity consumption with the Oxford Stringency Index, which measures the strictness of COVID-19 policies \citep{hale2020variation}. During this period, there was a notable relationship between the drop in consumption and the Stringency Index in Brazil. From April to June, electricity consumption showed a mean relative residual of $-8.87\% \pm1.2\%$, in accordance with economic records~\citep{gullo2020economia}.

\begin{figure}[!hb]
\centering
\includegraphics[width=0.5\linewidth]{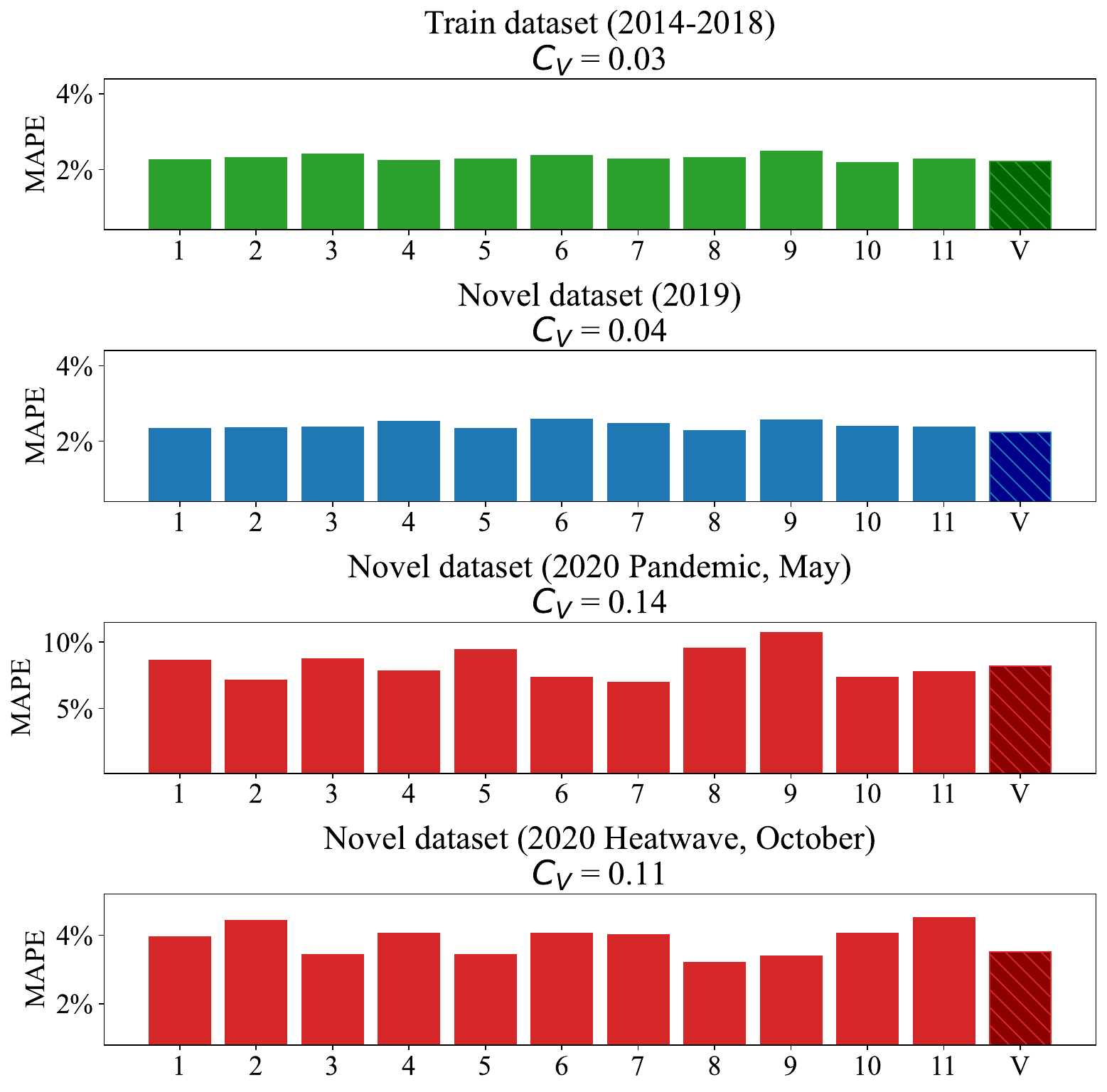}
\caption{Comparison of model performances across periods, with each constituent model represented by its cluster. In 2019, models showed low MAPE; in 2020, individual performances became erratic, indicating data from different distributions due to the October heatwave.}
\label{fig:2020pandemicens}
\end{figure}

The lack of demographic and behavioral variables enables us to assess the pandemic's impact through a counterfactual approach, with the expectation of uniform constituent errors in 2020. This period demonstrates the usefulness of the Rashomon ensembles and our proposed approach in contrast to other counterfactual techniques. Figure \ref{fig:2020pandemicens} shows ensemble and constituent performances when trained on data from 2014 to 2018 and applied to 2019 and 2020. As expected, errors for 2019 are similar to those seen in training, meaning similar weather variable distributions. However, in 2020, the Rashomon ensemble displayed erratic behavior, with a coefficient of variance rising from $0.04$ in 2019 to $0.14$ in May 2020, suggesting atypical weather patterns. Since this pertains to the high point of restrictions in Brazil, we can conclude that the pandemic effect shadowed this change in weather. In October, Brazil experienced one of its most intense heatwaves in history, breaking century-old temperature records~\cite{marengo2022heat}. Such extremes diverged from the 2014-2018 distribution. If May's prediction errors were solely pandemic-related, we would expect consistent performance across constituents. However, the erratic behavior implies that the weather throughout the year was unusual and evident months before the heatwave.
    
\subsection{Auditing medical bills in large healthcare companies}

Auditing hospital and outpatient bills is critical for identifying errors and overcharges, but the high volume of daily bills makes comprehensive auditing challenging \cite{zhang2017anomaly, van2016outlier, kirlidog2012fraud}. We collaborated with Unimed-BH, the leading private healthcare provider in Minas Gerais, Brazil, to address this issue. With over 1.5 million patients, only a small fraction of bills are manually reviewed, leading to missed discrepancies and inefficient use of expert auditors. Automating this process allows auditors to focus on the most complex cases.

We developed a tool to rank bills based on a “discrepancy score”, estimating the likelihood of inconsistencies~\cite{zuinunimed}. Bills from Unimed-BH’s accredited network are evaluated using a Rashomon ensemble, where each model contributes a distinct perspective on inconsistency. We define a dataset \(X = \{x_1, x_2, \dots, x_N \}, \ x_i \in \mathbb{R}^D\) and a representation space \(Z \in \mathbb{R}^K (K \ll N)\). The score function \(\tau(\cdot): X \rightarrow \mathbb{R}\) ranks bills by inconsistency likelihood, using an ensemble of models tuned to different features. Each model \(i\) maps data to \(Z\) (e.g., via autoencoders) or generates a score \(\tau_i(\cdot)\) (e.g., via isolation forests), which informs the ranking of \(X\).

Unlike in our previous experiments, we explored how to merge multiple algorithms in a single ensemble and weigh their importance. We employ a multi-armed bandit (MAB) approach, balancing exploration (auditing lower-ranked bills) and exploitation (focusing on high-ranked bills). This method allows us to incorporate varied algorithms without directly comparing their applicability \cite{bouneffouf2020survey}. The ensemble models inconsistency detection as a ranking task, where high-ranked bills are more likely to contain misaligned financial values. The MAB algorithm assigns weights to each model based on its performance with labeled data, combining models optimally to reflect diverse explanations. The final ensemble uses a weighted voting method, dynamically adapting to new data. Constituent models include:

\begin{itemize}
    \item Generative methods: PCA, Denoising AutoEncoders.
    \item Regressor methods: Elastic Net, Lasso, Support Vector Regressor, XGBoost, LightGBM.
    \item Isolation methods: Isolation Forest, K-Nearest Neighbors.
\end{itemize}

\begin{figure}
    \centering
    \includegraphics[width=0.6\linewidth]{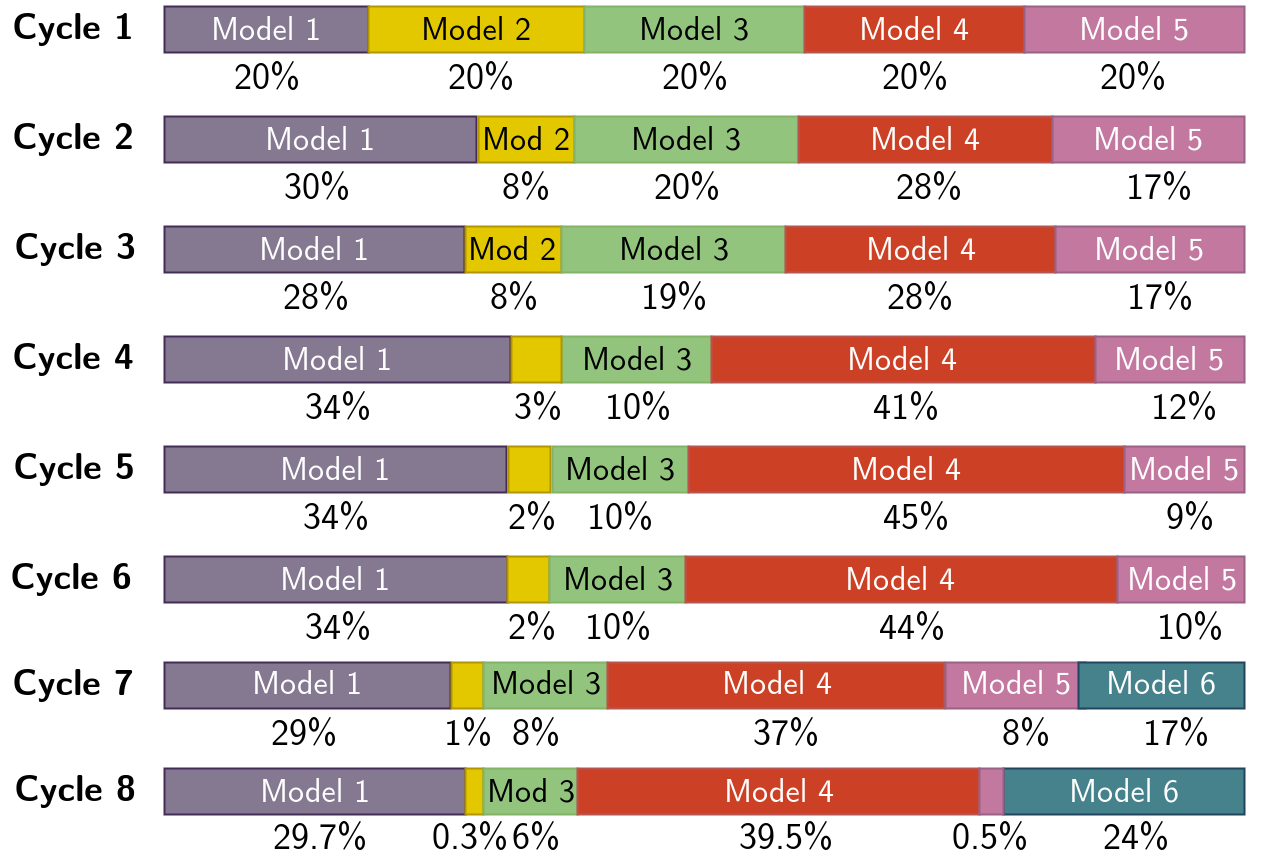}
    \caption{Multi-armed bandit approach to assign ensemble weights. Models with inferior performance are penalized, and new models can be added.}
    \label{fig:mab}
\end{figure}

Given that our data is only partially labeled (only monetary values of audited bills are known), we use this subset to tune model weights. As new patterns of inconsistency emerge monthly, the MAB Rashomon ensemble ensures that models with stronger predictive power receive higher weights, while less effective models are penalized, as shown in Figure \ref{fig:mab}. This setup also accommodates the integration of new models and maintains robustness against performance drift and degradation \cite{gashler2008decision, huang2009research}.

\begin{figure}
    \centering
    \includegraphics[width=0.5\linewidth]{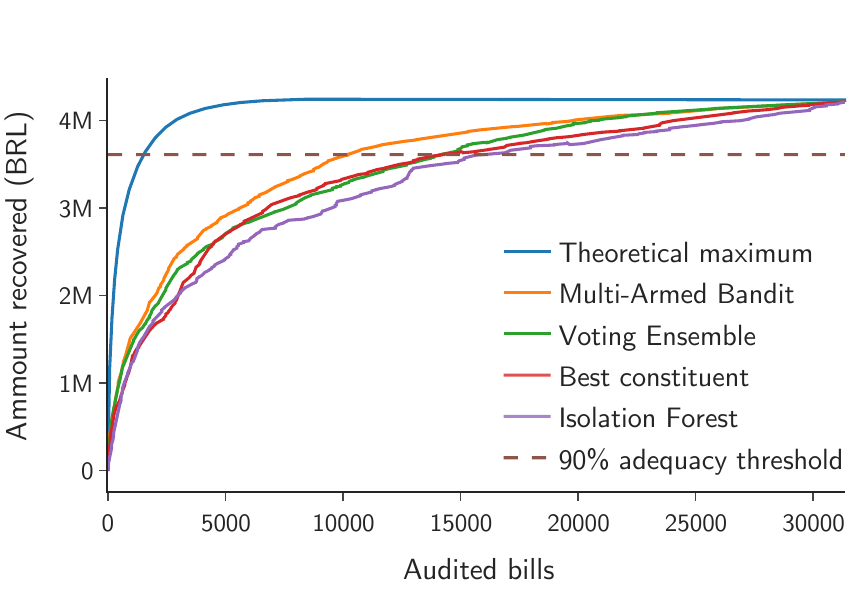}
    \caption{Performance of the proposed approach in Brazilian Reais (BRL). The ensemble requires 28\% fewer bills to reach the 90\% adequacy threshold compared to the best individual model.}
    \label{fig:recovered}
\end{figure}

Our initial experiments tested the hypothesis that combining models would outperform any single model in detecting inconsistencies and evaluated the monetary recovery potential of this approach. In February and March 2023, Unimed audited 31,355 hospital bills, with recorded adequacy values (in Brazilian Reais, BRL) indicating the recovery amount from inconsistencies. To establish a benchmark, we ranked bills by adequacy values to set a theoretical upper bound for recovery. We also used a baseline ensemble method where all constituent models contributed equally, allowing comparison with the MAB approach. Figure \ref{fig:recovered} illustrates the results of comparing our MAB Rashomon ensemble with other approaches in terms of adequacy. The baselines consisted of the theoretical maximum recovery, the voting scheme approach used previously, an oracle system selecting the best constituent before deployment, and a literature-based anomaly detection approach.

The analysis showed that reaching the 90\% adequacy threshold - the minimum level for practical applicability - required 2,178 bills by the theoretically optimal ranking. In comparison, the Multi-Armed Bandit (MAB) ranking needed 10,219 bills, while the baseline voting scheme required 14,190 bills to achieve the same threshold. Despite these differences, all methods significantly outperformed traditional anomaly detection approaches such as isolation forests, as well as a naive baseline in which bills are audited in order of arrival, configuring a system without a priority ranking. Using the MAB Rashomon ensemble could reduce the number of bills audited by nearly 62\% from the naive approach, from over 26,000 to approximately 10,000, while meeting the 90\% adequacy threshold.

We proceeded to deploy our approach in Unimed's pipeline under specific constraints. The model was trained on bills from the previous three months, and daily rankings of medical bills were generated for auditing. To account for processing delays, we use a D-3 sliding-window rule, which generates the final daily ranking using the last three days' worth of bills. We accompanied the results of this deployment for 5 months, from April to August 2022. During this period, the solution alone recovered R\$1,571,146 (USD 349,610) that would otherwise have been lost. From April to August, Unimed’s current rule-based system and the Rashomon solution analyzed 8,570 bills. Of these, 3,327 were flagged by our algorithm, identifying 665 more inconsistent bills than the rule-based system. Unimed's business area decided that all bills flagged by the algorithm should be audited monthly, although only an average attainment rate of 89.9\% was achieved due to workforce constraints. However, this resulted in an increase of 63.4\% in the number of bills flagged for auditing, with 39\% of these bills being flagged exclusively by our algorithm. The monetary recovery increased by 38\%, with the highest adequacy values aligning with the months where the auditors prioritized the bills recommended by our solution. These results further demonstrate the robustness and effectiveness of the Rashomon ensembles in real-world problems.

\section{Conclusion}

In this study, we proposed a novel approach for ensemble learning based on explainability that enables estimating the prediction risk in production. We address the challenge of model selection by identifying a Rashomon subset of models that perform similarly but process data differently. By inducing perturbations on a held-out test set, we simulate out-of-distribution data and assess ensemble loss of predictive power as constituent models diverge. Our approach relies on ensemble diversity, leveraging that our constituents' behavior may diverge when faced with data from distributions that do not match the one seen in training. 

An extensive evaluation across various tasks demonstrated the effectiveness of our Rashomon ensemble, especially in scenarios with multiple local structures. In such cases, our approach consistently outperformed other state-of-the-art tree-based ensembling techniques, showcasing the capabilities of our approach. Even in scenarios in which local structures were less prevalent, our ensembles proved robust, maintaining high-performance levels. This adaptability becomes particularly valuable when predicting in domains where the inherent data generation functions may differ from what was observed during training.

Nevertheless, we acknowledge instances where our approach faced challenges. Specifically, when the Rashomon ratio - the proportion of models retained in the Rashomon Set - was relatively small. The scarce diversity among the constituent models limited performance gains. This finding emphasizes the importance of carefully considering the Rashomon ratio and the diversity of explanatory factors. While our approach consistently achieved superior results in most of our experiments, care should be exercised when deploying it in domains with a low Rashomon ratio. Alternative ensemble methods or model selection strategies may be more suitable in such cases.
Therefore, we stress the importance of understanding each problem domain and evaluating the ensemble's diversity and performance before deployment. Our experiments have also revealed a direct relationship between model agreement and prediction accuracy.

Finally, we emphasize the importance of expert input in refining the final model and sets of variables, which resulted in a patent~\citep{liapatente}. Due to our focus on explicability, we observed that when inducing our Rashomon Ensembles, some expert-known patterns frequently emerged among the constituents. This not only helped gain the experts' trust but also led to more insightful discussions regarding the remaining learned patterns. This was a main factor in achieving a tangible impact on business and a core aspect of the patent. Its applicability is demonstrated in both the stainless steel case study, which resulted in significant improvements in production processes, and the Unimed one, with gains of over $R\$1.5$ million across 5 months.

\section*{Code and Data Availability}
\label{sec:availability}

The code used for all machine learning analyses, made available for non-commercial use, has been deposited at \url{https://doi.org/10.6084/m9.figshare.30081913}~\cite{figshareTKDD2025}. 
The datasets employed in this study are accessible as follows:
\begin{itemize}
    \item \textbf{Open datasets:} Available directly from the UCI Machine Learning Repository~\cite{asuncion2007uci} and the OpenML database~\cite{bischl2017openml}.
    \item \textbf{\textit{FAPESP} COVID-19 datasets:} Accessible upon request via \url{covid19datasharing@fapesp.br}.
    \item \textbf{Alzheimer datasets:} See details in~\cite{ismaelalzheimer2022}.
    \item \textbf{\textit{APERAM South America} stainless steels datasets:} See details in~\cite{zuinaperam}.
    \item \textbf{\textit{Grupo Fleury} COVID-19 datasets:} See details in~\cite{zuinnature}.
    \item \textbf{Brazilian energy datasets:} See details in~\cite{zuinenergystanford2}.
    \item \textbf{\textit{Unimed-BH} medical bills datasets:} See details in~\cite{zuinunimed}.
\end{itemize}

All datasets were used strictly in accordance with their respective terms of use, and any restrictions on data redistribution are noted in the corresponding references.

\section*{Acknowledgements}
This work was funded by the authors' individual grants from Kunumi. \textit{APERAM South America}, \textit{Grupo Fleury}, and \textit{Unimed-BH} kindly granted access to the data needed for the development of the case studies in this work. 

\bibliographystyle{ACM-Reference-Format}
\bibliography{bigbib}

\newpage
{\color{white}.}
\end{document}